\documentclass{article}

\usepackage[final,nonatbib]{neurips_2022}

\usepackage{caption}
\usepackage[utf8]{inputenc} %
\usepackage[T1]{fontenc}    %
\usepackage{hyperref}       %
\usepackage{url}            %
\usepackage{booktabs}       %
\usepackage{amsfonts}       %
\usepackage{nicefrac}       %
\usepackage{microtype}      %
\usepackage{xcolor}         %

\usepackage{color}
\usepackage{amssymb}%
\usepackage{pifont}%
\usepackage{multirow}
\usepackage[pdftex]{graphicx}
\usepackage{gensymb}

\usepackage{amsmath}
\usepackage{subcaption}
\usepackage{bm}
\usepackage{mathrsfs}
\usepackage{makecell}
\usepackage{colortbl}
\usepackage{wrapfig}
\usepackage{float}
\usepackage{soul}
\usepackage[permil]{overpic}
\usepackage{tikz}
\usepackage{tabularx}
\usepackage{lipsum}
\usepackage{siunitx}
\usepackage{textcomp}
\usepackage{enumitem}
\usepackage{booktabs}
\usepackage{array}\newcolumntype{C}{>{\centering\arraybackslash}X}

\definecolor{calmetallic}{RGB}{253,181,21}
\hypersetup{
    breaklinks=true,
    letterpaper=true,
    colorlinks,
    citecolor=calmetallic,
}

\newlength\savewidth
\newcommand{\tablestyle}[2]{\setlength{\tabcolsep}{#1}\renewcommand{\arraystretch}{#2}\centering\footnotesize}

\DeclareMathOperator*{\argmin}{arg\,min}

\definecolor{colorfirst}{rgb}{.866,.945, 0.831} %
\definecolor{colorsecond}{rgb}{1, 0.98, 0.83} %
\definecolor{colorthird}{rgb}{1, 0.9, 0.8} %

\newcommand{\texttd}{\raisebox{0.5ex}{\texttildelow}}

\newcommand{\eg}{\textit{e.g.}}
\newcommand{\ie}{\textit{i.e.}}
\newcommand{\roughly}{{\raise.20ex\hbox{$\scriptstyle\sim$}}}

\newcommand{\etal}{\textit{et al.}}
\newcommand{\authorskip}{\hspace{2.5mm}}

\newcommand{\dychecklink}{https://hangg7.com/dycheck}

\title{Monocular Dynamic View Synthesis: A Reality Check}

\author{
    Hang Gao$^{1,\dag}$ \authorskip
    Ruilong Li$^1$ \authorskip
    Shubham Tulsiani$^2$ \authorskip
    Bryan Russell$^3$ \authorskip
    Angjoo Kanazawa$^1$ \\[2mm]
    $^1$UC Berkeley \qquad
    $^2$Carnegie Mellon University \qquad
    $^3$Adobe Research
    \vspace{-4mm}
}

\begin{document}

\maketitle

\begin{abstract}
\looseness=-1

We study the recent progress on dynamic view synthesis (DVS) from monocular video.
Though existing approaches have demonstrated impressive results, we show a discrepancy between the practical capture process and the existing experimental protocols, which effectively leaks in multi-view signals during training.
We define {\em effective multi-view factors}~(EMFs) to quantify the amount of multi-view signal present in the input capture sequence based on the relative camera-scene motion.
We introduce two new metrics: co-visibility masked image metrics and correspondence accuracy, which overcome the issue in existing protocols.
We also propose a new iPhone dataset that includes more diverse real-life deformation sequences.
Using our proposed experimental protocol, we show that the state-of-the-art approaches observe a $1$-$\SI{2}{dB}$ drop in masked PSNR in the absence of multi-view cues and $4$-$\SI{5}{dB}$ drop when modeling complex motion.
Code and data can be found at~\url{\dychecklink}.

\end{abstract}

\section{Introduction}
\label{sec:intro}

Dynamic scenes are ubiquitous in our everyday lives -- people moving around, cats purring, and trees swaying in the wind. 
The ability to capture 3D dynamic sequences in a ``casual'' manner, particularly through monocular videos taken by a smartphone in an uncontrolled environment, will be a cornerstone in scaling up 3D content creation, performance capture, and augmented reality. 
{
    \let\thefootnote\relax\footnote{
        $^\dag$
        Work partially done as part of HG's internship at Adobe.
    }
}

\begin{figure}[t!]
    \includegraphics[width=\linewidth,bb=0 21 1555 540]{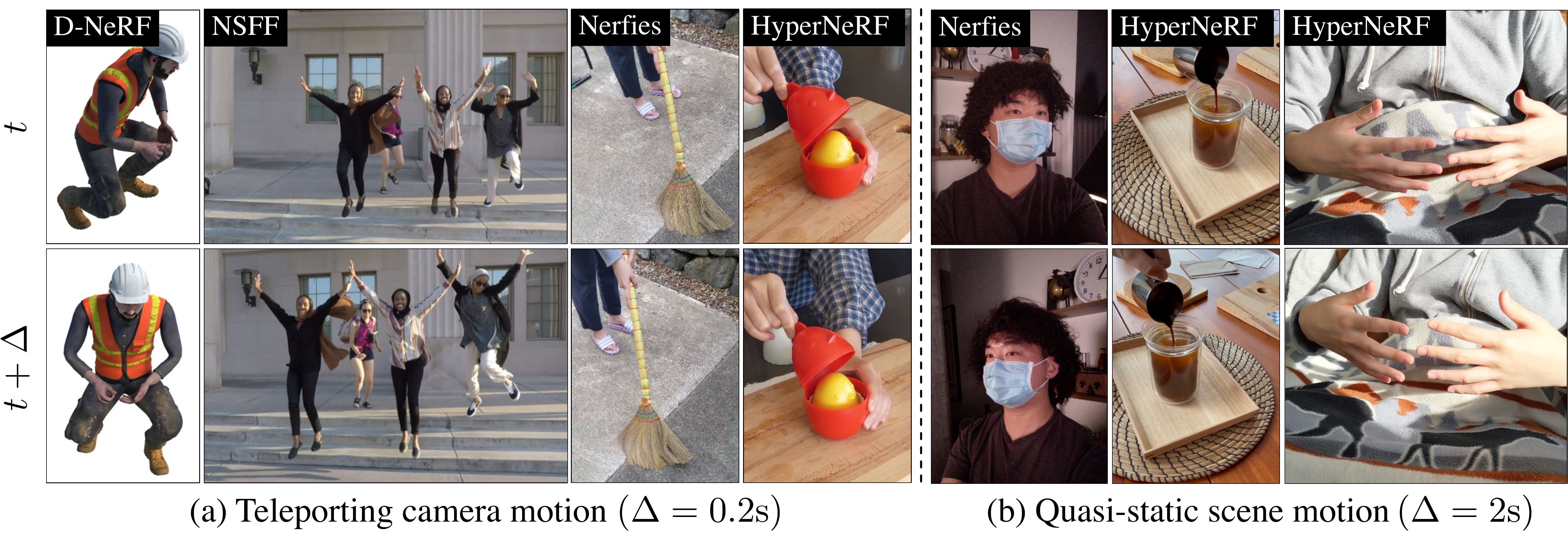}
    \vspace{-10pt}
    \caption{
        \textbf{Visualizing the training data of existing benchmarks.}
        Existing datasets operate under the effective multi-view regime:
        The sequences either have (a) teleporting camera motion or (b) quasi-static scene motion.
        These motions leak multi-view cues, \eg, the model can observe the human ($1^{\text{st}}$ column) and hands (last column) at roughly the same pose from different viewpoints.
    }
    \vspace{-1em}
    \label{fig:existing_benchmark_visuals}
\end{figure}

Recent works have shown promising results in dynamic view synthesis~(DVS) from a monocular video~\cite{xian2021space,tretschk2021non,pumarola2020dnerf,li2020nsff,park2021nerfies,gao2021dynamic,park2021hypernerf,wu2022d}. 
However, upon close inspection, we found that there is a discrepancy between the problem statement and the experimental protocol employed. As illustrated in Figure~\ref{fig:existing_benchmark_visuals}, the input data to these algorithms either contain frames that ``teleport'' between multiple camera viewpoints at consecutive time steps, which is impractical to capture from a single camera, or depict quasi-static scenes, which do not represent real-life dynamics.

In this paper, we provide a systematic means of characterizing the aforementioned discrepancy and propose a better set of practices for model fitting and evaluation. %
Concretely, we introduce {\em effective multi-view factors}~(EMFs) to quantify the amount of multi-view signal in a monocular sequence based on the relative camera-scene motion. %
With EMFs, we show that the current experimental protocols operate under an effectively multi-view regime. For example, our analysis reveals that 
the aforementioned practice of camera teleportation makes the existing capture setup akin to an Olympic runner taking a video of a moving scene without introducing any motion blur.

The reason behind the existing experimental protocol is that monocular DVS is a challenging problem that is also hard to evaluate.
Unlike static novel-view synthesis where one may simply evaluate on held-out views of the captured scene, 
in the dynamic case, since the scene changes over time, evaluation requires another camera that observes the scene from a different viewpoint at the same time. 
However, this means that the test views often contain regions that were never observed in the input sequence. 
Camera teleportation, \ie, constructing a temporal sequence by alternating samples from different cameras, addresses this issue at the expense of introducing multi-view cues, which are unavailable in the practical single-camera capture.

\looseness=-1 
We propose two sets of metrics to overcome this challenge without the use of camera teleportation. %
The first metric enables evaluating only on pixels that were seen in the input sequence by computing the co-visibility of every test pixel. 
The proposed co-visibility mask can be used to compute 
masked image metrics (PSNR, SSIM~\cite{wang2004image} and LPIPS~\cite{zhang2018unreasonable}).
While the masked image metrics measure the quality of rendering, they do not directly measure the quality of the inferred scene deformation.
Thus, we also propose a second metric that evaluates the quality of established point correspondences by the percentage of correctly transferred keypoints~(PCK-T)~\cite{kulkarni2019canonical}. 
The correspondences may be evaluated between the input and test frames or even within the input frames, which enable evaluation on sequences that are captured with only a single camera. 

We conduct extensive evaluation on existing datasets~\cite{park2021nerfies,park2021hypernerf} as well as a new dataset that includes more challenging motion and diverse scenes.
When tested on existing datasets without 
camera teleportation, %
the state-of-the-art methods observe a $1$-$\SI{2}{dB}$ drop in masked PSNR and $\texttd5\%$ drop in PCK-T.
When tested on complex motion with the proposed dataset, existing approaches observe another $4$-$\SI{5}{dB}$ drop in masked PSNR and $\texttd30\%$ drop in PCK-T, suggesting a large room for improvement.
We encourage future works to report EMFs on new data and adopt our experimental protocol to evaluate monocular DVS methods.
Code and data are available at our \href{\dychecklink}{project page}.

\section{Related work}
\label{sec:related}

\paragraph{Non-rigid structure from motion (NR-SfM).}
\looseness=-1
Traditional NR-SfM tackles the task of dynamic 3D inference by fitting parametric 3D morphable models~\cite{blanz1999morphable,ramakrishna2012reconstructing,bogo2016keep,kar2015category,cmrKanazawa18,kulkarni2020acsm,yang2021lasr,yang2021banmo}, or
fusing non-parametric depth scans of generic dynamic scenes~\cite{curless1996volumetric,newcombe2015dynamicfusion,li08global,gao18surfelwarp,bozic2020neuraltracking}.
All of these approaches aim to recover accurate surface geometry at each time step and their performance is measured with ground truth 3D geometry or 2D correspondences with PCK~\cite{yang2012articulated} when such ground truth is not available.
In this paper, we analyze recent dynamic view synthesis methods whose goal is to generate a photo-realistic novel view.
Due to their goal, these methods do not focus on evaluation against ground truth 3D geometry, but we take inspiration from prior NR-SfM works to evaluate the quality of the inferred 3D dynamic representation based on correspondences.
We also draw inspiration from previous NR-SfM work that analyzed camera/object speed and 3D reconstruction quality~\cite{park10moving,yucer15moving}.

\paragraph{Monocular dynamic neural radiance fields (dynamic NeRFs).}
\looseness=-1 
Dynamic NeRFs reconstruct moving scenes from multi-view inputs or given pre-defined deformation template~\cite{kwon2021neural,noguchi2021narf,peng2021neuralbody,su2021anerf,weng2022humannerf,gafni2021nerface,li2021neural,du2021neural}.
In contrast, there is a series of recent works that seek to synthesize high-quality novel views of generic dynamic scenes given a monocular video~\cite{xian2021space,tretschk2021non,pumarola2020dnerf,li2020nsff,park2021nerfies,gao2021dynamic,park2021hypernerf,wu2022d}.
These works can be classified into two categories: a deformed scene is directly modeled as a time-varying NeRF in the world space~\cite{xian2021space,li2020nsff,gao2021dynamic}
 or as a NeRF in canonical space with a time-dependent deformation~\cite{tretschk2021non,pumarola2020dnerf,park2021nerfies,park2021hypernerf,wu2022d}.
The evaluation protocol in these works inherit from the original static-scene NeRF~\cite{mildenhall2020nerf} that quantify the rendering quality of held-out viewpoints using image metrics, \textit{e.g.}, PSNR.
However, in dynamic scenes, PSNR from an unseen camera view may not be meaningful since the novel view may include regions that were never seen in the training view (unless the method can infer unseen regions using learning based approaches).
Existing approaches resolve this issue by incorporating views from multiple cameras during training, which we show results in an effectively multi-view setup.
We introduce metrics to measure the difficulties of an input sequence, a monocular dataset with new evaluation protocol and metrics, which show that existing methods have a large room for improvement.

\section{Effective multi-view in a monocular video}
\label{sec:effective}

We consider the problem of dynamic view synthesis~(DVS) from a monocular video.
A monocular dynamic capture consists of a single camera observing a moving scene.
The lack of simultaneous multi-view in the monocular video makes this problem more challenging compared to the multi-view setting, such as reconstructing moving people from multiple cameras~\cite{peng2021neuralbody,li2021neural,li2022tava}.

\looseness=-1
Contrary to the conventional perception that the effect of multi-view is binary for a capture (single versus multiple cameras), we show that it can be characterized on a continuous spectrum.
Our insight is that a monocular sequence contains \textit{effective} multi-view cues when the camera moves much faster than the scene, though technically the underlying scene is observed only once at each time step.

\subsection{Characterizing effective multi-view in a monocular video}
\begin{wrapfigure}{r}{0.6\linewidth}
    \centering
    \includegraphics[width=\linewidth,bb=10 30 894 240]{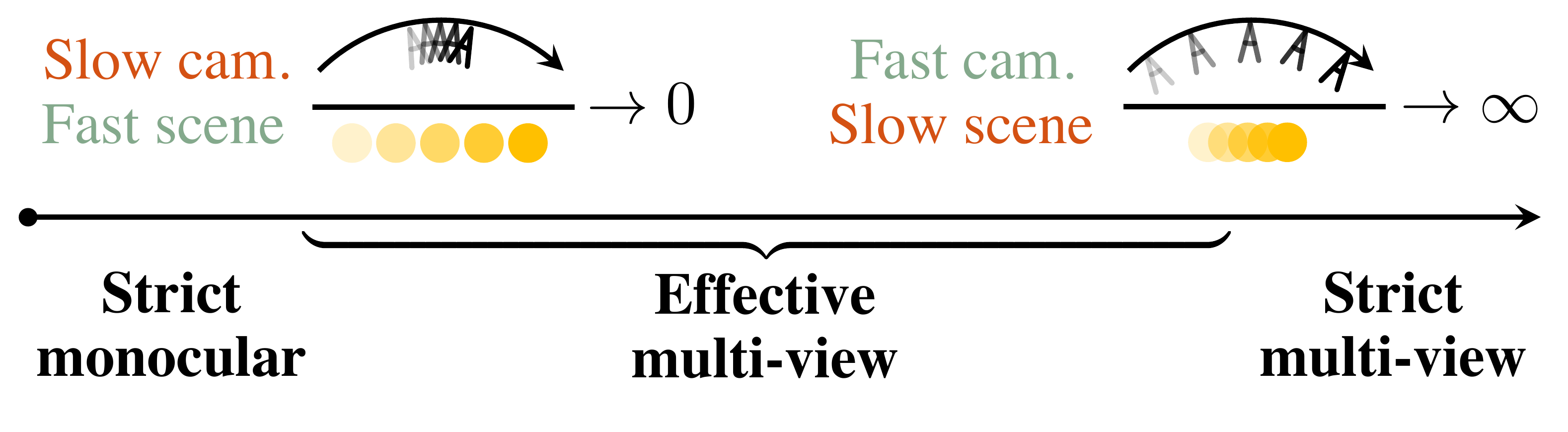}
    \vspace{-10pt}
    \caption{
        \textbf{The spectrum of effective multi-view in a monocular video.}
        A video captured by a single camera can still have multi-view cues when the camera moves much faster than the scene.We disentangle recent advances in DVS given a monocular video from such phenomena.
    }
    \label{fig:effective_multiview_spectrum}
\end{wrapfigure}

Although a monocular video only sees the scene from one viewpoint at a time, depending on the capture method, it can still contains cues that are effectively similar to those captured by a multi-view camera rig, which we call as effective multi-view.
As shown in Figure~\ref{fig:effective_multiview_spectrum},
when the scene moves significantly slower than the camera (to the far right end of the axis),  the same scene is observed from multiple views, resulting in multi-view capture. %
In this case, DVS reduces to a well-constrained multi-view stereo problem at each time step.
Consider another case where the camera moves significantly faster compared to the scene so that it observes roughly the same scene from different viewpoints. 
As the camera motion approaches the infinity this again reduces the monocular capture to a multi-view setup.
We therefore propose to characterize the amount of multi-view cues by the relative camera-scene motion.

\subsection{Quantifying effective multi-view in a monocular video}
For practicality, we propose two metrics, referred to as effective multi-view factors~(EMFs).
The first metric, full EMF $\Omega$ is defined as the relative ratio between the motion magnitude of the camera to the scene, which in theory characterizes the effective multi-view perfectly, but in practice can be expensive and challenging to compute. %
The second metric, angular EMF $\omega$ is defined as the camera angular velocity around the scene look-at point, which only considers the camera motion; while approximate, it is easy to compute and characterizes object-centric captures well.

\paragraph{Full EMF $\Omega$: ratio of camera-scene motion magnitude.}
\looseness=-1
Consider a monocular video of a moving scene over a set of time steps $\mathcal{T}$.
At each discrete time $t \in \mathcal{T}$, let the camera's 3D location be $\mathbf{o}_t$.
We consider each point $\mathbf{x}_t$ on the domain of observed scene surface~$\mathbb{S}_t^2 \subset \mathbb{R}^3$.
We define the camera-scene motion as the expected relative ratio,
\begin{equation}
    \Omega
    =
    \mathop{\mathbb{E}}_{t,t+1 \in \mathcal{T}}
    \bigg[
    \mathop{\mathbb{E}}_{\mathbf{x}_t \in \mathbb{S}^2_t}
    \Big[
    \frac
    {\|\mathbf{o}_{t+1} - \mathbf{o}_t\|}
    {\|\mathbf{x}_{t+1} - \mathbf{x}_t\|}
    \Big]
    \bigg],
\end{equation}
where the denominator $\mathbf{x}_{t+1} - \mathbf{x}_t$ denotes the 3D scene flow and the the numerator $\mathbf{o}_{t+1}- \mathbf{o}_t$ denotes the 3D camera motion, both over one time step forward. 
The 3D scene flow can be estimated via the 2D dense optical flow field and the metric depth map when available, or monocular depth map from off-the-shelf approaches~\cite{teed2020raft,ranftl2021vision} in the general case.
Please see the Appendix for more details. Note that $\Omega$ in theory captures the effective multi-view factor for any sequence.
However, in practice, 3D scene flow estimation is an actively studied problem and may suffer from noisy or costly predictions.

\paragraph{Angular EMF $\omega$: camera angular velocity.}
We introduce a second metric $\omega$ that is easy to compute in practice.
We make an additional assumption that the capture has a single look-at point in world space, which often holds true, particularly for captures involving a single centered subject.
Specifically, given a look-at point $\mathbf{a}$ by triangulating the optical-axes of all cameras (as per~\cite{park2021nerfies}) and the frame rate $N$, the camera angular velocity $\omega$ is computed as a scaled expectation,
\begin{equation}
    \omega
    =
    \mathop{\mathbb{E}}_{t,t+1 \in \mathcal{T}}
    \bigg[
    \arccos
    \Big(
    \frac
    {
    \langle\mathbf{a} - \mathbf{o}_t, \mathbf{a} - \mathbf{o}_{t+1}\rangle
    }
    {
    \|\mathbf{a} - \mathbf{o}_t\| \cdot \|\mathbf{a} - \mathbf{o}_{t+1}\|
    }
    \Big)
    \bigg]
    \cdot
    N.
\end{equation}
Note that even though $\omega$ only considers the camera motion, it is indicative of effective multi-view in the majority of existing captures, which we describe in Section~\ref{sec:closer_look}. 

For both $\Omega$ and $\omega$, the larger the value, the more multi-view cue the sequence contains. 
For future works introducing new input sequences, we recommend always reporting angular EMF for its simplicity and reporting full EMF when possible.
Next we inspect the existing experimentation practices under the lens of effective multi-view.

\section{Towards better experimentation practice}
\label{sec:existing}

\begin{figure}[t!]
    \begin{minipage}{\linewidth}
        \centering
        \tablestyle{4pt}{1.2}
        \small
        \begin{tabular}{lcccccc}
        \toprule
            &\#Train cam. &Duration &FPS &Depth &Kpt. &Sequences \\
            \midrule
            D-NeRF~\cite{pumarola2020dnerf} &$\texttd150$ &$1-3$\si{s} &$60$ &- &- &$8$ MV \\
            HyperNeRF~\cite{park2021hypernerf}  &$2$ &$8-15$\si{s} &$15$ &- &- &$3$ MV + $13$ SV \\
            Nerfies~\cite{park2021nerfies} &$2$ &$8-15$\si{s} &$5/15$ &- &- &$4$ MV \\
            NSFF~\cite{li2020nsff} &$24$ &$1-3$\si{s} &$15/30$ &Estimated~\cite{ranftl2021vision} &- &$8$ MV \\
            \hline
            iPhone~(proposed) &$1$ &$8-15$\si{s} &$30/60$ &Lidar &\checkmark &$7$ MV + $7$ SV \\
            \bottomrule
        \end{tabular}
        \vspace{-2pt}
        \captionof{table}{
            \textbf{Summary of the existing and proposed iPhone datasets.}
            Existing datasets operate under the effective multi-view regime, teleporting between multiple cameras during training to generate a synthetic monocular video.
            Unlike previous protocols, the proposed iPhone dataset consists of dynamic sequences captured by a \textit{single} smoothly moving camera.
            It also has accompanying depth maps for training supervision and labeled keypoints for evaluation.
            ``MV'' denotes multi-camera capture, and ``SV'' denotes single-camera capture.
        }
        \label{tab:benchmark_summary}
    \end{minipage}
\end{figure}

In this section, we reflect on the existing datasets and find that they operate under the effective multi-view regime, with either teleporting camera motion or quasi-static scene motion.
The reason behind the existing protocol is that monocular DVS is challenging from both the modeling and evaluation perspective.
While the former challenge is well known, the latter is less studied, as we expand below. %
To overcome the existing challenge in the evaluation and enable future research to experiment with casually captured monocular video, we propose a better toolkit, including two new metrics and a new dataset of complex motion in everyday lives.

\subsection{Closer look at existing datasets}
\label{sec:closer_look}

\begin{figure}
    \centering
    \includegraphics[width=\linewidth,bb=9 9 1125 210]{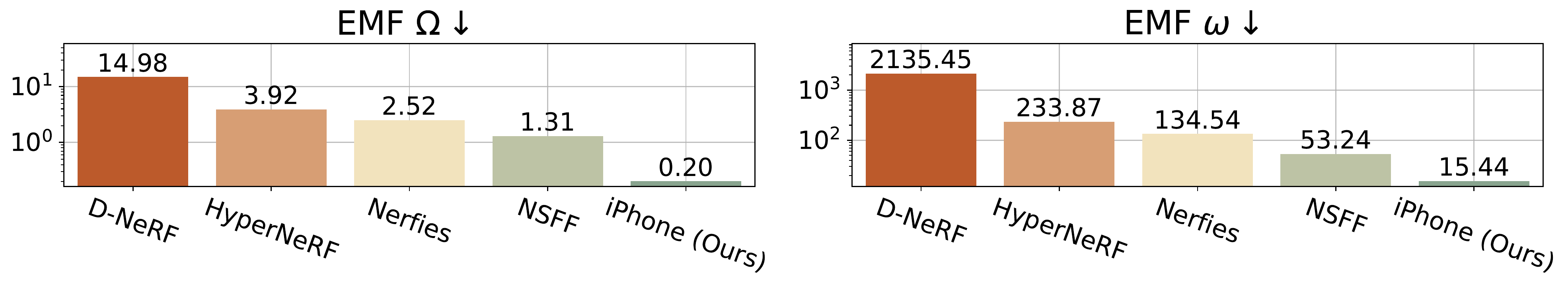}
    \vspace{-12pt}
    \captionof{figure}{
        \textbf{Statistics of effective multi-view factors (EMFs) across different datasets.}
        Existing datasets have high EMFs, indicating the abundance of multi-view cues during training (note that the y-axis is in log scale).
        The proposed iPhone dataset features a single camera, capturing the moving scene with a smooth motion, and thus has smaller EMFs.
        Our results help ground the difficulty of each dataset for DVS from a monocular video.
    }
    \label{fig:benchmark_emfs}
\end{figure}

We investigate the datasets used for evaluation in D-NeRF~\cite{pumarola2020dnerf}, HyperNeRF~\cite{park2021hypernerf}, Nerfies~\cite{park2021nerfies}, and NSFF~\cite{li2020nsff}.
Table~\ref{tab:benchmark_summary} shows their statistics.
We evaluate the amount of effective multi-view cues via the proposed EMFs, shown in Figure~\ref{fig:benchmark_emfs}.
We find that existing datasets have large EMF values on both metrics.
For example, the HyperNeRF dataset has an $\omega$ as large as $\texttd200^{\circ}/\si{s}$.
To put these numbers in context, a person imaging an object $3\si{m}$ away has to move at $1\si[per-mode=symbol]{m\per\second}$ to get an $\omega = 20^{\circ}/\si{s}$ (close to the statistics in the proposed dataset). Some datasets exhibit $\omega$ higher than $120^{\circ}/\si{s}$, which is equivalent to a camera motion faster than the Olympic $100\si{m}$ sprint record, without incurring any motion blur.

\looseness=-1 Visualizing the actual training data shown in Figure~\ref{fig:existing_benchmark_visuals} reveals that existing datasets feature non-practical captures of either (1) teleporting/fast camera motion or (2) quasi-static/slow scene motion.
The former is not representative of practical captures from a hand-held camera, \textit{e.g.}, a smartphone, while the latter is not representative of moving objects in daily life.
Note that, out of the $23$ multi-camera sequences that these prior works used for quantitative evaluation, $22$ have teleporting camera motion, and $1$ has quasi-static scene motion -- the \textsc{Curls} sequence shown at the $5^{\text{th}}$ column in Figure~\ref{fig:existing_benchmark_visuals}.
All $13$ single-camera sequences from HyperNeRF~\cite{park2021hypernerf} used for qualitative evaluation have quasi-static scene motion.

\begin{wrapfigure}{r}{0.45\linewidth}
    \centering
    \vspace{-16pt}
    \includegraphics[width=\linewidth,bb=15 0 421 166]{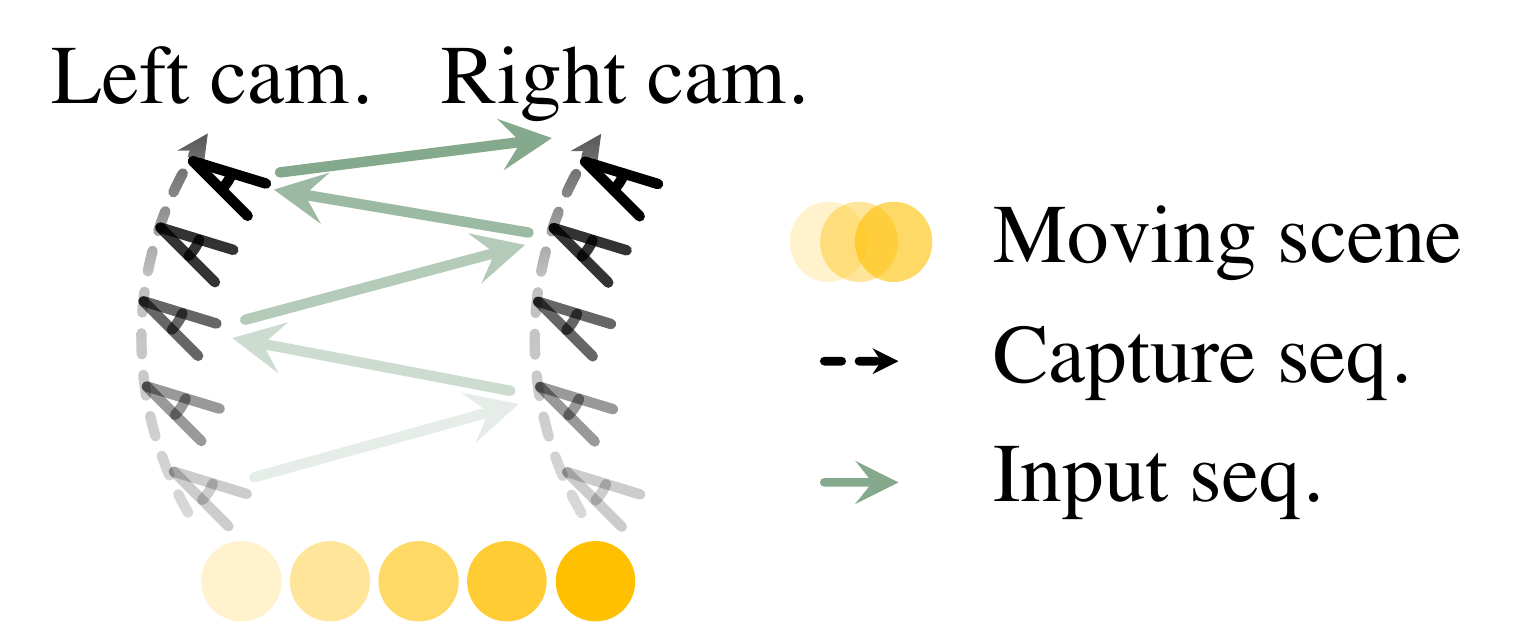}
    \vspace{-12pt}
    \caption{
        \textbf{Camera teleportation from a multi-camera rig.}
        As an example, we show the data protocol used in Nerfies~\cite{park2021nerfies} and HyperNeRF~\cite{park2021hypernerf}.
    }
    \label{fig:teleporting_camera_illustration}
\end{wrapfigure}

The four datasets also share a similar data protocol for generating effective multi-view input sequences from the original multi-camera rig capture.
In Figure~\ref{fig:teleporting_camera_illustration}, we illustrate the teleporting protocol used in Nerfies~\cite{park2021nerfies} and HyperNeRF~\cite{park2021hypernerf} as a canonical example.
They sample alternating frames from two physical cameras (left and right in this case) mounted on a rig to create the training data.
NSFF~\cite{li2020nsff} samples alternating frames from $24$ cameras based on the data released from Yoon~\etal~\cite{yoon2020novel}.
D-NeRF~\cite{pumarola2020dnerf} experiments on synthetic dynamic scenes where cameras are randomly placed on a fixed hemisphere at every time step, in effect teleporting between $100$-$200$ cameras.
We encourage you to visit our \href{\dychecklink}{project page} to view the input videos from these datasets.

\looseness=-1
Existing works adopt effective multi-view capture for two reasons.
First, it makes monocular DVS more tractable.
Second, it enables evaluating novel view on the full image, without worrying about the visibility of each test pixel, as all camera views were visible during training. %
We show this effect in Figure~\ref{fig:impact_of_covisible}.
When trained with camera teleportation ($3^{\text{rd}}$ column), the model can generate a high-quality full image from the test view.
However, when trained without camera teleportation ($4^{\text{th}}$ column), the model struggles to hallucinate unseen pixels since NeRFs~\cite{park2021nerfies,mildenhall2020nerf} are not designed to predict completely unseen portions of the scene, unless they are specifically trained for generalization~\cite{yu2021pixelnerf}.
Next, we propose new metrics that enable evaluation without using camera teleportation. Note that when the model is trained without camera teleportation, the rendering quality also degrades, which we also evaluate. %

\subsection{Our proposed metrics}
\label{sec:proposed_metrics}

While the existing setup allows evaluating on the full rendered image from the test view, the performance under such evaluation protocol, particularly with teleportation, confounds the efficacy of the proposed approaches and the multi-view signal present in the input sequence. %
To evaluate with an actual monocular setup, we propose two new metrics that evaluate only on seen pixels and measure the correspondence accuracy of the predicted deformation.
\providecommand\animage{}
\renewcommand{\animage}[2]{
    \frame{\includegraphics[width=\linewidth,clip,trim=#1]{figures/assets/impact_of_covisible/#2}}
}
\providecommand\textimage{}
\renewcommand{\textimage}[4]{
	\frame{\begin{overpic}[width=\linewidth,clip,trim=#1]{figures/assets/impact_of_covisible/#2}\put(0.3,868){\footnotesize\sethlcolor{black}\textcolor{white}{\hl{$#3/#4$}}}\end{overpic}}
}
\begin{figure}[t!]
    \setlength{\tabcolsep}{0.8pt}
    \renewcommand{\arraystretch}{0.1}
    \begin{tabularx}{\textwidth}{@{}c*{5}{C}@{}}
        \makebox[20pt]{\raisebox{32pt}{\rotatebox[origin=c]{90}{\textsc{3d Printer}}}}\hspace{-4pt} &
        \animage{0 0 0 148}{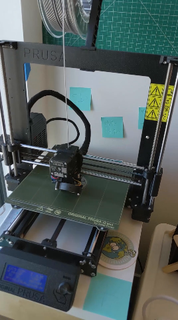} &
        \animage{0 0 0 148}{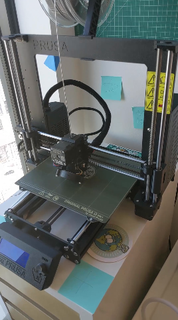} &
        \textimage{0 0 0 148}{hypernerf/vrig-3dprinter/24_intl_nv_pred.png}{3.13}{251.36} &
        \textimage{0 0 0 148}{hypernerf/vrig-3dprinter/24_nv_pred.png}{1.22}{59.40} &
        \animage{0 0 0 148}{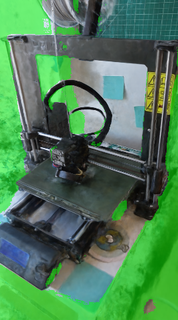}
        \\
        [2pt] &
        {\small Training view} &
        {\small Test view} &
        {\footnotesize \makecell{Nerfies\\(teleporting)~\cite{park2021nerfies}}} &
        {\footnotesize \makecell{Nerfies\\(non-teleporting)}} &
        {\footnotesize \makecell{Nerfies w/ mask\\(non-teleporting)}}
    \end{tabularx}
    \vspace{-4pt}
    \caption{
        \textbf{Evaluation without teleportation with the  co-visibility mask.}
        $\Omega/\omega$ metrics of the input sequence are annotated on the top-left.
        Existing works avoid evaluating on unseen pixels by camera teleportation ($3^{\text{rd}}$ column).
        Naively training with the non-teleporting (smooth) camera trajectory causes evaluation issues on the full image ($4^{\text{th}}$ column), since a NeRF model cannot hallucinate unseen regions.
        We propose to only evaluate seen regions during training by a co-visibility mask (we show non-visible regions in green at the last column).
        Note that the model performs poorly on seen 
        regions as well when trained without camera teleportation.
    }
    \label{fig:impact_of_covisible}
\end{figure}

\paragraph{Co-visibility masked image metrics.}
Existing works evaluate DVS models with image metrics on the \emph{full} image, \eg, PSNR, SSIM~\cite{wang2004image} and LPIPS~\cite{zhang2018unreasonable}, following novel-view synthesis evaluation on static scenes~\cite{mildenhall2020nerf}.
However, in dynamic scenes, particularly for monocular capture with multi-camera validation, the 
test view contains regions that may not have been observed at all by the training camera.
To circumvent this issue without resorting to camera teleportation, for each pixel in the test image, we propose \emph{co-visibility} masking, which tests how many times a test pixel has been observed in the training images.
Specifically, we use optical flow to compute correspondences between every test image and the training images, and only keep test pixels that have enough correspondences in the training images via thresholding. 
This results in a mask, illustrated in Figure~\ref{fig:impact_of_covisible}, which we use to confine the image metrics.
We follow the common practice from the image generation literature and adopt masked metrics, mPSNR and mLPIPS~\cite{gatys2017controlling,huh2020transforming}.
Note that NSFF~\cite{li2020nsff} adopts similar metrics but for evaluating the rendering quality on foreground versus background regions.
We additionally report mSSIM by partial convolution~\cite{liu2018image}, which only considers seen regions during its computation. %
More details are in the Appendix.
Using masked image metrics, we quantify the performance gap in rendering when a model is trained with or without multi-view cues in Section~\ref{sec:reality_check_on_existing}.

\paragraph{Percentage of correctly transferred keypoints~(PCK-T).}
Correspondences lie at the heart of traditional non-rigid reconstruction~\cite{newcombe2015dynamicfusion}, which is overlooked in the current image-based evaluation.
We propose to evaluate 2D correspondences across training frames with the percentage of correctly transferred keypoints~(PCK-T)~\cite{kulkarni2019canonical}, which directly evaluates the quality of the inferred deformation.
Specifically, we sparsely annotate 2D keypoints across input frames to ensure that each keypoint is fully observed during training.
For correspondence readout from existing methods, we use either root finding~\cite{chen2021snarf} or scene flow chaining. Please see the Appendix for details on our keypoint annotation, correspondence readout, and metric computation.
As shown in Figure~\ref{fig:good_pixels_bad_corrs}, evaluating correspondences reveal that high quality image rendering does not necessarily result in accurate correspondences, which indicates issues in the underlying surface, due to the ambiguous nature of the problem.
\providecommand\cbar{}
\renewcommand{\cbar}[1]{
	\includegraphics[height=78pt,bb=-1 0 86 365]{figures/assets/good_pixels_bad_corrs/#1}
}
\providecommand\animage{}
\renewcommand{\animage}[3]{
	\frame{\begin{overpic}[width=\linewidth,clip,trim=#1]{figures/assets/good_pixels_bad_corrs/#2}\put(0.3,33){\footnotesize\sethlcolor{black}\textcolor{white}{\hl{ #3 }}}\end{overpic}}
}
\providecommand\textimage{}
\renewcommand{\textimage}[5]{
	\frame{\begin{overpic}[width=\linewidth,clip,trim=#1]{figures/assets/good_pixels_bad_corrs/#2}\put(0.3,915){\footnotesize\sethlcolor{black}\textcolor{white}{\hl{$#3/#4$}}}\put(0.3,33){\footnotesize\sethlcolor{black}\textcolor{white}{\hl{ #5 }}}\end{overpic}}
}
\begin{figure}[t!]
    \setlength{\tabcolsep}{0.8pt}
    \renewcommand{\arraystretch}{0.1}
    \begin{tabularx}{\textwidth}{@{}c*{6}{C}c@{}}
        \makebox[20pt]{\raisebox{35pt}{\rotatebox[origin=c]{90}{\textsc{Toby Sit}}}}\hspace{-6pt} &
        \textimage{0 68 0 13}{nerfies/toby-sit/60_src.png}{3.30}{160.55}{$t$} &
        \animage{0 68 0 13}{nerfies/toby-sit/60_nv.png}{$t$} &
        \animage{0 68 0 13}{nerfies/toby-sit/60_nv_pred.png}{$t$} &
        \animage{0 68 0 13}{nerfies/toby-sit/60_kpt_src.png}{$t$} &
        \animage{0 68 0 13}{nerfies/toby-sit/60_kpt_dst.png}{$t + 2\si{s}$} &
        \animage{0 68 0 13}{nerfies/toby-sit/60_kpt_dst_pred.png}{$t + 2\si{s}$} &
        \cbar{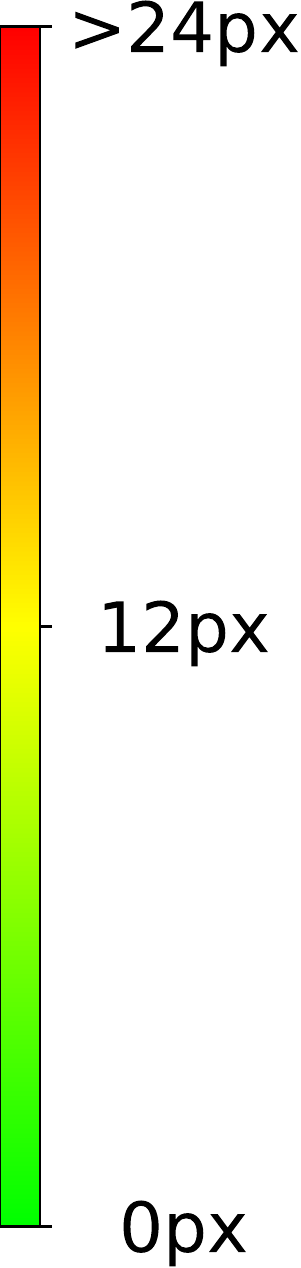}
        \\
        [2pt] &
        {\small Training view} &
        {\small Test view} &
        {\small Prediction~\cite{park2021nerfies}} &
        {\small Source Kpt.} &
        {\small Target Kpt.} &
        {\small Pred. Kpt.~\cite{park2021nerfies}}
    \end{tabularx}
    \vspace{-4pt}
    \caption{
        \textbf{High-quality novel-view synthesis does not imply accurate correspondence modeling.}
        $\Omega/\omega$ metrics of the input sequence are shown on the top-left.
        The time steps of the ground-truth data and predictions are shown on the bottom-left.
        Using Nerfies~\cite{park2021nerfies} as an example, we show that the model renders high-quality results ($3^{\text{rd}}$ column) without modeling accurate correspondences (last column).
        Transferred keypoints are colorized by a heatmap of end-point error, overlaid on the ground-truth target frame.
    }
    \vspace{-1em}
    \label{fig:good_pixels_bad_corrs}
\end{figure}

\subsection{Proposed iPhone dataset}
\label{sec:iphone_dataset}
\begin{figure}[t!]
    \includegraphics[width=\linewidth,bb=0 13 946 329]{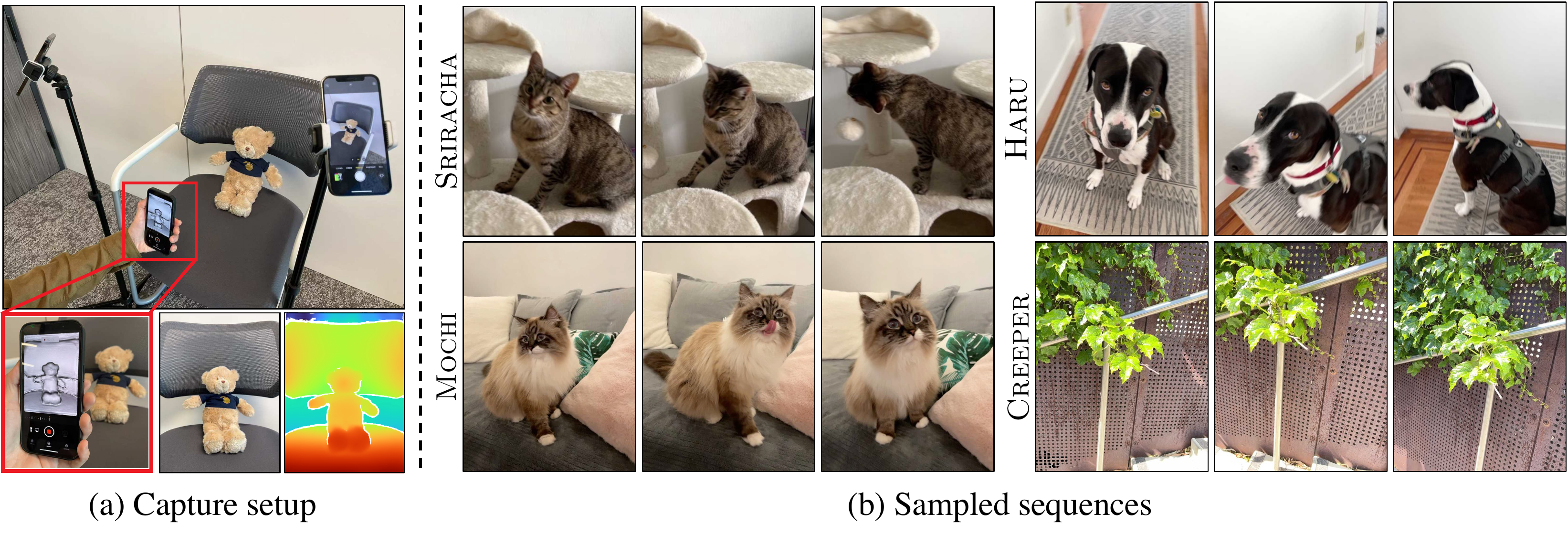}
    \vspace{-15pt}
    \caption{
        \textbf{Capture setup and sampled sequences from our proposed iPhone dataset.}
        The proposed iPhone dataset has a single hand-held camera for training and multiple cameras with a large baseline for validation.
        It has accompanying depth from the iPhone sensor and features diverse and complex real-life motions.
    }
    \vspace{-1em}
    \label{fig:iphone_dataset_visuals}
\end{figure}

Existing datasets can be rectified by removing camera teleportation and evaluated using the proposed metrics, as we do in Section~\ref{sec:reality_check_on_existing}.
However, even after removing camera teleportation, the existing datasets are still not representative of practical in-the-wild capture.
First, the existing datasets are limited in motion diversity.
Second, the evaluation baseline in existing datasets is small, which can hide issues in incorrect deformation and resulting geometry. %
For these reasons, we propose a new dataset called
the {\em iPhone dataset} shown in Figure~\ref{fig:iphone_dataset_visuals}.
In contrast to existing datasets with repetitive object motion, 
we collect $14$ sequences featuring non-repetitive motion, from various categories such as generic objects, humans, and pets.
We deploy three cameras for multi-camera capture -- one hand-held moving camera for training and two static cameras of large baseline for evaluation. Furthermore, our iPhone dataset comes with metric depth from the lidar sensors,
which we use to provide ground-truth depth for supervision.
In Section~\ref{sec:reality_check_on_iphone}, we show that depth supervision, together with other regularizations, is beneficial for training DVS models. Please see the Appendix for details on our multi-camera capture setup, data processing, and more visualizations.  %

\section{Reality check: re-evaluating the state of the art}
\label{sec:reality}

In this section, we conduct a series of empirical studies to disentangle the recent progress in dynamic view synthesis~(DVS) given a monocular video from effective multi-view in the training data.
We evaluate current state-of-the-art methods when the effective multi-view factor~(EMF) is low.

\paragraph{Existing approaches and baselines.}
We consider the following state-of-the-art approaches for our empirical studies: NSFF~\cite{li2020nsff}, Nerfies~\cite{park2021nerfies} and HyperNeRF~\cite{park2021hypernerf}.
We choose them as canonical examples for other approaches~\cite{xian2021space,tretschk2021non,pumarola2020dnerf,gao2021dynamic,wu2022d,li2021neural,du2021neural}, discussed in Section~\ref{sec:related}.
We also evaluate time-conditioned NeRF (T-NeRF) as a common baseline~\cite{xian2021space,pumarola2020dnerf,li2020nsff}. 
Unlike the state-of-the-art methods, it is not possible to extract correspondences from a T-NeRF. 
A summary of these methods can be found in the Appendix.

\paragraph{Datasets.}
We evaluate on the existing datasets as well as the proposed dataset. %
For existing datasets, we use the multi-camera captures accompanying Nerfies~\cite{park2021nerfies} and HyperNeRF~\cite{park2021hypernerf} for evalulation.
Due to their similar capture protocol, we consider them as a single dataset in our experiment (denoted as the Nerfies-HyperNeRF dataset). 
It consists of $7$ sequences in total, which we augment with keypoint annotations.
Our dataset has $7$ multi-camera captures and $7$ single-camera captures.
We evaluate novel-view synthesis on the multi-camera captures and correspondence on all captures.
Our data adopts the data format from the Nerfies-HyperNeRF dataset, with additional support for depth and correspondence labels.
All videos are at $480\si{p}$ resolution and all dynamic scenes are inward-facing.

\paragraph{Masked image and correspondence metrics.}
Following Section~\ref{sec:proposed_metrics}, we evaluate co-visibility masked image metrics and the correspondence metric.
We report masked image metrics: mPSNR, mSSIM~\cite{wang2004image,liu2018image}, and mLPIPS~\cite{li2020nsff,zhang2018unreasonable,gatys2017controlling,huh2020transforming}.
We visualize the rendering results with the co-visibility mask.
For the correspondence metric, we report the percentage of correctly transferred keypoints~(PCK-T)~\cite{kulkarni2019canonical} with threshold ratio $\alpha = 0.05$.
Additional visualizations of full image rendering and inferred correspondences can be found in the Appendix.

\paragraph{Implementation details.}
We consolidate Nerfies~\cite{park2021nerfies} and HyperNeRF~\cite{park2021hypernerf} in one codebase using JAX~\cite{jax2018github}.
Compared to the original official code releases, our implementation aligns all training and evaluation details between models and allows correspondence readout.
Our implementation reproduces the quantitative results in the original papers.
We implement T-NeRF in the same codebase.
For NSFF~\cite{li2020nsff}, we tried both the official code base~\cite{nsff} and a public third-party re-implementation~\cite{nsff_pl}, where the former fails to converge on our proposed iPhone dataset while the latter works well.
We thus report results using the third-party re-implementation.
However, note that both the original and the third-party implementation represent the dynamic scene in normalized device coordinates~(NDC). 
As NDC is designed for forward-facing but not considered inward-facing scenes, layered artifacts may appear due to its log-scale sampling rate in the world space, as shown in Figure~\ref{fig:nerfies_and_hypernerf_benchmark}.
More details about aligning the training procedure and remaining differences are provided in the Appendix.
Code, pretrained models, and data are available on the \href{\dychecklink}{project page}.

\subsection{Reality check on the Nerfies-HyperNeRF dataset}
\label{sec:reality_check_on_existing}

\paragraph{Impact of effective multi-view.}
\providecommand\animage{}
\renewcommand{\animage}[2]{
    \frame{\includegraphics[width=\linewidth,clip,trim=#1]{figures/assets/impact_of_effective_multiview/#2}}
}
\providecommand\textimage{}
\renewcommand{\textimage}[4]{
	\frame{\begin{overpic}[width=\linewidth,clip,trim=#1]{figures/assets/impact_of_effective_multiview/#2}\put(0,622){\footnotesize\sethlcolor{black}\textcolor{white}{\hl{$#3/#4$}}}\end{overpic}}
}
\begin{figure}[t!]
    \centering
    \includegraphics[width=\linewidth,bb=6 0 1345 218]{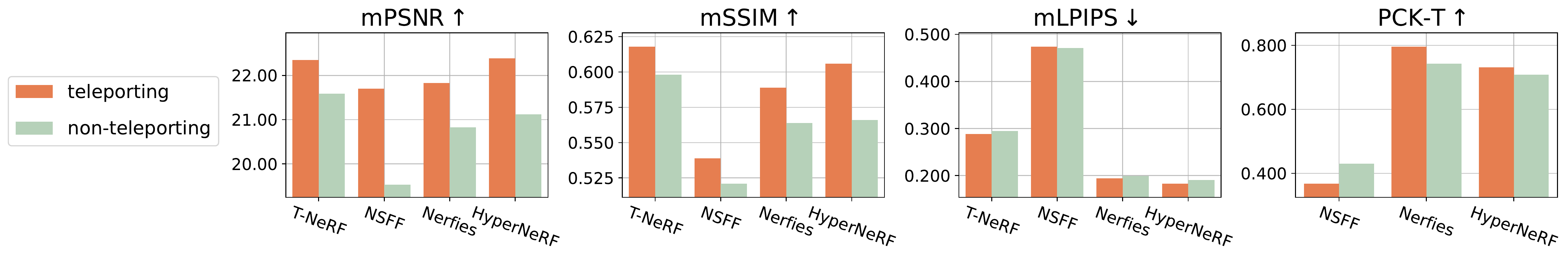}
    \begin{minipage}{\linewidth}
        \setlength{\tabcolsep}{0.8pt}
        \renewcommand{\arraystretch}{0.5}
        \begin{tabularx}{\textwidth}{@{}c*{4}{C}@{}}
            \makebox[20pt]{\raisebox{30pt}{\rotatebox[origin=c]{90}{\textsc{Broom}}}}\hspace{-6pt} &
            \animage{0 0 0 196}{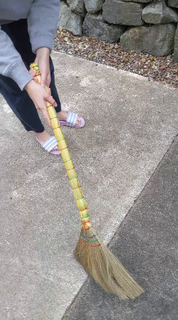} &
            \animage{0 0 0 196}{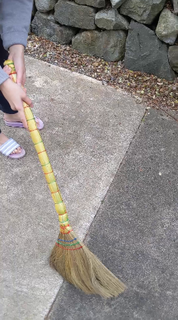} &
            \textimage{0 0 0 196}{nerfies/broom/66_masked_nf_intl.png}{3.40}{128.54} &
            \textimage{0 0 0 196}{nerfies/broom/66_masked_nf_mono.png}{2.57}{60.36}
            \\
            \makebox[20pt]{\raisebox{30pt}{\rotatebox[origin=c]{90}{\textsc{3D Printer}}}}\hspace{-6pt} &
            \animage{0 182 0 14}{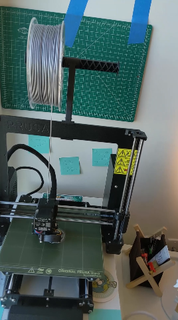} &
            \animage{0 182 0 14}{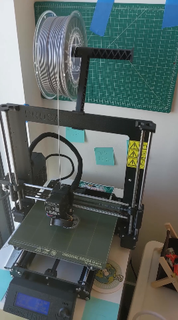} &
            \textimage{0 182 0 14}{hypernerf/vrig-3dprinter/93_masked_nf_intl.png}{3.13}{251.37} &
            \textimage{0 182 0 14}{hypernerf/vrig-3dprinter/93_masked_nf_mono.png}{1.22}{59.40}
            \\
            [2pt] &
            {\small Training view} &
            {\small Test view} &
            {\small Nerfies (teleporting)~\cite{park2021nerfies}} &
            {\small Nerfies (non-teleporting)}
        \end{tabularx}
    \end{minipage}
    \vspace{-4pt}
    \caption{
        \textbf{Impact of effective multi-view on the Nerfies-HyperNeRF dataset.}
        $\Omega/\omega$ metrics of the input sequence are shown on the top-left.
        We compare the existing camera teleporting setting and our non-teleporting setting.
        (Top): Quantitative results of different models using our proposed evaluation metrics.
        (Bottom): Qualitative comparison using Nerfies as an example.
        Two settings use the same set of co-visibility masks computed from common training images.
        Visualizations of other models are in the Appendix.
    }
    \label{fig:impact_of_effective_multiview}
\end{figure}

\providecommand\animage{}
\renewcommand{\animage}[2]{
    \frame{\includegraphics[width=\linewidth,clip,trim=#1]{figures/assets/nerfies_and_hypernerf_benchmark/#2}}
}
\providecommand\textimage{}
\renewcommand{\textimage}[4]{
	\frame{\begin{overpic}[width=\linewidth,clip,trim=#1]{figures/assets/nerfies_and_hypernerf_benchmark/#2}\put(0.3,920){\footnotesize\sethlcolor{black}\textcolor{white}{\hl{$#3/#4$}}}\end{overpic}}
}
\begin{figure}[t!]
    \setlength{\tabcolsep}{0.8pt}
    \renewcommand{\arraystretch}{0.5}
    \begin{tabularx}{\textwidth}{@{}c*{6}{C}@{}}
        \makebox[20pt]{\raisebox{40pt}{\rotatebox[origin=c]{90}{\textsc{Chicken}}}}\hspace{-6pt} &
        \textimage{0 68 0 13}{hypernerf/vrig-chicken/0_src.png}{1.52}{33.46} &
        \animage{0 68 0 13}{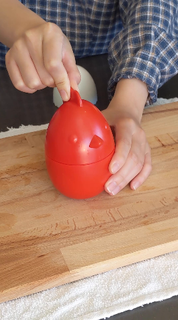} &
        \animage{0 68 0 13}{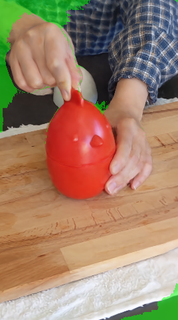} &
        \animage{0 68 0 13}{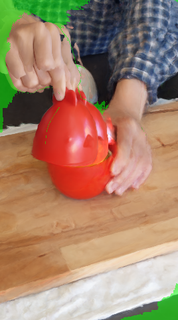} &
        \animage{0 68 0 13}{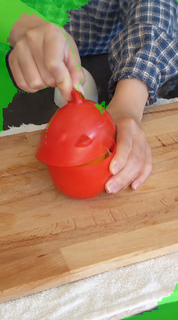} &
        \animage{0 68 0 13}{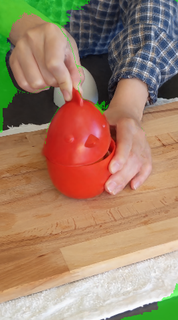}
        \\
        \makebox[13pt]{\raisebox{40pt}{\rotatebox[origin=c]{90}{\textsc{Peel Banana}}}}\hspace{-6pt} &
        \textimage{0 68 0 13}{hypernerf/vrig-peel-banana/443_src.png}{0.33}{33.77} &
        \animage{0 68 0 13}{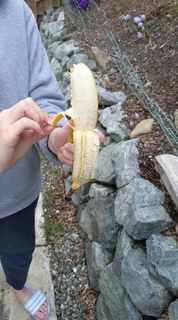} &
        \animage{0 68 0 13}{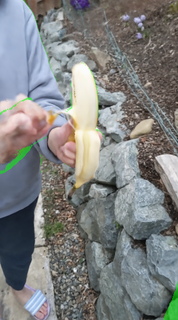} &
        \animage{0 68 0 13}{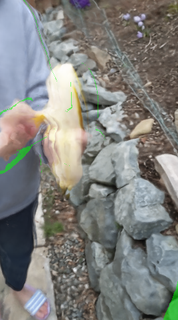} &
        \animage{0 68 0 13}{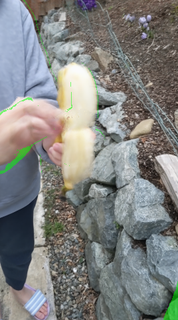} &
        \animage{0 68 0 13}{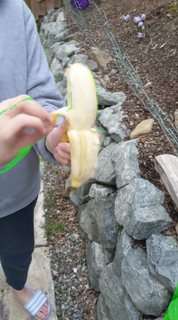}
        \\
        [2pt] &
        {\small Training view} &
        {\small Test view} &
        {\small T-NeRF} &
        {\small NSFF~\cite{li2021neural}} &
        {\small Nerfies~\cite{park2021nerfies}} &
        {\small HyperNeRF~\cite{park2021hypernerf}}
    \end{tabularx}
    \vspace{-4pt}
    \caption{
        \textbf{Qualitative results on the Nerfies-HyperNeRF dataset without camera teleportation.}
        $\Omega/\omega$ metrics of the input sequence are shown on the top-left.
        Existing approaches struggle at modeling dynamic regions.
    }
    \vspace{-1em}
    \label{fig:nerfies_and_hypernerf_benchmark}
\end{figure}

We first study the impact of effective multi-view on the Nerfies-HyperNeRF~\cite{park2021nerfies,park2021hypernerf} dataset.
In this experiment, we rectify the effective multi-view sequences by only using the left camera during training as opposed to both the left and right cameras, illustrated in Figure~\ref{fig:teleporting_camera_illustration}.
We denote the original setting as ``teleporting'' and the rectified sequences as ``non-teleporting''.
We train all approaches under these two settings with the same held-out validation frames and same set of co-visibility masks computed from common training frames. 
In Figure~\ref{fig:impact_of_effective_multiview} (Top), all methods perform better across all metrics when trained under the teleporting setting compared to the non-teleporting one, with the exception of PCK-T for NSFF. 
We conjecture that this is because that NSFF has additional optical flow supervision, which is more accurate without camera teleportation.
In Figure~\ref{fig:impact_of_effective_multiview} (Bottom), we show qualitative results using Nerfies (we include visualizations of the other methods in the Appendix).
Without effective multi-view, Nerfies fails at modeling physically plausible shape for broom and wires.
Our results show that the effective multi-view in the existing experimental protocol inflates the synthesis quality of prior methods, and that truly monocular captures are more challenging.

\paragraph{Benchmark results without camera teleportation.}

\begin{table}
\centering
\makebox[0pt][c]{\parbox{\textwidth}{%
    \begin{minipage}[b]{0.49\hsize}\centering
        \begin{table}[H]
    \centering
    \tablestyle{4pt}{1.2}
    \scriptsize
    \begin{tabular}{lccccc}
    \toprule
        {\makecell[l]{$\Omega=1.30$\\$\omega=51.53$}}\textbf{}
        &mPSNR$\uparrow$ &mSSIM$\uparrow$ &mLPIPS$\downarrow$ &PCK-T$\uparrow$ \\
        \midrule
        T-NeRF &$\mathbf{21.55}$ &$\mathbf{0.595}$ &$0.297$ &- \\
        \hline
        NSFF~\cite{li2020nsff} &$19.53$ &$0.521$ &$0.471$ &$0.422$ \\
        Nerfies~\cite{park2021nerfies} &$20.85$ &$0.562$ &$0.200$ &$0.756$ \\
        HyperNeRF~\cite{park2021hypernerf} &$21.16$ &$0.565$ &$\mathbf{0.192}$ &$\mathbf{0.764}$ \\
        \bottomrule
    \end{tabular}
    \vspace{.5em}
    \caption{
        \textbf{Benchmark results on the rectified Nerfies-HyperNeRF dataset.}
        Please see the Appendix for the breakdown over $7$ multi-camera sequences.
    }
   \vspace{-1em}
    \label{tab:nerfies_and_hypernerf_benchmark}
\end{table}

    \end{minipage}
    \hfill
    \begin{minipage}[b]{0.49\hsize}\centering
        \begin{table}[H]
    \centering
    \tablestyle{4pt}{1.2}
    \scriptsize
    \begin{tabular}{lccccc}
    \toprule
        {\makecell[l]{$\Omega=0.24$\\$\omega=15.18$}}
        &mPSNR$\uparrow$ &mSSIM$\uparrow$ &mLPIPS$\downarrow$ &PCK-T$\uparrow$ \\
        \midrule
        T-NeRF &$\mathbf{16.96}$ &$\mathbf{0.577}$ &$0.379$ &- \\
        \hline
        NSFF~\cite{li2020nsff} &$15.46$ &$0.551$ &$0.396$ &$0.256$ \\
        Nerfies~\cite{park2021nerfies} &$16.45$ &$0.570$ &$0.339$ &$\mathbf{0.453}$ \\
        HyperNeRF~\cite{park2021hypernerf} &$16.81$ &$0.569$ &$\mathbf{0.332}$ &$0.400$ \\
        \bottomrule
    \end{tabular}
    \vspace{.5em}
    \caption{
        \textbf{Benchmark results on the proposed iPhone dataset.} 
        Please see the Appendix for the breakdown over $7$ multi-camera sequences of complex motion.
    }
    \vspace{-1em}
    \label{tab:iphone_benchmark}
\end{table}

    \end{minipage}
}}
\end{table}

In Table~\ref{tab:nerfies_and_hypernerf_benchmark} and Figure~\ref{fig:nerfies_and_hypernerf_benchmark}, we report the quantitative and qualitative results under the non-teleporting setting.
Note that our implementation of the T-NeRF baseline performs the best among all four evaluated models in terms of mPSNR and mSSIM.
In Figure~\ref{fig:nerfies_and_hypernerf_benchmark}, we confirm this result since T-NeRF renders high-quality novel view for both sequences.
HyperNeRF produces the most photorealistic renderings, measured by mLPIPS.
However it also produces distorted artifacts that do not align well with the ground truth (\textit{e.g.}, the incorrect shape in the \textsc{Chicken} sequence).

\subsection{Reality check on the proposed iPhone dataset}
\label{sec:reality_check_on_iphone}
\paragraph{Ablation study on improving the state of the art.}
\providecommand\animage{}
\renewcommand{\animage}[2]{
    \frame{\includegraphics[width=\linewidth,clip,trim=#1]{figures/assets/ablations/#2}}
}
\providecommand\textimage{}
\renewcommand{\textimage}[4]{
	\frame{\begin{overpic}[width=\linewidth,clip,trim=#1]{figures/assets/ablations/#2}\put(0.3,918){\footnotesize\sethlcolor{black}\textcolor{white}{\hl{$#3/#4$}}}\end{overpic}}
}
\begin{figure}[t!]
\vspace{-1em}
    \centering
    \includegraphics[width=\linewidth,bb=6 0 1297 230]{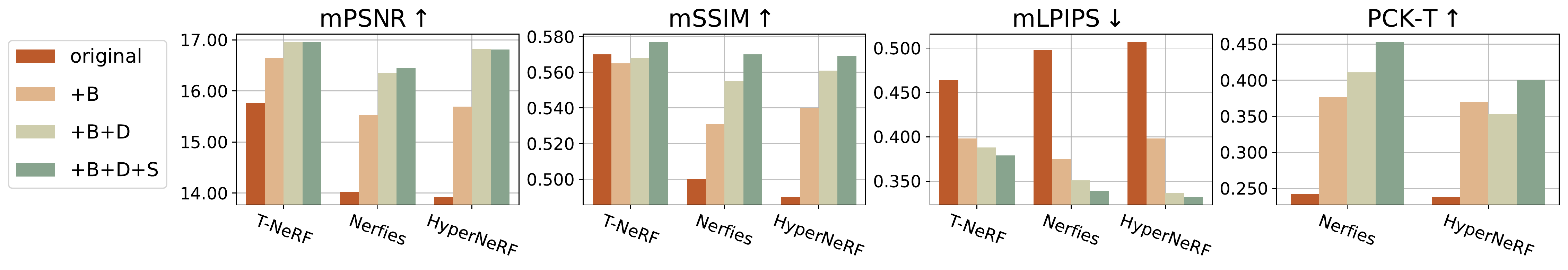}
    \begin{minipage}{\linewidth}
        \setlength{\tabcolsep}{0.8pt}
        \renewcommand{\arraystretch}{0.5}
        \begin{tabularx}{\textwidth}{@{}c*{6}{C}@{}}
            \makebox[20pt]{\raisebox{40pt}{\rotatebox[origin=c]{90}{\textsc{Teddy}}}}\hspace{-6pt} &
            \textimage{0 0 0 0}{iphone2/teddy/311_src.png}{0.20}{7.62} &
            \animage{0 0 0 0}{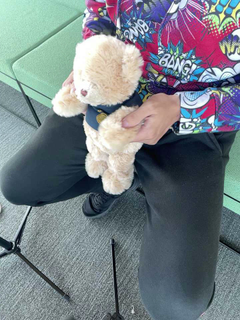} &
            \animage{0 0 0 0}{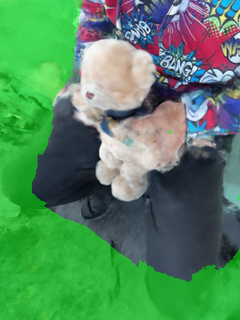} &
            \animage{0 0 0 0}{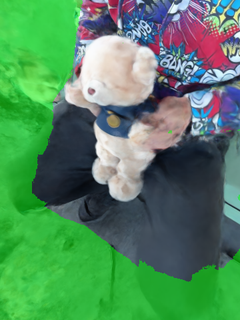} &
            \animage{0 0 0 0}{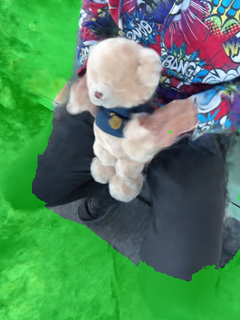} &
            \animage{0 0 0 0}{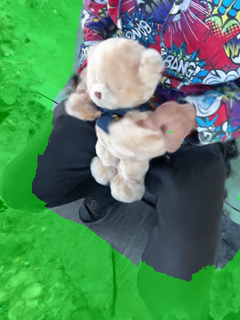}
            \\
            \makebox[20pt]{\raisebox{40pt}{\rotatebox[origin=c]{90}{\textsc{Paper Windmill}}}}\hspace{-6pt} &
            \textimage{0 0 0 0}{iphone2/paper-windmill/391_src.png}{0.38}{10.71} &
            \animage{0 0 0 0}{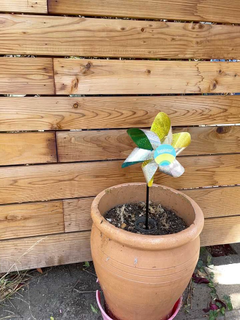} &
            \animage{0 0 0 0}{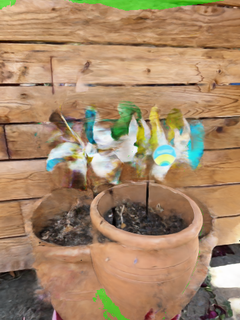} &
            \animage{0 0 0 0}{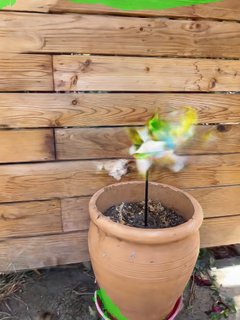} &
            \animage{0 0 0 0}{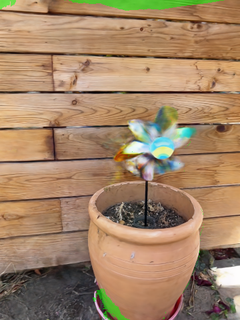} &
            \animage{0 0 0 0}{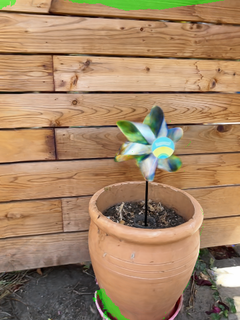}
            \\
            [2pt] &
            {\small Training view} &
            {\small Test view} &
            {\small HyperNeRF~\cite{park2021hypernerf}} &
            {\small \textsc{+B}} &
            {\small \textsc{+B+D}} &
            {\small \textsc{+B+D+S}}
        \end{tabularx}
    \end{minipage}
    \vspace{-4pt}
    \caption{
        \textbf{Ablation study on improving the state of the art on the proposed iPhone dataset.}
        $\Omega/\omega$ metrics of the input sequence are shown on the top-left.
        \textsc{+B}, \textsc{+D}, \textsc{+S} denotes random background compositing~\cite{weng2022humannerf}, additional metric depth supervision~\cite{xian2021space,li2020nsff} from iPhone sensor, and surface sparsity regularizer~\cite{barron2021mip}, respectively.
    }
    \vspace{-1em}
    \label{fig:ablations}
\end{figure}

We find that existing methods perform poorly out-of-the-box on the proposed iPhone dataset with more diverse and complex real-life motions.
In Figure~\ref{fig:ablations} (Bottom), we demonstrate this finding with HyperNeRF~\cite{park2021hypernerf} for it achieves the highest mLPIPS metric on the Nerfies-HyperNeRF dataset.
Shown in the $3^{\text{rd}}$ column, HyperNeRF produces visually implausible results with ghosting effects. %
Thus we explored incorporating additional regularizations from recent advances in neural rendering.
Concretely, we consider the following: (\textsc{+B}) random background compositing~\cite{weng2022humannerf}; (\textsc{+D}) a depth loss on the ray matching distance~\cite{xian2021space,li2020nsff}; and (\textsc{+S}) a sparsity regularization for scene surface~\cite{barron2021mip}.
In Figure~\ref{fig:ablations} (Top), we show quantitative results from the ablation.
In Figure~\ref{fig:ablations} (Bottom), we show visualizations of the impact of each regularization.
Adding additional regularizations consistently boosts model performance.
While we find the random background compositing regularizations particularly helpful, extra depth supervision and surface regularization further improve the quality, \textit{e.g.}, the fan region of the paper windmill.

\paragraph{Benchmarked results.}
\providecommand\animage{}
\renewcommand{\animage}[2]{
    \frame{\includegraphics[width=\linewidth,clip,trim=#1]{figures/assets/iphone_benchmark/#2}}
}
\providecommand\textimage{}
\renewcommand{\textimage}[4]{
	\frame{\begin{overpic}[width=\linewidth,clip,trim=#1]{figures/assets/iphone_benchmark/#2}\put(0.3,905){\footnotesize\sethlcolor{black}\textcolor{white}{\hl{$#3/#4$}}}\end{overpic}}
}
\begin{figure}[t!]
    \setlength{\tabcolsep}{0.8pt}
    \renewcommand{\arraystretch}{0.5}
    \begin{tabularx}{\textwidth}{@{}c*{7}{C}@{}}
        \makebox[20pt]{\raisebox{35pt}{\rotatebox[origin=c]{90}{\textsc{Apple}}}}\hspace{-6pt} &
        \textimage{0 0 0 0}{iphone2/apple/185_src.png}{0.75}{3.79} &
        \animage{0 0 0 0}{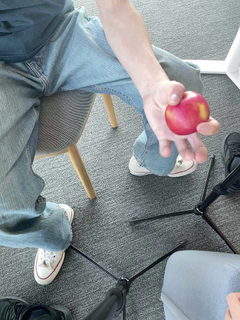} &
        \animage{0 0 0 0}{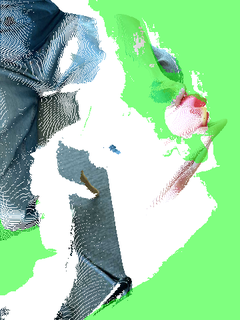} &
        \animage{0 0 0 0}{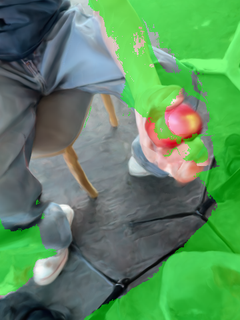} &
        \animage{0 0 0 0}{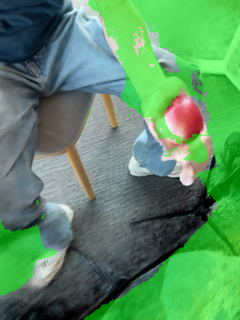} &
        \animage{0 0 0 0}{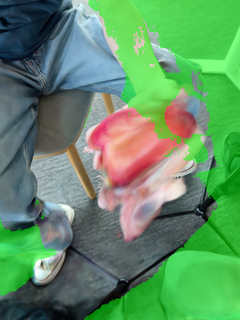} &
        \animage{0 0 0 0}{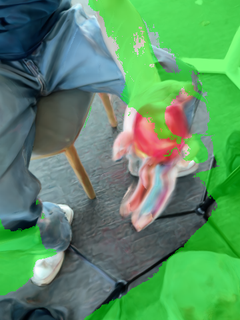}
        \\
        \makebox[20pt]{\raisebox{35pt}{\rotatebox[origin=c]{90}{\textsc{Space Out}}}}\hspace{-6pt} &
        \textimage{10 8 10 8}{iphone2/space-out/425_src.png}{0.13}{6.37} &
        \animage{10 8 10 8}{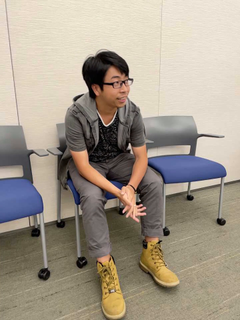} &
        \animage{10 8 10 8}{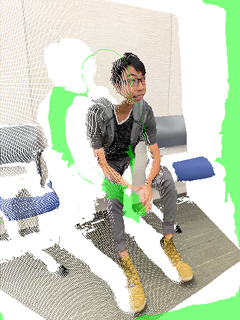} &
        \animage{10 8 10 8}{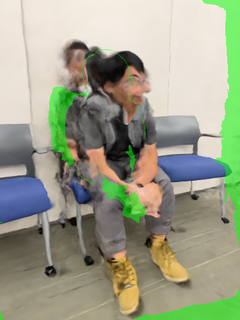} &
        \animage{10 8 10 8}{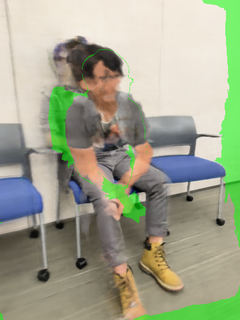} &
        \animage{10 8 10 8}{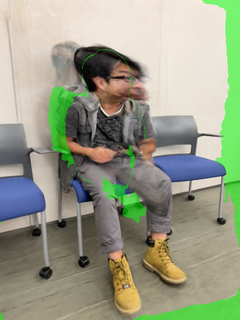} &
        \animage{10 8 10 8}{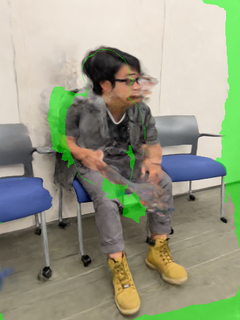}
        \\
        [2pt] &
        {\small Training view} &
        {\small Test view} &
        {\small Depth re-proj.} &
        {\small T-NeRF++} &
        {\small NSFF~\cite{li2020nsff}} &
        {\small Nerfies++} &
        {\small HyperNeRF++}
    \end{tabularx}
    \vspace{-4pt}
    \caption{
        \textbf{Qualitative results on the proposed iPhone dataset.}
        $\Omega/\omega$ metrics of the input sequence are shown on the top-left.
        The models shown here are trained with all the additional regularizations (\textsc{+B+D+S}) except NSFF.
        However, existing approaches still struggle to produce high-quality results.
    }
    \vspace{-1em}
    \label{fig:iphone_benchmark}

\end{figure}

In Figure~\ref{fig:iphone_benchmark}, we show qualitative results from our benchmark using the best model settings from the ablation study, denoted as ``++''.
Note that it is non-trivial to apply the same enhancements to NSFF for its NDC formulation so we keep it as-is.
We visualize the lidar depth re-projection from the training view ($1^{\text{st}}$ column) to the test view ($2^{\text{nd}}$ column), as a reference for qualitative comparison ($3^{\text{rd}}$ column).
Note that the white region is occluded from the input view, whereas the green region is occluded from the most of input video frames.
We observe that existing approaches do not handle complex deformation well.
For example, all models fail at fusing a valid human shape on the \textsc{Space Out} sequence.
In Table~\ref{tab:iphone_benchmark}, we find a similar trend as in the Nerfies-HyperNeRF dataset: the baseline T-NeRF performs the best in terms of mPSNR and mSSIM while HyperNeRF produces the most photorealistic renderings in terms of mLPIPS.
The overall synthesis quality and correspondence accuracy of all methods drop considerably compared to the results on the Nerfies-HyperNeRF dataset.
Taking Nerfies as an example, it drops $\SI{4.4}{dB}$ in mPSNR, $69.6\%$ in mLPIPS, and $40.1\%$ in PCK-T.
Our study suggests an opportunity for large improvement when modeling complex motion.

\section{Discussion and recommendation for future works}
\label{sec:recommendations}
In this work, we expose issues in the common practice and establish systematic means to calibrate performance metrics of existing and future works, in the spirit of papers like~\cite{divvala2009empirical,tatarchenko2019single,musgrave2020metric}. We provide initial attempts toward characterizing the difficulty of a monocular video for dynamic view synthesis~(DVS) in terms of effective multi-view factors~(EMFs).
In practice, there are other challenging factors such as variable appearance, lighting condition, motion complexity and more. We leave their characterization for future works.
We recommend future works to visualize the input sequences and report EMFs when demonstrating the results. We also recommend future works to evaluate the correspondence accuracy and strive for establishing better correspondences for DVS.

\paragraph{Acknowledgements.}
We would like to thank Zhengqi Li and Keunhong Park for valuable feedback and discussions; Matthew Tancik and Ethan Weber for proofreading.
We are also grateful to our pets: Sriracha, Haru, and Mochi, for being good during capture.
This project is generously supported in part by the CONIX Research Center, sponsored by DARPA, as well as the BDD and BAIR sponsors.

{\small
\bibliographystyle{unsrt}
\bibliography{egbib}
}

\newpage
\appendix
\section{Outline}
In this Appendix, we describe in detail the following:
\begin{itemize}
    \setlength\itemsep{-1pt}
    \item Computation for effective multi-view factors~(EMFs) in Section~\ref{sec:app_comp_emf}.
    \item Computation for co-visibility mask and masked image metrics in Section~\ref{sec:app_comp_covisible}.
    \item Summary of existing works and correspondence readout in Section~\ref{sec:app_comp_corr}.
    \item Summary of the capture setup and data processing for our
iPhone dataset in Section~\ref{sec:app_summ_iphone}.
    \item Summary of the implementation details and remain differences in Section~\ref{sec:app_summ_impl}.
    \item Additional results on the impact of effective multi-view in Section~\ref{sec:app_emv_results}.
    \item Additional results on per-sequence performance breakdown in Section~\ref{sec:app_breakdown_results}.
    \item Additional results on novel-view synthesis in Section~\ref{sec:app_nvs_results}.
    \item Additional results on inferred correspondence in Section~\ref{sec:app_corr_results}.
\end{itemize}
For better demonstration, we strongly recommend visiting our \href{\dychecklink}{project page} for videos of the capture and result visualizations.

\section{Computation for effective multi-view factors~(EMFs)}
\label{sec:app_comp_emf}

To quantify the amount of effective multi-view in a sequence by the camera and scene motion magnitude, we propose two metrics as effective multi-view factors (EMFs), \ie, the Full EMF $\Omega$ and the angular EMF $\omega$.
Note that we design our metrics to be scale-agnostic such that we can compare them across different sequences of different world scales.

As in the main paper, we define a point $\mathbf{x}_t \in \mathbb{S}_t^2$ on the visible object's surface and a camera parameterized by its origin $\mathbf{o}_t$ at time $t \in \mathcal{T}$, where $\mathcal{T}$ is the set of possible time steps.

\subsection{Full EMF $\Omega$: Ratio of camera-scene motion magnitude}
\label{sec:app_comp_emf_Omega}
We are interested in the relative scale of the camera motion compared to the object.
Recall that we define $\Omega$ as the expected ratio over all visible pixels over time,
\begin{equation}
    \Omega
    =
    \mathop{\mathbb{E}}_{t,t+1 \in \mathcal{T}}
    \bigg[
    \mathop{\mathbb{E}}_{\mathbf{x}_t \in \mathbb{S}^2_t}
    \Big[
    \frac
    {\|\mathbf{o}_{t+1} - \mathbf{o}_t\|}
    {\|\mathbf{x}_{t+1} - \mathbf{x}_t\|}
    \Big]
    \bigg].
\end{equation}
The numerator is trivially computable given the camera information.
We thus focus on the denominator, \textit{i.e.}, the foreground 3D scene flow $\mathbf{x}_{t+1} - \mathbf{x}_t$.

We estimate 3D scene flow by combining the known cameras, dense 2D optical flow, and per-frame depth maps.
We estimate the 2D optical flow using RAFT~\cite{teed2020raft}.
When metric depth is not available, \textit{e.g.}, on previous datasets~\cite{li2020nsff,park2021hypernerf,park2021nerfies}, we use DPT~\cite{ranftl2021vision} for monocular depth estimation.
Additionally, we need a foreground mask for the object, which we obtain through a video segmentation network~\cite{botach2021end}.
For each pixel location $\mathbf{u}_t$ at time $t$ in the foreground mask, we can compute its 3D position $\mathbf{x}_t$ by back-projection with the depth $z_t$.
We then get the 2D pixel correspondence at time $t+1$ by simply following the 2D optical flow $\mathbf{u}_{t+1} = \mathbf{u}_t + \mathbf{f}_{t \rightarrow t+1}(\mathbf{u}_t)$, where $\mathbf{f}_{t\rightarrow t+1}$ is a bilinearly interpolated forward flow map.
After back-projection, we obtain the corresponding 3D point position $\mathbf{x}_{t+1}$ at frame $t+1$.
In practice, extra care is needed for handling the unknown depth scale from model prediction and occlusion, discussed next.

\paragraph{Aligning depth maps of unknown scales.}
The DPT~\cite{ranftl2021vision} model predicts a disparity map in Euclidean space with an unknown scale $a$ and shift $b$.
To resolve the scale and shift ambiguity in the predicted disparity maps, we make use of the sparse 3D points extracted by COLMAP.
For a frame at time $t$, we first calculate the actual disparity $1 / \tilde{z}_t$ of the sparse 3D points by projecting them onto the image, which usually results in sub-pixels.
We then bilinearly interpolate the predicted disparity map to get the predicted disparity $1 / \tilde{z}^*_t$.
Scale $a$ and shift $b$ can then be estimated through linear regression via the relation:
\begin{equation*}
    \frac{1}{\tilde{z}_t}  = a \cdot \frac{1}{\tilde{z}^*_t} + b.
\end{equation*}
Note that the sparse 3D points from COLMAP~\cite{schoenberger2016sfm} are all located on the static background.
When projecting the sparse 3D points onto the image, some points might be occluded by the moving objects in the foreground.
We handle occluded points by fitting $a$ and $b$ using RANSAC~\cite{fischler1981random}, which ignores outliers and is robust in practice.

\looseness=-1
\paragraph{Handling occlusions.}
We identify occlusions using a forward-backward consistency check following the method of Brox \etal~\cite{brox2004high}. 
We briefly summarize their method here.

Concretely, we identify an occlusion by chaining the forward flow $\mathbf{f}_{t \rightarrow t+1}$ and backward flow $\mathbf{f}_{t+1 \rightarrow t}$ and thresholding based on warp consistency.
For those pixels that have inconsistent forward and backward optical flows, defined by regions where chained forward and backward flows result in non-zero flow values, satisfying the following inequality:
\begin{equation}
    \begin{aligned}
    \|\mathbf{f}_{t \rightarrow t+1}(\mathbf{u}_t) +
    \mathbf{f}_{t+1\rightarrow t}(&\mathbf{u}_t + \mathbf{f}_{t \rightarrow t+1}(\mathbf{u}_t))\|_2^2
    \\
    &\geqslant
    0.01 \cdot (\|\mathbf{f}_{t \rightarrow t+1}(\mathbf{u}_t)\|_2^2 +
    \|\mathbf{f}_{t+1\rightarrow t}(\mathbf{u}_t + \mathbf{f}_{t \rightarrow t+1}(\mathbf{u}_t))\|_2^2)
    + 0.5.
    \end{aligned}
    \label{eq:forward_backward_check}
\end{equation}
The occluded pixels, along with the background pixels not belonging to the foreground mask, are excluded from the 3D scene flow computation.

\paragraph{Discussion.}
In practice, we find that the $\Omega$ metric relies on the model estimation quality, in particular, the monocular depth prediction.
We therefore design a second metric by measuring camera angular speed $\omega$.
With some practical assumptions, it circumvents $\Omega$'s limitation and does not rely on any external model estimates.

\setlength{\tabcolsep}{4pt}
\begin{table}
\tablestyle{4pt}{1.2}
\begin{center}
\resizebox{0.75\textwidth}{!}{%
\begin{tabular}{lllccc}\toprule
Dataset &Sequence &\#Frames &FPS &$\Omega$ &$\omega$ \\
\midrule
D-NeRF~\cite{pumarola2020dnerf} 
&\textsc{Bouncingballs} &$150$ &$30$ &$15.52$ &$1945.52$ \\
&\textsc{Hellwarrior} &$100$ &$30$ &$10.32$ &$2984.15$ \\
&\textsc{Hook} &$100$ &$30$ &$25.82$ &$1996.95$ \\
&\textsc{Jumpingjacks} &$200$ &$30$ &$10.64$ &$1969.47$ \\
&\textsc{Lego} &$50$ &$30$ &$17.00$ &$2133.78$ \\
&\textsc{Mutant} &$150$ &$30$ &$12.64$ &$1908.67$ \\
&\textsc{Standup} &$150$ &$30$ &$14.22$ &$2011.31$ \\
&\textsc{Trex} &$200$ &$30$ &$13.70$ &$2133.78$ \\
\midrule
HyperNeRF~\cite{park2021hypernerf} 
&\textsc{3D Printer} &$207$ &$15$ &$3.13$ &$251.37$ \\
&\textsc{Chicken} &$164$ &$15$ &$7.38$ &$212.58$ \\
&\textsc{Peel Banana} &$513$ &$15$ &$1.26$ &$237.66$ \\
\midrule
Nerfies~\cite{park2021nerfies} 
&\textsc{Broom} &$197$ &$15$ &$3.40$ &$128.54$ \\
&\textsc{Curls} &$57$ &$5$ &$1.20$ &$138.55$ \\
&\textsc{Tail} &$238$ &$15$ &$3.30$ &$160.55$ \\
&\textsc{Toby Sit} &$308$ &$15$ &$2.18$ &$110.51$ \\
\midrule
NSFF~\cite{li2020nsff} 
&\textsc{Balloon1} &$24$ &$15$ &$2.44$ &$57.63$ \\
&\textsc{Balloon2} &$24$ &$30$ &$0.76$ &$76.97$ \\
&\textsc{Dynamic Face} &$24$ &$15$ &$4.57$ &$83.17$ \\
&\textsc{Jumping} &$24$ &$30$ &$0.68$ &$53.79$ \\
&\textsc{Playground} &$24$ &$30$ &$0.36$ &$71.56$ \\
&\textsc{Skating} &$24$ &$30$ &$0.79$ &$42.76$ \\
&\textsc{Truck} &$24$ &$30$ &$0.22$ &$19.41$ \\
&\textsc{Umbrella} &$24$ &$15$ &$0.62$ &$20.66$ \\
\midrule
iPhone (Ours) 
&\textsc{Apple} &$475$ &$30$ &$0.75$ &$3.79$ \\
&\textsc{Backpack} &$180$ &$30$ &$0.26$ &$5.59$ \\
&\textsc{Block} &$350$ &$30$ &$0.04$ &$11.50$ \\
&\textsc{Creeper} &$210$ &$30$ &$0.23$ &$14.05$ \\
&\textsc{Handwavy} &$303$ &$30$ &$0.05$ &$13.66$ \\
&\textsc{Haru} &$200$ &$60$ &$0.30$ &$30.32$ \\
&\textsc{Mochi} &$180$ &$60$ &$0.07$ &$14.49$ \\
&\textsc{Paper Windmill} &$277$ &$30$ &$0.38$ &$10.71$ \\
&\textsc{Pillow} &$330$ &$30$ &$0.06$ &$13.19$ \\
&\textsc{Space Out} &$429$ &$30$ &$0.13$ &$6.37$ \\
&\textsc{Spin} &$426$ &$30$ &$0.15$ &$7.86$ \\
&\textsc{Sriracha} &$220$ &$30$ &$0.18$ &$18.56$ \\
&\textsc{Teddy} &$350$ &$30$ &$0.20$ &$7.62$ \\
&\textsc{Wheel} &$250$ &$30$ &$0.03$ &$58.45$ \\
\bottomrule
\end{tabular}
}
\end{center}
\caption{
\textbf{Per-sequence breakdowns of the statistics of different datasets.} 
As in the main paper, we consider the multi-camera captures from three representative existing datasets: D-NeRF~\cite{pumarola2020dnerf}, Nerfies~\cite{park2021nerfies} and HyperNeRF~\cite{park2021hypernerf}.
We also provide per-sequence breakdowns for both the multi-camera and single-camera captures from our proposed iPhone dataset.
}
\vspace*{-1.5em}

\label{tab:data_stats_breakdown}
\end{table}
\setlength{\tabcolsep}{1.4pt}
\subsection{Angular EMF $\omega$: Camera angular velocity}
We propose to measure camera angular speed $\omega$ given the camera parameters, frame rate $N$ and a single 3D look-at point $\mathbf{a}$ obtained by triangulating all cameras, following Nerfies~\cite{park2021nerfies}.
Recall that $\omega$ is computed as the scaled expectation,
\begin{equation}
    \omega
    =
    \mathop{\mathbb{E}}_{t,t+1 \in \mathcal{T}}
    \bigg[
    \arccos
    \Big(
    \frac
    {
    \langle\mathbf{a} - \mathbf{o}_t, \mathbf{a} - \mathbf{o}_{t+1}\rangle
    }
    {
    \|\mathbf{a} - \mathbf{o}_t\| \cdot \|\mathbf{a} - \mathbf{o}_{t+1}\|
    }
    \Big)
    \bigg]
    \cdot
    N.
\end{equation}

When computing this metric, we assume that (1) the object moves at roughly constant speed, (2) the camera always fixates on the object, and (3) the distance between the camera and the object remains approximately the same over time.
All sequences from existing works as well as ours meet these assumptions, except those from the NSFF~\cite{li2020nsff} and NV-DYN~\cite{yoon2020novel} datasets.
In their case, the cameras are always facing forward, breaking the assumption (2).
However, we find that even though cameras are not fixated on the object since they are static, we can still compute the look-at point $\mathbf{a}$ by considering the center of mass of the foreground visible surfaces in 3D.
Both datasets provide accurate foreground segmentations and MVS depth, which we use to identify and back-project foreground pixels into 3D space.
The final look-at point is computed as the average foreground points over all frames.

\looseness=-1
Note that existing works only provide extracted frames from each sequence without specifying the frame rate.
We identify the frame rates by re-assembling the original video using different FPS candidates and hand-picking the one that results in the most natural object and camera motion, which are verified by the original authors~\cite{li2020nsff,park2021hypernerf,park2021nerfies}.
We document per-sequence FPS for future reference in Table~\ref{tab:data_stats_breakdown}.

\section{Computation for co-visibility mask and masked image metrics}
\label{sec:app_comp_covisible}
Code for both the co-visibility mask and masked image metrics are made publicly available on our \href{\dychecklink}{project page}.
In this section, we provide further details for their computation processes.

\subsection{Co-visibility mask}
\begin{figure}[t!]
    \includegraphics[width=\linewidth,bb=0 0 2190 930]{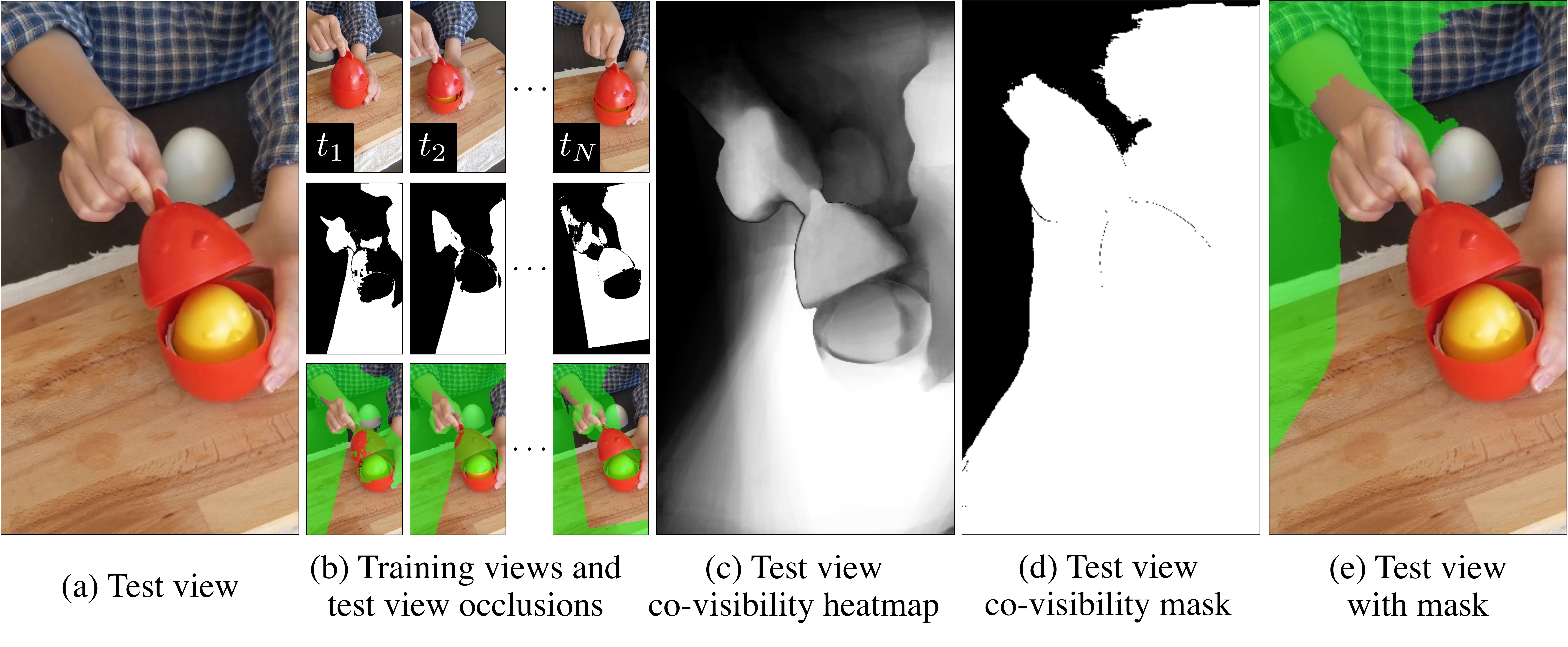}
    \vspace{-20pt}
    \caption{
        \textbf{Illustration of the computation process for co-visibility.}
        Given a (a) test view, we first compute its pairwise (b) occlusions in all training views the forward-backward consistency check~\cite{brox2004high} based on the optical flow estimation from pre-trained RAFT~\cite{teed2020raft}.
        The occlusions are visualized as binary masks in (b)'s second row, where black color indicates pixels without correspondence. 
        We also visualize their overlays over the original test image.
        Then by summing up all occlusion maps, we compute the (c) test view co-visibility heatmap, which stores the number of times each test pixel is seen in training frames.
        Finally, we apply a threshold on the heatmap and obtain a binary (d) co-visibility mask.
        We also visualize its (e) overlay on the test image.
        Note that the occlusion maps are usually inaccurate due to noise in optical flow prediction, \eg, they miss the cover of the chicken toy in this example.
        Our conservative threshold strategy overcomes the noise and ensures that adequately seen regions are included in the final mask.
    }
    \label{fig:covisible_comp_visuals}
\end{figure}

\looseness=-1
In dynamic scenes, particularly for monocular capture with multi-camera validation, the test view contains regions that may not have been observed at all by the training camera.
To circumvent this issue without resorting to camera teleportation, for each pixel in the test image, we propose ``co-visibility'' masking, which tests how many times a test pixel has been observed in the training images.

We visualize the computation process of the co-visibility mask in Figure~\ref{fig:covisible_comp_visuals}.
Concretely, for each (a) test frame, we first check its (b) occlusion in each training frame by the forward-backward flow consistency check according to Equation~\ref{eq:forward_backward_check}.
We use RAFT~\cite{teed2020raft} for optical flow estimation between each test frame and each training frame.
Note that we visualize occlusion as both a binary mask and its overlay on the test image.
For occlusion mask visualization, the black color indicates pixels with no correspondence in the training views.
We then compute the (c) co-visibility heatmap by simply summing up all test view occlusion masks.
This co-visibility heatmap stores the number of times that each pixel is seen in training views.
For visualization purpose, we normalize the heatmap by the number of the training frames $N$.
Finally, we apply a threshold $\beta$ to the heatmap and obtain a (d) binary co-visibility mask, which we also visualize with (e) its overlay on the test image.
We adopt a conservative strategy and set $\beta = \max (5, 0.1\cdot N)$, meaning that we deem a pixel ``seen'' during training and valid for evaluation when it is seen in $5$ or $10\%$ of training frames, whichever is larger.
This strategy ensures high recall in the masking result, \ie, the final co-visible regions are adequately seen during training when the flow estimation is noisy.
For example, as shown in the third row of Figure~\ref{fig:covisible_comp_visuals} (b), the test view occlusions are inaccurate and miss the red cover of the chicken toy when it is visible in both frames.
However, since the red cover is adequately seen over the whole sequence, it is still included in the final co-visibility mask.

\subsection{Masked image metrics}

In this work, we propose to only evaluate on regions that are adequately seen during training by co-visibility masking.
We employ three masked image metrics, namely mPSNR, mSSIM and mLPIPS, which extend from their original definition, which we discuss next.

\paragraph{PSNR $\rightarrow$ mPSNR.} 
PSNR is originally defined as per-pixel mean squared error (MSE) in the log scale (with a constant negative multiplier). 
We compute mPSNR by simply taking the average of per-pixel PSNR scores over the masked region.

\paragraph{SSIM~\cite{wang2004image} $\rightarrow$ mSSIM.} 
Comparing to PSNR, SSIM is defined on the patch level: it considers the structural similarity within each patch.
In practice, it is usually implemented as convolutions where kernels are defined by the pixels in each patch.
We take inspiration from Liu~\etal~\cite{liu2018image} and follow exactly their partial convolution implementation for this operation, where only the masked pixels are accounted for the final result.

\paragraph{LPIPS~\cite{zhang2018unreasonable} $\rightarrow$ mLPIPS~\cite{li2020nsff,gatys2017controlling,huh2020transforming}.} 
\looseness=-1
LPIPS is also defined on patch level.
Given two images, it computes their similarity distance in the feature space across different spatial resolution using a pretrained AlexNet model~\cite{krizhevsky2017imagenet}.
The final similarity score is the average over all distance maps.
To compute mLPIPS, we follow the previous works~\cite{li2020nsff,gatys2017controlling,huh2020transforming} and first apply the co-visibility mask on the input images by zeroing out the unseen regions.
Given the output distance maps at each spatial resolution, we then apply the same mask with downsampling and compute the masked average distance score.
It should be noted that the pretrained AlexNet has a receptive field of $195^2$.
Thus when the co-visibility mask is small (most of the pixels are not seen during training), this metric can be artificially low due to the zeroing operation.

\section{Correspondence readout from existing works}
\label{sec:app_comp_corr}

In this section, we first review the formulation of the existing works and then describe the computation to read out correspondence from these models.

\subsection{Formulation of existing works}

A neural radiance field (NeRF)~\cite{mildenhall2020nerf} represents a \textit{static} scene as a continuous volumetric field $F$ that transforms a point's position $\mathbf{x}$ and auxiliary variables $\mathbf{w}$ (\eg, view direction, latent appearance vector) to color $\mathbf{c}$ and density $\sigma$,
\begin{equation}
  F: (\mathbf{x}, \mathbf{w}) \mapsto (\mathbf{c}, \sigma).
\end{equation}

Here we briefly review representative approaches that extend NeRFs to dynamic scenes.

\noindent \paragraph{Nerfies~\cite{park2021nerfies} and HyperNeRF~\cite{park2021hypernerf}.}
Similarly to traditional non-rigid reconstruction methods that explains non-rigid scenes with a static {\em canonical} space and a per-frame deformation model~\cite{newcombe2015dynamicfusion}, Nerfies~\cite{park2021nerfies} capture a non-rigid scene with one canonical NeRF $F$ and a per-time step view-to-canonical deformation $W_{t\rightarrow c}$ that takes a point $\mathbf{x}$ with a time-conditioned latent vector $\bm{\varphi}_t$ to a canonical point $\mathbf{x}_c$,
\begin{equation}
    W_{t\rightarrow c}: (\mathbf{x}, \bm{\varphi}_t)
    \mapsto
    \mathbf{x}_c.
\end{equation}
At each time step the resulting volumetric field is
 $F_t = F \circ W_{t\rightarrow c}$.
HyperNeRF~\cite{park2021hypernerf} addresses topological change  on top of Nerfies by outputting a two-dimensional ``ambient'' coordinate $\mathbf{w}$ encoding the topological change in addition to the canonical point $\mathbf{x}_c$,
\begin{equation}
    W_{t\rightarrow c}:
    (\mathbf{x}, \bm{\varphi}_t)
    \mapsto
    (\mathbf{x}_c, \mathbf{w}).
\end{equation}
These two output variables are passed to the (topologically varying) canonical space mapping $F$.

\noindent \textbf{Time-conditioned NeRF and NSFF~\cite{li2020nsff}.}
Another way to handle non-rigid scenes is to directly map space-time to the output color and density by a time-conditioned latent vector $\bm{\varphi}_t$, which we refer to as T-NeRF:
\begin{equation}
    F_t:
    (\mathbf{x}, \bm{\varphi}_t)
    \mapsto
    (\mathbf{c}, \sigma).
\end{equation}
Note that since T-NeRF implicitly handles deformation, it is difficult to compute correspondences over time.
NSFF~\cite{li2020nsff} augments T-NeRF's implicit function $F_t$ to output an explicit scene flow field $W_{t \rightarrow t+\delta}$ between adjacent time steps $t$ and $t+\delta$,
\begin{equation}
    W_{t \rightarrow t+\delta}: %
    \mathbf{x}
    \mapsto
    \mathbf{x}^\prime, \ \ \delta\in\{+1, -1\}.
\end{equation}
This explicit flow field is used to regularize motion and, as shown below, can be chained to compute long-range point correspondences across views and times.

\subsection{Correspondence readout}
Our goal is to find view-to-view correspondences such that given a set of key-points on a source image at time $t_1$, we can find their correspondence on a target image at time $t_2$.

For clarity, we start with assuming a known 3D view-to-view warp $W_{t_1\rightarrow t_2}$, outlined in the last sub-section.
The 2D correspondence $\mathbf{u}_{t_2}$ given $\mathbf{u}_{t_1}$ can be obtained by three steps, which we describe as ``warp-integrate-project''.
In the ``warp'' step, given the pixel location $\mathbf{u}_{t_1}$ and camera $\pi_{t_1}$, we sample points on the ray passing from the camera center through the pixel $\pi_{t_1}^{-1}(\mathbf{u}_{t_1})$.  Then, we warp the sampled points toward their 3D correspondences in the target frame using the known 3D warp $W_{t_1\rightarrow t_2}$.
In the ``integrate'' step, we compute the expected 3D location for the source samples weighted by the probability mass $w_{t_1}$ by volume rendering, as per NeRF~\cite{mildenhall2020nerf}.
We can use densities from either source or target frame, a choice that we find insensitive in practice.
In our formulation, we use the densities from the source frame.
Finally, in the ``project'' step, we project the expected 3D location to the target frame through the target camera $\pi_{t_2}$.
The ``warp-integrate-project'' process can be written as
\begin{equation}
    \mathbf{u}_{t_2}
    =
    \pi_{t_2}
    \bigg(
    \mathop{\mathbb{E}}_{\mathbf{x}_{t_1} \in \pi_{t_1}^{-1}(\mathbf{u}_{t_1})}
    \big[
    w_{t_1}(\mathbf{x}_{t_1}) \cdot W_{t_1 \rightarrow t_2}(\mathbf{x}_{t_1})
    \big]
    \bigg).
\end{equation}
Note that there are also other alternatives such as ``warp-project-integrate'' where integration happens after projecting warped points to 2D.
We find in practice that these different approaches make little difference to the final results when the surface is dense such that $w_t$ is concentrated near one point (almost one-hot) for each ray.

\paragraph{Nerfies~\cite{park2021nerfies} and HyperNeRF~\cite{park2021hypernerf}.}
We can compose $W_{t_1\rightarrow t_2}$ by an inverse map $W_{t_1\rightarrow c}$ and a forward map $\tilde{W}_{c\rightarrow t_2}$,
\begin{equation}
    W_{t_1\rightarrow t_2}(\mathbf{x}_{t_1})
    =
    \tilde{W}_{c\rightarrow t_2}(W_{t_1\rightarrow c}(\mathbf{x}_{t_1})).
\end{equation}
We solve for the forward map given the inverse map through optimization:
\begin{equation}
    \tilde{W}_{c\rightarrow t}(\mathbf{x}_t)
    =
    \mathop{\argmin}_{\mathbf{x}_c}
    \|
    W_{t\rightarrow c}(\mathbf{x}_t)
    -
    \mathbf{x}_c
    \|_2^2.
\end{equation}
We use the Broyden solver for root-finding, as per SNARF~\cite{chen2021snarf}, and initialize $\mathbf{x}_c$ with $\mathbf{x}_t$.

\paragraph{NSFF~\cite{li2020nsff}.}
We can compose $W_{t_1\rightarrow t_2}$ by chaining the scene flow predictions through time.
Concretely we have
\begin{equation}
    W_{t_1\rightarrow t_2}(\mathbf{x}_{t_1})
    =
    W_{t_2-1\rightarrow t_2}\bigg(
    \cdots
    W_{t_1+1\rightarrow t_1+2}\big(
    W_{t_1\rightarrow t_1 + 1}(
    \mathbf{x}_{t_1}
    )
    \big)
    \bigg).
\end{equation}

\section{Summary of the capture setup and data processing for our
iPhone dataset}
\label{sec:app_summ_iphone}

\providecommand\animage{}
\renewcommand{\animage}[2]{
	\frame{\begin{overpic}[width=\linewidth,clip,trim=#1]{figures/assets/appendix/app_summ_iphone_visuals/#2}\end{overpic}}
}
\providecommand\textimage{}
\renewcommand{\textimage}[2]{
	\frame{\begin{overpic}[width=\linewidth,clip,trim=#1]{figures/assets/appendix/app_summ_iphone_visuals/#2}\put(0.3,915){\footnotesize\sethlcolor{black}\textcolor{white}{\hl{Excluded}}}\end{overpic}}
}
\providecommand\missingimage{}
\renewcommand{\missingimage}[2]{
	\frame{\begin{overpic}[width=\linewidth,clip,trim=#1]{figures/assets/appendix/app_summ_iphone_visuals/#2}\put(0.3,915){\footnotesize\sethlcolor{black}\textcolor{white}{\hl{Unavailable}}}\end{overpic}}
}
\begin{figure}[t!]
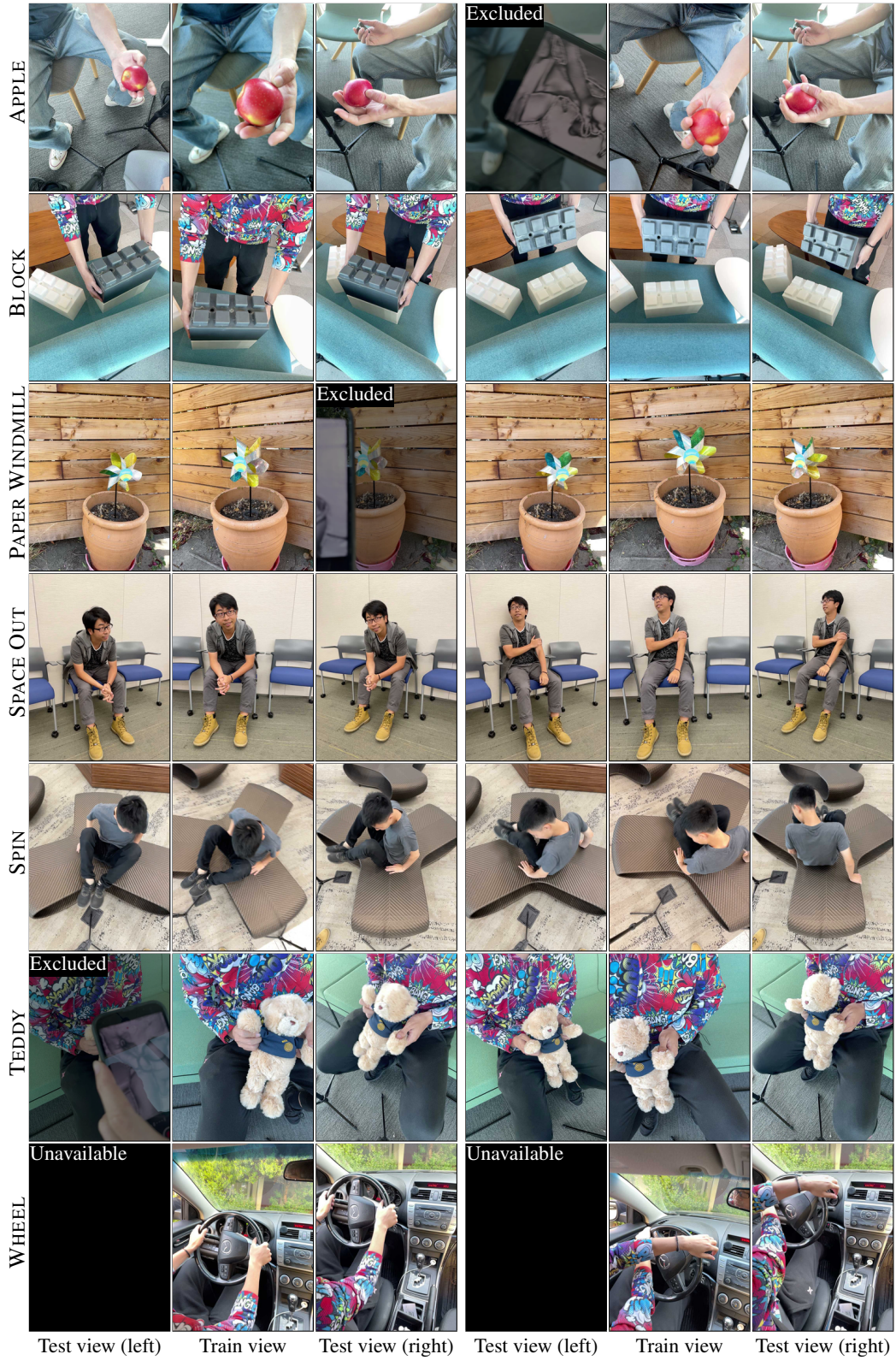

    \setlength{\tabcolsep}{0.8pt}
    \renewcommand{\arraystretch}{0.5}
    \begin{tabularx}{\textwidth}{@{}c*{3}{C}@{}c*{3}{C}@{}}
        \makebox[20pt]{\raisebox{40pt}{\rotatebox[origin=c]{90}{\textsc{Apple}}}}\hspace{-6pt} &
        \animage{0 0 0 0}{iphone2/apple/205_right_val_img.png} &
        \animage{0 0 0 0}{iphone2/apple/205_train_img.png} &
        \animage{0 0 0 0}{iphone2/apple/205_left_val_img.png} &
        &
        \textimage{0 0 0 0}{iphone2/apple/282_right_val_img.png} &
        \animage{0 0 0 0}{iphone2/apple/282_train_img.png} &
        \animage{0 0 0 0}{iphone2/apple/282_left_val_img.png}
        \\
        \makebox[20pt]{\raisebox{40pt}{\rotatebox[origin=c]{90}{\textsc{Block}}}}\hspace{-6pt} &
        \animage{0 0 0 0}{iphone2/block/413_left_val_img.png} &
        \animage{0 0 0 0}{iphone2/block/413_train_img.png} &
        \animage{0 0 0 0}{iphone2/block/413_right_val_img.png} &
        &
        \animage{0 0 0 0}{iphone2/block/65_left_val_img.png} &
        \animage{0 0 0 0}{iphone2/block/65_train_img.png} &
        \animage{0 0 0 0}{iphone2/block/65_right_val_img.png}
        \\
        \makebox[20pt]{\raisebox{40pt}{\rotatebox[origin=c]{90}{\textsc{Paper Windmill}}}}\hspace{-6pt} &
        \animage{0 0 0 0}{iphone2/paper-windmill/481_left_val_img.png} &
        \animage{0 0 0 0}{iphone2/paper-windmill/481_train_img.png} &
        \textimage{0 0 0 0}{iphone2/paper-windmill/481_right_val_img.png} &
        &
        \animage{0 0 0 0}{iphone2/paper-windmill/80_left_val_img.png} &
        \animage{0 0 0 0}{iphone2/paper-windmill/80_train_img.png} &
        \animage{0 0 0 0}{iphone2/paper-windmill/80_right_val_img.png}
        \\
        \makebox[20pt]{\raisebox{40pt}{\rotatebox[origin=c]{90}{\textsc{Space Out}}}}\hspace{-6pt} &
        \animage{0 0 0 0}{iphone2/space-out/180_left_val_img.png} &
        \animage{0 0 0 0}{iphone2/space-out/180_train_img.png} &
        \animage{0 0 0 0}{iphone2/space-out/180_right_val_img.png} &
        &
        \animage{0 0 0 0}{iphone2/space-out/345_left_val_img.png} &
        \animage{0 0 0 0}{iphone2/space-out/345_train_img.png} &
        \animage{0 0 0 0}{iphone2/space-out/345_right_val_img.png}
        \\
        \makebox[20pt]{\raisebox{40pt}{\rotatebox[origin=c]{90}{\textsc{Spin}}}}\hspace{-6pt} &
        \animage{0 0 0 0}{iphone2/spin/148_left_val_img.png} &
        \animage{0 0 0 0}{iphone2/spin/148_train_img.png} &
        \animage{0 0 0 0}{iphone2/spin/148_right_val_img.png} &
        &
        \animage{0 0 0 0}{iphone2/spin/340_left_val_img.png} &
        \animage{0 0 0 0}{iphone2/spin/340_train_img.png} &
        \animage{0 0 0 0}{iphone2/spin/340_right_val_img.png}
        \\
        \makebox[20pt]{\raisebox{40pt}{\rotatebox[origin=c]{90}{\textsc{Teddy}}}}\hspace{-6pt} &
        \textimage{0 0 0 0}{iphone2/teddy/179_left_val_img.png} &
        \animage{0 0 0 0}{iphone2/teddy/179_train_img.png} &
        \animage{0 0 0 0}{iphone2/teddy/179_right_val_img.png} &
        &
        \animage{0 0 0 0}{iphone2/teddy/300_left_val_img.png} &
        \animage{0 0 0 0}{iphone2/teddy/300_train_img.png} &
        \animage{0 0 0 0}{iphone2/teddy/300_right_val_img.png}
        \\
        \makebox[20pt]{\raisebox{40pt}{\rotatebox[origin=c]{90}{\textsc{Wheel}}}}\hspace{-6pt} &
        \missingimage{0 0 0 0}{iphone2/wheel/0_left_val_img.png} &
        \animage{0 0 0 0}{iphone2/wheel/0_train_img.png} &
        \animage{0 0 0 0}{iphone2/wheel/0_right_val_img.png} &
        &
        \missingimage{0 0 0 0}{iphone2/wheel/422_left_val_img.png} &
        \animage{0 0 0 0}{iphone2/wheel/422_train_img.png} &
        \animage{0 0 0 0}{iphone2/wheel/422_right_val_img.png}
        \\
        [2pt] &
        {\small Test view (left)} &
        {\small Train view} &
        {\small Test view (right)} &
        \; &
        {\small Test view (left)} &
        {\small Train view} &
        {\small Test view (right)}
        \\
        \cmidrule(lr){2-4}\cmidrule(lr){6-8}
    \end{tabularx}
    \vspace{-4pt}
    \caption{
        \textbf{Visualizations of the multi-camera captures after time synchronization from the proposed iPhone dataset.}
        In each row, we visualize the frames from both the training camera and two testing cameras at two time steps.
        We intentionally move the training camera in front of each test camera at certain times to ensure that our input sequence covers most of the scene in evaluation.
        When a particular test frame depicts the training camera, we exclude the test frame (denoted as ``Excluded'').
        For the \textsc{Wheel} sequence in the last row, we only employ the right test camera due to limited space to set up the multi-camera rig.
    }
    \vspace{-1em}
    \label{fig:app_summ_iphone_visuals}

\end{figure}
Our capture setup has $7$ multi-camera captures (MV) and $7$ single camera captues (SV).
We evaluate novel-view synthesis on the multi-camera captures and correspondence on all captures.

\paragraph{Multi-camera captures.}
For multi-camera captures, we employ three cameras: one hand-held camera to capture monocular video for training and two stationary mounted cameras for validation.
The two validation cameras face inward from two distinct viewpoints with large baseline.
This wide-baseline setup enables us to better evaluate the shape modeling quality for novel-view synthesis.
We use the ``Record3D'' app~\cite{record3d} on iPhone to record both RGB and depth information at each time step. 
Note that we only collect depth information for training views given that we will only use the depths for supervision.
We discuss the preprocessing procedure for the training video sequence in the ``Single-camera captures'' paragraph below.

To synchronize multiple cameras, we leverage the ``audio-based multi-camera synchronization'' functionality in Adobe Premiere Pro, as per~\cite{ouyang2022real}, which achieves millisecond-level accuracy.
In Figure~\ref{fig:app_summ_iphone_visuals}, we show visualizations of our multi-camera captures after time synchronization.
To ensure that our input sequence covers most of the scene regions in evaluation, we intentionally move the training camera in front of each test camera at certain frames.
When we do so, that particular test frame is excluded due to severe occlusion (shown as ``Excluded'' in the figure).
For the \textsc{Wheel} sequence (last row), we only employ the right camera due to the limited physical space to set up the multi-camera rig in that scene.

After time synchronization, we calibrate the multi-camera system.
The Record3D app provides camera parameters and poses at each time step, but the poses only relate to each other \textit{within} each capture sequence.
In fact, each camera pose is recorded as relative pose to the first frame in each sequence, with the first pose being identity.
We therefore need to solve the relative $SE(3)$ transforms between the first frame in each test sequence with respect to the first training frame.
This problem can be formulated as a Perspective-n-Point~(PnP) problem where, given a set of 3D points and their corresponding 2D pixels in two sequences, we aim to solve the camera pose.
In practice, given a training RGBD frame and a testing RGB frame, we compute a set of 2D correspondences by SIFT feature matching~\cite{lowe2004distinctive} and obtain their 3D point positions (in the training sequence's world space) by back-projecting the 2D keypoints with the training frame depth map.
This process is repeated for all time steps.
We exploit our problem structure by constraining the camera poses within each test sequence to be the same, \ie, static camera.
We use the RANSAC PnP solver~\cite{zuliani2009ransac} in OpenCV~\cite{opencv_library}.

\paragraph{Single-camera captures.}
\providecommand\animage{}
\renewcommand{\animage}[2]{
	\frame{\begin{overpic}[width=\linewidth,clip,trim=#1]{figures/assets/appendix/app_summ_iphone_depth_visuals/#2}\end{overpic}}
}
\begin{figure}[t!]
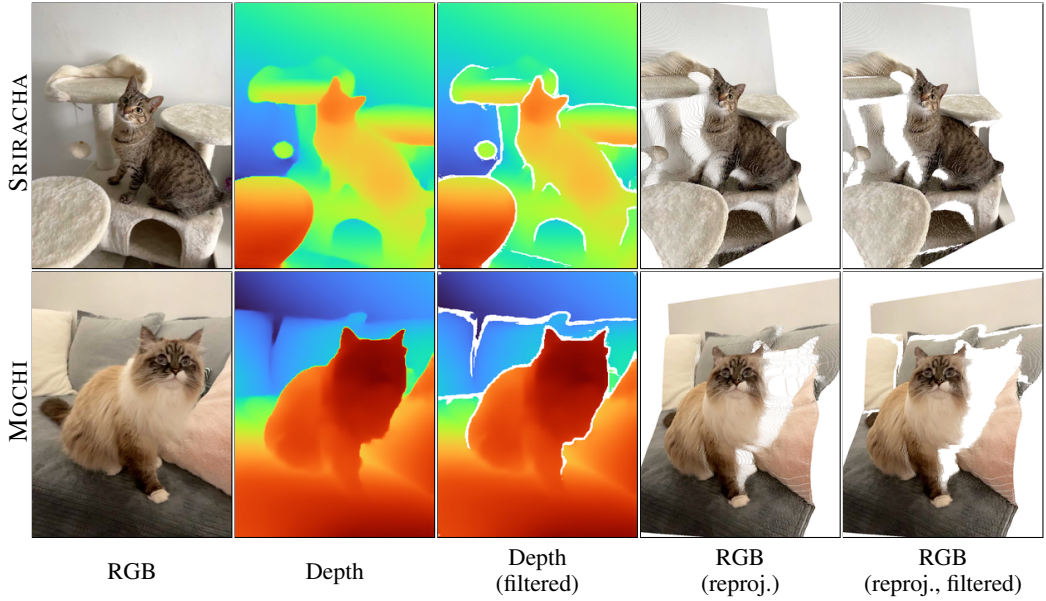

    \setlength{\tabcolsep}{0.8pt}
    \renewcommand{\arraystretch}{0.5}
    \begin{tabularx}{\textwidth}{@{}c*{5}{C}@{}}
        \makebox[20pt]{\raisebox{48pt}{\rotatebox[origin=c]{90}{\textsc{Sriracha}}}}\hspace{-6pt} &
        \animage{0 0 0 0}{iphone2/sriracha-tree/67_rgb.png} &
        \animage{0 0 0 0}{iphone2/sriracha-tree/67_depth.png} &
        \animage{0 0 0 0}{iphone2/sriracha-tree/67_depth_filtered.png} &
        \animage{0 0 0 0}{iphone2/sriracha-tree/67_rgb_reproj.png} &
        \animage{0 0 0 0}{iphone2/sriracha-tree/67_rgb_reproj_filtered.png}
        \\
        \makebox[20pt]{\raisebox{48pt}{\rotatebox[origin=c]{90}{\textsc{Mochi}}}}\hspace{-6pt} &
        \animage{0 0 0 0}{iphone2/mochi-high-five/75_rgb.png} &
        \animage{0 0 0 0}{iphone2/mochi-high-five/75_depth.png} &
        \animage{0 0 0 0}{iphone2/mochi-high-five/75_depth_filtered.png} &
        \animage{0 0 0 0}{iphone2/mochi-high-five/75_rgb_reproj.png} &
        \animage{0 0 0 0}{iphone2/mochi-high-five/75_rgb_reproj_filtered.png}
        \\
        [2pt] &
        {\small RGB} &
        {\small Depth} &
        {\small \makecell{Depth\\(filtered)}} &
        {\small \makecell{RGB\\(reproj.)}} &
        {\small \makecell{RGB\\(reproj., filtered)}}
    \end{tabularx}
    \vspace{-4pt}
    \caption{
        \textbf{Visualizations of the depth filtering during data preprocessing of the proposed iPhone dataset.}
        The depth sensing is particularly noisy around the object edges, which we filtered out by Sobel filter~\cite{sobel2014an}.
        We visualize the re-projected RGB image with the original ($2^{\text{nd}}$ column) or filtered depth ($3^{\text{rd}}$ column) from the captured view to the first view in each sequence at the last two columns.
        Without filtering ($4^{\text{th}}$ column), there are erroneous floaters which cause too much noise for training supervision.
        With filtering (last column), we have a crisper depth map which is used for improving the state-of-the-art methods.
    }
    \vspace{-1em}
    \label{fig:app_summ_iphone_depth_visuals}

\end{figure}
\providecommand\animage{}
\renewcommand{\animage}[2]{
	\frame{\begin{overpic}[width=\linewidth,clip,trim=#1]{figures/assets/appendix/app_summ_iphone_keypoint_visuals/#2}\end{overpic}}
}
\begin{figure}[t!]
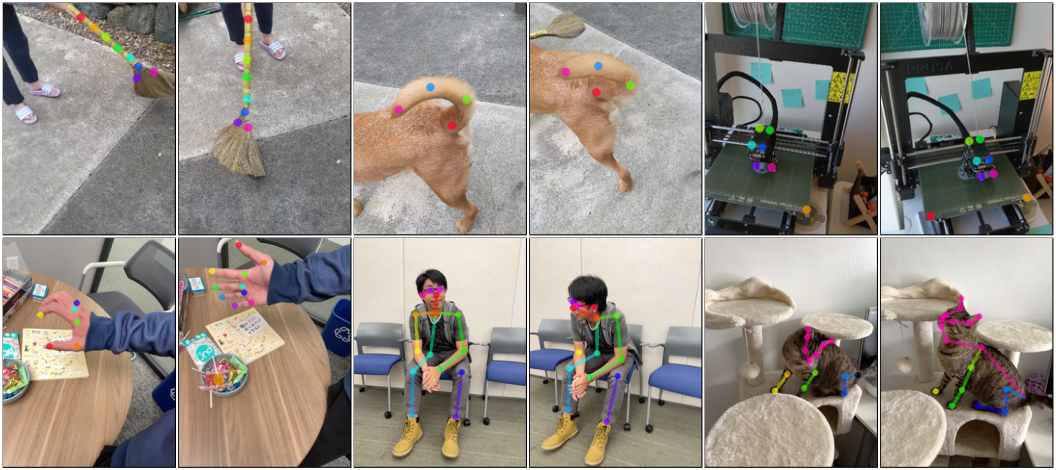

    \setlength{\tabcolsep}{0.8pt}
    \renewcommand{\arraystretch}{0.5}
    \begin{tabularx}{\textwidth}{*{6}{C}@{}}
        \animage{0 40 0 40}{nerfies/broom/4.png} &
        \animage{0 40 0 40}{nerfies/broom/5.png} &
        \animage{0 40 0 40}{nerfies/tail/0.png} &
        \animage{0 40 0 40}{nerfies/tail/7.png} &
        \animage{0 40 0 40}{hypernerf/vrig-3dprinter/1.png} &
        \animage{0 40 0 40}{hypernerf/vrig-3dprinter/6.png}
        \\
        \animage{0 0 0 0}{iphone2/handwavy/6.png} &
        \animage{0 0 0 0}{iphone2/handwavy/8.png} &
        \animage{0 0 0 0}{iphone2/space-out/4.png} &
        \animage{0 0 0 0}{iphone2/space-out/6.png} &
        \animage{0 0 0 0}{iphone2/sriracha-tree/1.png} &
        \animage{0 0 0 0}{iphone2/sriracha-tree/2.png}
    \end{tabularx}
    \vspace{-4pt}
    \caption{
        \textbf{Visualizations of the keypoint annotation during data preprocessing of the proposed iPhone dataset.}
        We manually annotate keypoints for correspondence evaluation.
        For sequences of humans and quadrupeds (dogs or cats), we annotate based on the skeleton defined in the COCO challenge~\cite{lin2014microsoft} and StanfordExtra~\cite{biggs2020left}. For sequences that focus on more general objects, we manually identify and annotate $5$ to $15$ trackable keypoints across frames.
    }
    \vspace{-1em}
    \label{fig:app_summ_iphone_keypoint_visuals}

\end{figure}
We treat the single-camera captures as the training sequence in our multi-camera capture setup.
In effect, the single-camera capture setup will not have validation data for novel-view synthesis evaluation.
We preprocess the depth data for the training sequence by applying a Sobel filter~\cite{sobel2014an} to filter out inaccurate depth values around object edges.
In Figure~\ref{fig:app_summ_iphone_depth_visuals}, we visualize our depth data before and after filtering.
We find that NeRF is particularly sensitive to depth noise and this filtering step is necessary.
Finally, we manually annotate keypoints for correspondence evaluation.
For sequences of humans and quadrupeds (dogs or cats), we annotate keypoints based on the skeleton defined in the COCO challenge~\cite{lin2014microsoft} and StanfordExtra~\cite{biggs2020left}.
For sequences that focus on more general objects (\eg, our \textsc{Block} and \textsc{Teddy} sequences), we manually identify and annotate 5 to 15 trackable keypoints across frames. 
We visualize keypoint annotations (with skeleton if available) for both our proposed iPhone dataset and the Nerfies-HyperNeRF dataset in Figure~\ref{fig:app_summ_iphone_keypoint_visuals}.

Note that both Nerfies~\cite{park2021nerfies} and HyperNeRF~\cite{park2021hypernerf} use background regularization which requires a point cloud of the background static scene.
We first extract the object mask over time by MTTR, an off-the-shelf video segmentation network~\cite{botach2021end}, which takes a text prompt of the foreground object as input.
Since our foreground objects are quite diverse (\eg, backpack and block), the segmentation results are usually noisy.
Thus we apply TSDF Fusion~\cite{newcombe2011kinectfusion} to the background point clouds over the whole sequence to get a completed background point cloud.
We find that this point cloud can be noisy when segmentation fails, and that it is necessary to manually filter the background point cloud to make sure that it does not include any foreground regions.
We consider this manual process a weakness of the previous background regularization~\cite{park2021nerfies}.

\section{Summary of the implementation details and remaining differences}
\label{sec:app_summ_impl}

\looseness=-1
To ensure a fair comparison, we align numerous training details between the models that we investigate in this paper: T-NeRF, NSFF~\cite{li2020nsff}, Nerfies~\cite{park2021nerfies} and HyperNeRF~\cite{park2021hypernerf}.
Code and checkpoints are available on our \href{\dychecklink}{project page}.

To start with, we align the total number of rays seen during training. 
We add support of ray undistortion~\cite{park2021hypernerf} in the third-party implementation of NSFF~\cite{nsff_pl} to make sure that the training rays are the same across codebases.
All models are trained with view-dependency modeling turned on.
We did not find appearance encoding~\cite{martin2021nerf} helpful in terms of quantitative results.
This might due to the lighting difference between training and validation captures -- a common issue in evaluation discussed in mip-NeRF 360~\cite{barron2021mip}.

\begin{table}[t!]
\centering
\tablestyle{4pt}{1.2}
\resizebox{\textwidth}{!}{%
\setlength{\tabcolsep}{4pt}
\begin{tabular}{l ccc ccc ccc ccc}
\toprule
& 
\multicolumn{3}{c}{\makecell{\textsc{Broom}\\$\Omega=3.40, \omega=128.54$}} &
\multicolumn{3}{c}{\makecell{\textsc{Curls}\\$\Omega=1.20, \omega=138.55$}} &
\multicolumn{3}{c}{\makecell{\textsc{Tail}\\$\Omega=3.30, \omega=160.55$}} &
\multicolumn{3}{c}{\makecell{\textsc{Toby Sit}\\$\Omega=2.18, \omega=110.51$}}
\\
Method &
PSNR$\uparrow$ &SSIM$\uparrow$ &LPIPS$\downarrow$ &
PSNR$\uparrow$ &SSIM$\uparrow$ &LPIPS$\downarrow$ &
PSNR$\uparrow$ &SSIM$\uparrow$ &LPIPS$\downarrow$ &
PSNR$\uparrow$ &SSIM$\uparrow$ &LPIPS$\downarrow$
\\
\cmidrule(lr){1-1}\cmidrule(lr){2-4}\cmidrule(lr){5-7}\cmidrule(lr){8-10}\cmidrule(lr){11-13}
Nerfies~\cite{park2021nerfies} &
$19.40$ & - & $0.323$ &
- & - & - &
- & - & - &
- & - & -
\\
Nerfies~(repo) &
$19.40$ & - & $0.325$ &
$\mathbf{24.40}$ & - & $0.392$ &
$\mathbf{21.90}$ & - & $0.245$ &
$18.44$ & - & $0.384$
\\
Nerfies~(our reimpl.) &
$\mathbf{19.70}$ & $\mathbf{0.216}$ & $\mathbf{0.296}$ &
$24.04$ & $\mathbf{0.670}$ & $\mathbf{0.245}$ &
$21.79$ & $\mathbf{0.314}$ & $\mathbf{0.236}$ &
$\mathbf{18.48}$ & $\mathbf{0.355}$ & $\mathbf{0.375}$
\\
\cmidrule(lr){1-1}\cmidrule(lr){2-4}\cmidrule(lr){5-7}\cmidrule(lr){8-10}\cmidrule(lr){11-13}
HyperNeRF~\cite{park2021hypernerf} &
$19.30$ & - & $\mathbf{0.296}$ &
- & - & - &
- & - & - &
- & - & -
\\
HyperNeRF~(repo) &
$19.30$ & - & $0.308$ &
$\mathbf{24.60}$ & - & $0.363$ &
$22.10$ & - & $\mathbf{0.226}$ &
$18.40$ & - & $\mathbf{0.330}$
\\
HyperNeRF~(our reimpl.) &
$\mathbf{19.36}$ & $\mathbf{0.210}$ & $0.314$ &
$24.59$ & $\mathbf{0.686}$ & $\mathbf{0.247}$ &
$\mathbf{22.16}$ & $\mathbf{0.329}$ & $0.231$ &
$\mathbf{18.41}$ & $\mathbf{0.345}$ & $0.339$
\\
\bottomrule
\end{tabular}%
}
\newline\newline
\resizebox{0.8\textwidth}{!}{%
\setlength{\tabcolsep}{4pt}
\begin{tabular}{l ccc ccc ccc}
\toprule
& 
\multicolumn{3}{c}{\makecell{\textsc{3D Printer}\\$\Omega=3.13, \omega=251.37$}} &
\multicolumn{3}{c}{\makecell{\textsc{Chicken}\\$\Omega=7.38, \omega=212.58$}} &
\multicolumn{3}{c}{\makecell{\textsc{Peel Banana}\\$\Omega=1.26, \omega=237.66$}}
\\
Method &
PSNR$\uparrow$ &SSIM$\uparrow$ &LPIPS$\downarrow$ &
PSNR$\uparrow$ &SSIM$\uparrow$ &LPIPS$\downarrow$ &
PSNR$\uparrow$ &SSIM$\uparrow$ &LPIPS$\downarrow$
\\
\cmidrule(lr){1-1}\cmidrule(lr){2-4}\cmidrule(lr){5-7}\cmidrule(lr){8-10}
Nerfies~\cite{park2021nerfies} &
$20.20$ & - & $\mathbf{0.115}$ &
$26.00$ & - & $0.084$ &
$21.70$ & - & $\mathbf{0.157}$
\\
Nerfies~(repo) &
$20.20$ & - & $0.118$ &
$\mathbf{26.80}$ & - & $0.081$ &
$\mathbf{22.00}$ & - & $0.179$
\\
Nerfies~(our reimpl.) &
$\mathbf{20.30}$ & $\mathbf{0.639}$ & $\mathbf{0.115}$ &
$26.54$ & $\mathbf{0.823}$ & $\mathbf{0.079}$ &
$21.11$ & $\mathbf{0.693}$ & $0.174$
\\
\cmidrule(lr){1-1}\cmidrule(lr){2-4}\cmidrule(lr){5-7}\cmidrule(lr){8-10}
HyperNeRF~\cite{park2021hypernerf} &
$20.00$ & - & $0.111$ &
$26.90$ & - & $0.079$ &
$\mathbf{23.30}$ & - & $\mathbf{0.133}$
\\
HyperNeRF~(repo) &
$20.10$ & - & $\mathbf{0.110}$ &
$27.70$ & - & $\mathbf{0.076}$ &
$22.20$ & - & $0.140$
\\
HyperNeRF~(our reimpl.) &
$\mathbf{20.12}$ & $\mathbf{0.638}$ & $\mathbf{0.110}$ &
$\mathbf{27.74}$ & $\mathbf{0.834}$ & $0.077$ &
$22.25$ & $\mathbf{0.729}$ & $0.144$
\\
\bottomrule
\end{tabular}%
}
\vspace{0.5em}
\caption{
    \textbf{Our re-implementation reproduces Nerfies~\cite{park2021nerfies}'s and HyperNeRF~\cite{park2021hypernerf}'s results.}
    The official numbers for both Nerfies and HyperNeRF are taken from the HyperNeRF paper.
    Our results matches closely to their numbers and the ones that we obtained by running the officially released repositories (denoted as ``repo'').
    All models are trained under the teleporting setting.    
}
\label{tab:app_summ_impl_nfhn_match}
\end{table}
T-NeRF, Nerfies, and HyperNeRF share the exact same training setup since they are implemented within our codebase. 
We follow the hyper-parameters specified in their official repositories.
We use a batch size $B = 6144$ for a total number of iterations $N = 2.5 \times 10^5$, optimized by ADAM~\cite{kingma2014adam} with an initial learning rate $\eta = 1 \times 10^{-3}$ exponentially decayed to $1 \times^{-4}$ at the end.
We use this training recipe for all of our experiments across all datasets.
On $4$ NVIDIA RTX A4000 or $2$ NVIDIA A100 GPUs with $24\si{GB}$ memory, it takes roughly $12$ hours to train a T-NeRF and $24$ hours to train a Nerfies or a HyperNeRF.
In Table~\ref{tab:app_summ_impl_nfhn_match}, we show that our codebase reproduces the numbers from the original papers and official repositories.

\begin{table}[t!]
\centering
\tablestyle{4pt}{1.2}
\resizebox{0.4\textwidth}{!}{%
\setlength{\tabcolsep}{4pt}
\begin{tabular}{l ccc}
\toprule
& 
\multicolumn{3}{c}{\makecell{\textsc{Jumping}\\$\Omega=0.68, \omega=53.79$}}
\\
Method &
PSNR$\uparrow$ &SSIM$\uparrow$ &LPIPS$\downarrow$
\\
\cmidrule(lr){1-1}\cmidrule(lr){2-4}
NSFF~(repo) &
$27.41$ & $0.900$ & $0.057$
\\
NSFF~(our reimpl.) &
$\mathbf{27.80}$ & $\mathbf{0.908}$ & $\mathbf{0.051}$
\\
\bottomrule
\end{tabular}%
}
\vspace{0.5em}
\caption{
    \textbf{Our modified third-party re-implementation reproduces NSFF~\cite{li2020nsff}'s results on one sequence from Yoon~\etal~\cite{yoon2020novel}.}
    Due to the absence of per-sequence results in the original paper, we compare to the numbers that we obtained by evaluating the officially released checkpoints (denoted as ``repo'').
    Our results matches closely to their numbers.
    All models are trained under the teleporting setting.    
}
\label{tab:app_summ_impl_nsff_match}
\end{table}
Due to no publicly available code to train NSFF on the Nerfies-HyperNeRF dataset.
We adapt and extend the third-party implementation of NSFF (which we find to perform better than the official repo~\cite{nsff}).
We confirm the finding from HyperNeRF that the default hyper-parameters in the NSFF paper are not suitable for long video sequences, and use their hyper-parameters instead.
In Table~\ref{tab:app_summ_impl_nsff_match}, we check on one sequence that our modified re-implementation of NSFF can reproduce the numbers from the ones we obtain by running the released code.
On $1$ NVIDIA RTX A4000 or NVIDIA A100 GPU, it takes roughly $72$ to train a NSFF.
With better implementation, we hypothesize that the training process can be largely accelerated.

\looseness=-1
While we try to ensure the fairness in our comparison, there are still four main remaining differences, namely: (1) static scene stablization, (2) sampling and rendering, (3) NeRF coordinates, and (4) flow supervision.
First, Nerfies and HyperNeRF use additional background points from SfM system as supervision to stabilize the static region of the scene, which we find sensitive to foreground segmentation errors as mentioned in Section~\ref{sec:app_summ_iphone}.
On the other hand, NSFF stabilizes the static region by composing the samples from a time-invariant static NeRF and a time-varying dynamic NeRF. 
Second, Nerfies and HyperNeRF sample $S = 128$ points during the coarse stage, and another $2 S$ points during the fine stage, evaluating $3 S = 384$ points in total.
NSFF, on the other hand, only samples $S$ points for dynamic NeRF and another $S$ points for static NeRF,  without coarse-to-fine sampling, evaluating $2 S = 256$ points in total.
Third, Nerfies and HyperNeRF sample points in world space, while NSFF samples in normalized device coordinates~(NDC), which can cause issues when applying to non-forward-facing scenes like the ones we use in this paper.
Finally, NSFF uses additional optical flow supervision, while Nerfies and HyperNeRF do not.
In fact, we consider the fact that NSFF can leverage correspondence supervision as a merit in the sense that it is non-trivial to apply optical flow supervision to Nerfies and HyperNeRF since their warp representation is not fully invertible.

\section{Additional results on the impact of effective multi-view}
\label{sec:app_emv_results}
\providecommand\animage{}
\renewcommand{\animage}[2]{
    \frame{\includegraphics[width=\linewidth,clip,trim=#1]{figures/assets/impact_of_effective_multiview/#2}}
}
\providecommand\textimage{}
\renewcommand{\textimage}[4]{
	\frame{\begin{overpic}[width=\linewidth,clip,trim=#1]{figures/assets/impact_of_effective_multiview/#2}\put(0,908){\footnotesize\sethlcolor{black}\textcolor{white}{\hl{$#3/#4$}}}\end{overpic}}
}
\begin{figure}[t!]
    \centering
    \begin{minipage}{\linewidth}
        \setlength{\tabcolsep}{0.8pt}
        \renewcommand{\arraystretch}{0.5}
        \begin{tabularx}{\textwidth}{@{}c*{6}{C}@{}}
            \multirow{2}{*}[15pt]{\makebox[20pt]{\raisebox{0pt}{\rotatebox[origin=c]{90}{\textsc{Broom}}}}\hspace{-6pt}} &
            \animage{0 48 0 60}{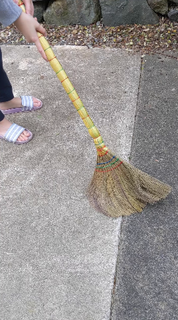} &
            \animage{0 48 0 60}{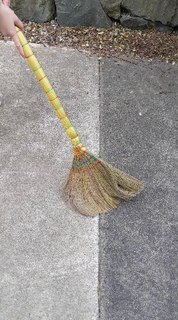} &
            \textimage{0 48 0 60}{nerfies/broom/0_masked_tn_intl.png}{3.40}{128.54} &
            \animage{0 48 0 60}{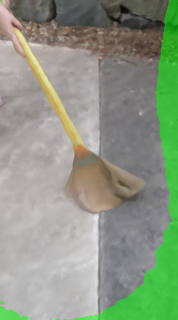} &
            \animage{0 48 0 60}{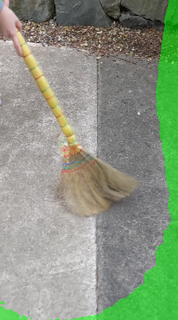} &
            \animage{0 48 0 60}{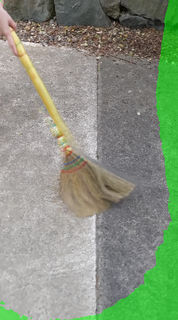}
            \\
            &
            &
            &
            \textimage{0 48 0 60}{nerfies/broom/0_masked_tn_mono.png}{2.57}{60.36} &
            \animage{0 48 0 60}{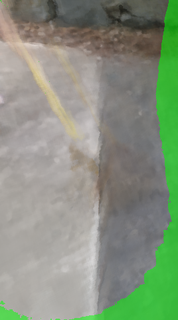} &
            \animage{0 48 0 60}{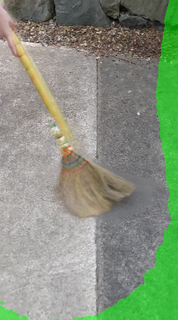} &
            \animage{0 48 0 60}{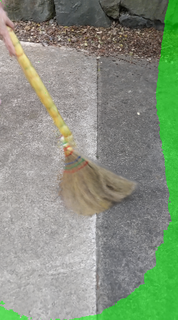}
            \\
            \multirow{2}{*}[24pt]{\makebox[20pt]{\raisebox{0pt}{\rotatebox[origin=c]{90}{\textsc{3D Printer}}}}\hspace{-6pt}} &
            \animage{0 48 0 60}{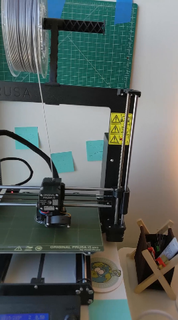} &
            \animage{0 48 0 60}{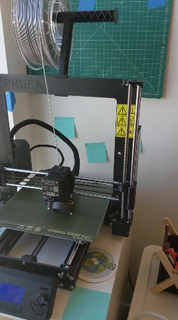} &
            \textimage{0 48 0 60}{hypernerf/vrig-3dprinter/46_masked_tn_intl.png}{3.13}{251.37} &
            \animage{0 48 0 60}{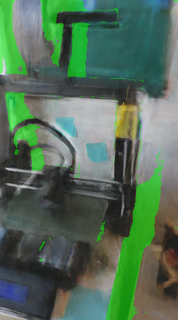} &
            \animage{0 48 0 60}{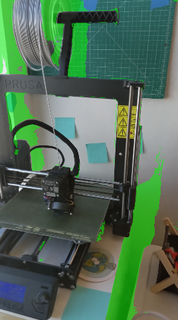} &
            \animage{0 48 0 60}{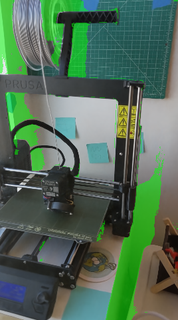}
            \\
            &
            &
            &
            \textimage{0 48 0 60}{hypernerf/vrig-3dprinter/46_masked_tn_mono.png}{1.22}{59.40} &
            \animage{0 48 0 60}{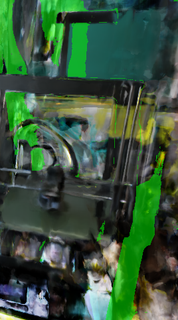} &
            \animage{0 48 0 60}{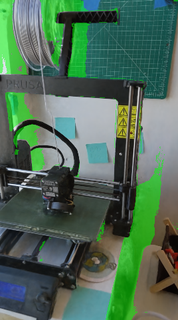} &
            \animage{0 48 0 60}{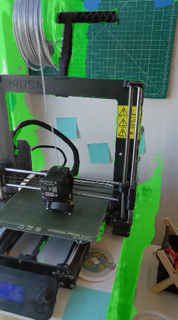}
            \\
            \multirow{2}{*}[18pt]{\makebox[20pt]{\raisebox{0pt}{\rotatebox[origin=c]{90}{\textsc{Chicken}}}}\hspace{-6pt}} &
            \animage{0 48 0 60}{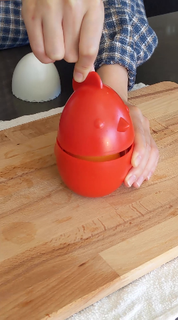} &
            \animage{0 48 0 60}{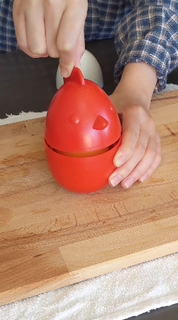} &
            \textimage{0 48 0 60}{hypernerf/vrig-chicken/69_masked_tn_intl.png}{7.38}{212.58} &
            \animage{0 48 0 60}{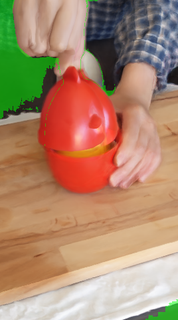} &
            \animage{0 48 0 60}{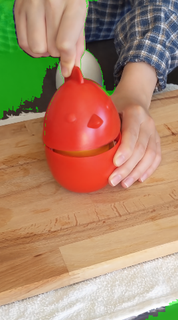} &
            \animage{0 48 0 60}{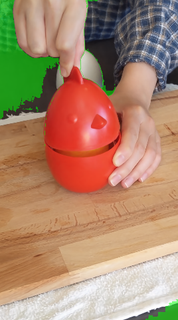}
            \\
            &
            &
            &
            \textimage{0 48 0 60}{hypernerf/vrig-chicken/69_masked_tn_mono.png}{1.52}{33.46} &
            \animage{0 48 0 60}{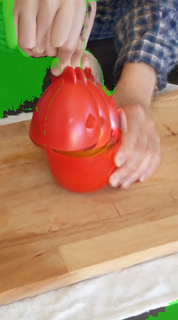} &
            \animage{0 48 0 60}{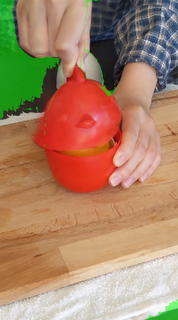} &
            \animage{0 48 0 60}{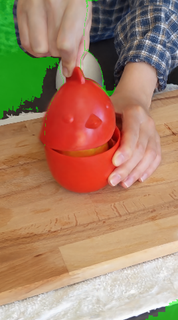}
            \\
            \multirow{2}{*}[28pt]{\makebox[20pt]{\raisebox{0pt}{\rotatebox[origin=c]{90}{\textsc{Peel Banana}}}}\hspace{-6pt}} &
            \animage{0 48 0 60}{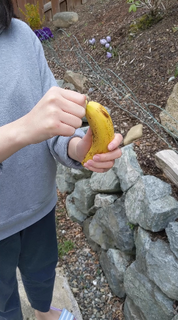} &
            \animage{0 48 0 60}{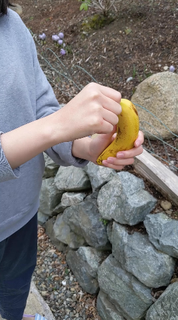} &
            \textimage{0 48 0 60}{hypernerf/vrig-peel-banana/0_masked_tn_intl.png}{1.26}{237.66} &
            \animage{0 48 0 60}{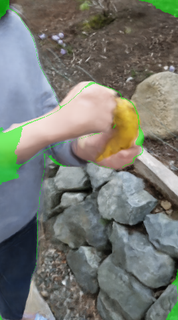} &
            \animage{0 48 0 60}{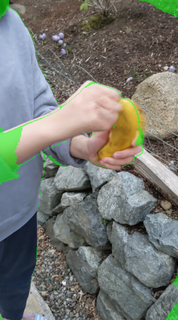} &
            \animage{0 48 0 60}{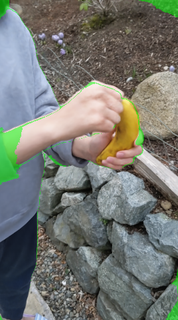}
            \\
            &
            &
            &
            \textimage{0 48 0 60}{hypernerf/vrig-peel-banana/0_masked_tn_mono.png}{0.33}{33.77} &
            \animage{0 48 0 60}{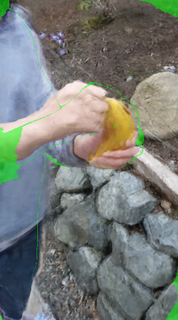} &
            \animage{0 48 0 60}{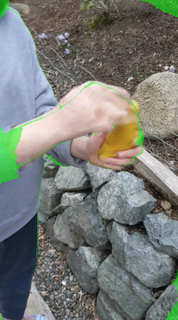} &
            \animage{0 48 0 60}{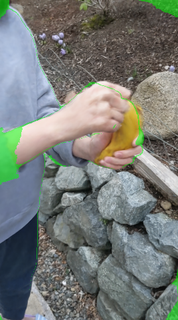}
            \\
            [2pt] &
            {\small Training view} &
            {\small Test view} &
            {\small T-NeRF} &
            {\small NSFF~\cite{li2020nsff}} &
            {\small Nerfies~\cite{park2021nerfies}} &
            {\small HyperNeRF~\cite{park2021hypernerf}}
        \end{tabularx}
    \end{minipage}
    \vspace{-2pt}
    \caption{
        \textbf{Additional qualitative results on the impact of effective multi-view on the Nerfies-HyperNeRF dataset.}
        $\Omega/\omega$ metrics of the input sequence are shown on the top-left.
        We compare the existing camera teleporting setting and our non-teleporting setting.
        For every two rows, we show the results trained with and without camera teleportation in the first and second rows.
        Two settings use the same set of co-visibility masks computed from common training images.
    }
    \label{fig:app_emv_results}
\end{figure}

In Figure~\ref{fig:app_emv_results}, we provide more qualitative comparisons between models that are trained with and without camera teleportation on the Nerfies-HyperNeRF dataset.

\section{Additional results on per-sequence quantitative performance breakdown}
\label{sec:app_breakdown_results}
\begin{table}[ht]
\centering
\tablestyle{4pt}{1.2}
\resizebox{\textwidth}{!}{  
\setlength{\tabcolsep}{4pt}
\begin{tabular}{l cccc cccc}
\toprule
& \multicolumn{4}{c}{\textsc{Broom} ($\Omega=2.57$, $\omega=60.4$)} & \multicolumn{4}{c}{\textsc{Curls} ($\Omega=0.90$, $\omega=118.7$)}\\
Method &mPSNR$\uparrow$ &mSSIM$\uparrow$ &mLPIPS$\downarrow$ &PCK-T$\uparrow$ &mPSNR$\uparrow$ &mSSIM$\uparrow$ &mLPIPS$\downarrow$ &PCK-T$\uparrow$\\
\cmidrule(lr){1-1}\cmidrule(lr){2-5}\cmidrule(lr){6-9}
T-NeRF &$20.04${\color{gray}\scriptsize($20.17$)} &$\mathbf{0.344}${\color{gray}\scriptsize($0.257$)} &$0.590${\color{gray}\scriptsize($0.624$)} &- &$21.86${\color{gray}\scriptsize($21.75$)} &$0.677${\color{gray}\scriptsize($0.597$)} &$0.284${\color{gray}\scriptsize($0.341$)} &- \\
NSFF &$\mathbf{20.36}${\color{gray}\scriptsize($20.46$)} &$0.335${\color{gray}\scriptsize($0.247$)} &$\mathbf{0.776}${\color{gray}\scriptsize($0.813$)} &$0.119$ &$18.74${\color{gray}\scriptsize($18.85$)} &$0.616${\color{gray}\scriptsize($0.531$)} &$\mathbf{0.378}${\color{gray}\scriptsize($0.423$)} &$0.212$ \\
Nerfies &$19.34${\color{gray}\scriptsize($19.51$)} &$0.293${\color{gray}\scriptsize($0.202$)} &$0.294${\color{gray}\scriptsize($0.327$)} &$0.460$ &$\mathbf{23.28}${\color{gray}\scriptsize($23.03$)} &$\mathbf{0.707}${\color{gray}\scriptsize($0.630$)} &$0.220${\color{gray}\scriptsize($0.266$)} &$0.782$ \\
HyperNeRF &$19.04${\color{gray}\scriptsize($19.23$)} &$0.288${\color{gray}\scriptsize($0.197$)} &$0.279${\color{gray}\scriptsize($0.313$)} &$\mathbf{0.471}$ &$23.13${\color{gray}\scriptsize($22.98$)} &$0.700${\color{gray}\scriptsize($0.625$)} &$0.220${\color{gray}\scriptsize($0.266$)} &$\mathbf{0.838}$ \\
\bottomrule
\end{tabular}
}
\newline\newline
\resizebox{\textwidth}{!}{  
\setlength{\tabcolsep}{4pt}
\begin{tabular}{l cccc cccc}
\toprule
& \multicolumn{4}{c}{\textsc{Tail} ($\Omega=1.31$, $\omega=28.6$)} & \multicolumn{4}{c}{\textsc{Toby-sit} ($\Omega=1.28$, $\omega=26.4$)}\\
Method &mPSNR$\uparrow$ &mSSIM$\uparrow$ &mLPIPS$\downarrow$ &PCK-T$\uparrow$ &mPSNR$\uparrow$ &mSSIM$\uparrow$ &mLPIPS$\downarrow$ &PCK-T$\uparrow$\\
\cmidrule(lr){1-1}\cmidrule(lr){2-5}\cmidrule(lr){6-9}
T-NeRF &$\mathbf{22.56}${\color{gray}\scriptsize($22.11$)} &$0.460${\color{gray}\scriptsize($0.385$)} &$0.305${\color{gray}\scriptsize($0.365$)} &- &$18.53${\color{gray}\scriptsize($18.53$)} &$0.428${\color{gray}\scriptsize($0.330$)} &$0.421${\color{gray}\scriptsize($0.471$)} &- \\
NSFF &$21.94${\color{gray}\scriptsize($21.72$)} &$\mathbf{0.461}${\color{gray}\scriptsize($0.388$)} &$\mathbf{0.522}${\color{gray}\scriptsize($0.579$)} &$0.323$ &$\mathbf{18.66}${\color{gray}\scriptsize($18.65$)} &$\mathbf{0.429}${\color{gray}\scriptsize($0.329$)} &$\mathbf{0.600}${\color{gray}\scriptsize($0.634$)} &$0.666$ \\
Nerfies &$21.46${\color{gray}\scriptsize($21.17$)} &$0.385${\color{gray}\scriptsize($0.305$)} &$0.213${\color{gray}\scriptsize($0.261$)} &$\mathbf{0.645}$ &$18.45${\color{gray}\scriptsize($18.41$)} &$0.423${\color{gray}\scriptsize($0.326$)} &$0.249${\color{gray}\scriptsize($0.307$)} &$\mathbf{0.914}$ \\
HyperNeRF &$21.54${\color{gray}\scriptsize($21.13$)} &$0.382${\color{gray}\scriptsize($0.301$)} &$0.218${\color{gray}\scriptsize($0.263$)} &$0.623$ &$18.40${\color{gray}\scriptsize($18.33$)} &$0.422${\color{gray}\scriptsize($0.324$)} &$0.242${\color{gray}\scriptsize($0.300$)} &$0.883$ \\
\bottomrule
\end{tabular}
}
\newline\newline
\resizebox{\textwidth}{!}{  
\setlength{\tabcolsep}{4pt}
\begin{tabular}{l cccc cccc}
\toprule
& \multicolumn{4}{c}{\textsc{3Dprinter} ($\Omega=1.22$, $\omega=59.4$)} & \multicolumn{4}{c}{\textsc{Chicken} ($\Omega=1.52$, $\omega=33.5$)}\\
Method &mPSNR$\uparrow$ &mSSIM$\uparrow$ &mLPIPS$\downarrow$ &PCK-T$\uparrow$ &mPSNR$\uparrow$ &mSSIM$\uparrow$ &mLPIPS$\downarrow$ &PCK-T$\uparrow$\\
\cmidrule(lr){1-1}\cmidrule(lr){2-5}\cmidrule(lr){6-9}
T-NeRF &$\mathbf{19.69}${\color{gray}\scriptsize($18.60$)} &$\mathbf{0.665}${\color{gray}\scriptsize($0.591$)} &$0.205${\color{gray}\scriptsize($0.238$)} &- &$\mathbf{25.54}${\color{gray}\scriptsize($24.41$)} &$\mathbf{0.802}${\color{gray}\scriptsize($0.764$)} &$0.131${\color{gray}\scriptsize($0.158$)} &- \\
NSFF &$16.89${\color{gray}\scriptsize($16.26$)} &$0.526${\color{gray}\scriptsize($0.426$)} &$\mathbf{0.443}${\color{gray}\scriptsize($0.492$)} &$0.797$ &$21.47${\color{gray}\scriptsize($20.72$)} &$0.671${\color{gray}\scriptsize($0.619$)} &$\mathbf{0.290}${\color{gray}\scriptsize($0.325$)} &$0.604$ \\
Nerfies &$19.67${\color{gray}\scriptsize($18.81$)} &$0.661${\color{gray}\scriptsize($0.588$)} &$0.148${\color{gray}\scriptsize($0.175$)} &$\mathbf{0.998}$ &$23.78${\color{gray}\scriptsize($22.71$)} &$0.784${\color{gray}\scriptsize($0.742$)} &$0.114${\color{gray}\scriptsize($0.142$)} &$0.978$ \\
HyperNeRF &$19.58${\color{gray}\scriptsize($18.73$)} &$0.656${\color{gray}\scriptsize($0.583$)} &$0.147${\color{gray}\scriptsize($0.175$)} &$0.994$ &$24.90${\color{gray}\scriptsize($23.88$)} &$0.792${\color{gray}\scriptsize($0.753$)} &$0.101${\color{gray}\scriptsize($0.125$)} &$\mathbf{1.000}$ \\
\bottomrule
\end{tabular}
}
\newline\newline
\resizebox{0.6\textwidth}{!}{  
\setlength{\tabcolsep}{4pt}
\begin{tabular}{l cccc}
\toprule
& \multicolumn{4}{c}{\textsc{Peel-banana} ($\Omega=0.33$, $\omega=33.8$)}\\
Method &mPSNR$\uparrow$ &mSSIM$\uparrow$ &mLPIPS$\downarrow$ &PCK-T$\uparrow$\\
\cmidrule(lr){1-1}\cmidrule(lr){2-5}
T-NeRF &$\mathbf{22.64}${\color{gray}\scriptsize($22.07$)} &$\mathbf{0.787}${\color{gray}\scriptsize($0.721$)} &$0.142${\color{gray}\scriptsize($0.185$)} &- \\
NSFF &$18.68${\color{gray}\scriptsize($18.62$)} &$0.613${\color{gray}\scriptsize($0.530$)} &$\mathbf{0.293}${\color{gray}\scriptsize($0.335$)} &$0.233$ \\
Nerfies &$19.97${\color{gray}\scriptsize($19.85$)} &$0.677${\color{gray}\scriptsize($0.609$)} &$0.161${\color{gray}\scriptsize($0.206$)} &$0.514$ \\
HyperNeRF &$21.34${\color{gray}\scriptsize($21.08$)} &$0.707${\color{gray}\scriptsize($0.641$)} &$0.135${\color{gray}\scriptsize($0.173$)} &$\mathbf{0.540}$ \\
\bottomrule
\end{tabular}
}
\vspace{0.5em}
\caption{\textbf{Per-scene breakdowns of the quantitative results on the Nerfies-HyperNeRF dataset.} Numbers in gray are calculated without using the co-visibility mask.
All models are trained under the non-teleporting setting.
}
\label{tab:app_breakdown_nfhn_results}
\end{table}

\begin{table}[h]
\centering
\tablestyle{4pt}{1.2}
\resizebox{\textwidth}{!}{  
\setlength{\tabcolsep}{4pt}
\begin{tabular}{l cccc cccc}
\toprule
& \multicolumn{4}{c}{\textsc{Apple} ($\Omega=0.75$, $\omega=3.8$)} & \multicolumn{4}{c}{\textsc{Block} ($\Omega=0.04$, $\omega=11.5$)}\\
Method &mPSNR$\uparrow$ &mSSIM$\uparrow$ &mLPIPS$\downarrow$ &PCK-T$\uparrow$ &mPSNR$\uparrow$ &mSSIM$\uparrow$ &mLPIPS$\downarrow$ &PCK-T$\uparrow$\\
\cmidrule(lr){1-1}\cmidrule(lr){2-5}\cmidrule(lr){6-9}
T-NeRF &$17.43${\color{gray}\scriptsize($15.98$)} &$0.728${\color{gray}\scriptsize($0.375$)} &$\mathbf{0.508}${\color{gray}\scriptsize($0.598$)} &- &$17.52${\color{gray}\scriptsize($17.15$)} &$0.669${\color{gray}\scriptsize($0.521$)} &$0.346${\color{gray}\scriptsize($0.449$)} &- \\
NSFF &$17.54${\color{gray}\scriptsize($16.50$)} &$0.750${\color{gray}\scriptsize($0.432$)} &$0.478${\color{gray}\scriptsize($0.548$)} &$\mathbf{0.599}$ &$16.61${\color{gray}\scriptsize($16.34$)} &$0.639${\color{gray}\scriptsize($0.494$)} &$0.389${\color{gray}\scriptsize($0.482$)} &$\mathbf{0.274}$ \\
Nerfies &$\mathbf{17.64}${\color{gray}\scriptsize($16.34$)} &$0.743${\color{gray}\scriptsize($0.411$)} &$0.478${\color{gray}\scriptsize($0.563$)} &$0.318$ &$\mathbf{17.54}${\color{gray}\scriptsize($17.35$)} &$\mathbf{0.670}${\color{gray}\scriptsize($0.528$)} &$0.331${\color{gray}\scriptsize($0.424$)} &$0.216$ \\
HyperNeRF &$16.47${\color{gray}\scriptsize($16.07$)} &$\mathbf{0.754}${\color{gray}\scriptsize($0.425$)} &$0.414${\color{gray}\scriptsize($0.505$)} &$0.132$ &$14.71${\color{gray}\scriptsize($14.93$)} &$0.606${\color{gray}\scriptsize($0.460$)} &$\mathbf{0.438}${\color{gray}\scriptsize($0.517$)} &$0.180$ \\
\bottomrule
\end{tabular}
}
\newline\newline
\resizebox{\textwidth}{!}{  
\setlength{\tabcolsep}{4pt}
\begin{tabular}{l cccc cccc}
\toprule
& \multicolumn{4}{c}{\textsc{Paper-windmill} ($\Omega=0.38$, $\omega=10.7$)} & \multicolumn{4}{c}{\textsc{Space-out} ($\Omega=0.13$, $\omega=6.4$)}\\
Method &mPSNR$\uparrow$ &mSSIM$\uparrow$ &mLPIPS$\downarrow$ &PCK-T$\uparrow$ &mPSNR$\uparrow$ &mSSIM$\uparrow$ &mLPIPS$\downarrow$ &PCK-T$\uparrow$\\
\cmidrule(lr){1-1}\cmidrule(lr){2-5}\cmidrule(lr){6-9}
T-NeRF &$\mathbf{17.55}${\color{gray}\scriptsize($17.55$)} &$0.367${\color{gray}\scriptsize($0.349$)} &$0.258${\color{gray}\scriptsize($0.268$)} &- &$17.71${\color{gray}\scriptsize($17.04$)} &$0.591${\color{gray}\scriptsize($0.521$)} &$\mathbf{0.377}${\color{gray}\scriptsize($0.438$)} &- \\
NSFF &$17.34${\color{gray}\scriptsize($17.35$)} &$0.378${\color{gray}\scriptsize($0.362$)} &$0.211${\color{gray}\scriptsize($0.218$)} &$0.113$ &$17.79${\color{gray}\scriptsize($17.25$)} &$0.622${\color{gray}\scriptsize($0.560$)} &$0.303${\color{gray}\scriptsize($0.359$)} &$0.812$ \\
Nerfies &$17.38${\color{gray}\scriptsize($17.39$)} &$\mathbf{0.382}${\color{gray}\scriptsize($0.366$)} &$0.209${\color{gray}\scriptsize($0.215$)} &$0.107$ &$\mathbf{17.93}${\color{gray}\scriptsize($18.10$)} &$0.605${\color{gray}\scriptsize($0.546$)} &$0.320${\color{gray}\scriptsize($0.369$)} &$\mathbf{0.859}$ \\
HyperNeRF &$14.94${\color{gray}\scriptsize($14.98$)} &$0.272${\color{gray}\scriptsize($0.254$)} &$\mathbf{0.348}${\color{gray}\scriptsize($0.361$)} &$\mathbf{0.163}$ &$17.65${\color{gray}\scriptsize($17.79$)} &$\mathbf{0.636}${\color{gray}\scriptsize($0.578$)} &$0.341${\color{gray}\scriptsize($0.390$)} &$0.598$ \\
\bottomrule
\end{tabular}
}
\newline\newline
\resizebox{\textwidth}{!}{  
\setlength{\tabcolsep}{4pt}
\begin{tabular}{l cccc cccc}
\toprule
& \multicolumn{4}{c}{\textsc{Spin} ($\Omega=0.15$, $\omega=7.9$)} & \multicolumn{4}{c}{\textsc{Teddy} ($\Omega=0.20$, $\omega=7.6$)}\\
Method &mPSNR$\uparrow$ &mSSIM$\uparrow$ &mLPIPS$\downarrow$ &PCK-T$\uparrow$ &mPSNR$\uparrow$ &mSSIM$\uparrow$ &mLPIPS$\downarrow$ &PCK-T$\uparrow$\\
\cmidrule(lr){1-1}\cmidrule(lr){2-5}\cmidrule(lr){6-9}
T-NeRF &$19.16${\color{gray}\scriptsize($18.17$)} &$0.567${\color{gray}\scriptsize($0.441$)} &$\mathbf{0.443}${\color{gray}\scriptsize($0.490$)} &- &$13.71${\color{gray}\scriptsize($13.32$)} &$\mathbf{0.570}${\color{gray}\scriptsize($0.331$)} &$0.429${\color{gray}\scriptsize($0.565$)} &- \\
NSFF &$18.38${\color{gray}\scriptsize($16.97$)} &$\mathbf{0.585}${\color{gray}\scriptsize($0.445$)} &$0.309${\color{gray}\scriptsize($0.380$)} &$\mathbf{0.177}$ &$13.65${\color{gray}\scriptsize($12.91$)} &$0.557${\color{gray}\scriptsize($0.302$)} &$0.372${\color{gray}\scriptsize($0.508$)} &$\mathbf{0.801}$ \\
Nerfies &$\mathbf{19.20}${\color{gray}\scriptsize($18.59$)} &$0.561${\color{gray}\scriptsize($0.436$)} &$0.325${\color{gray}\scriptsize($0.377$)} &$0.115$ &$\mathbf{13.97}${\color{gray}\scriptsize($13.91$)} &$0.568${\color{gray}\scriptsize($0.327$)} &$0.350${\color{gray}\scriptsize($0.479$)} &$0.775$ \\
HyperNeRF &$17.26${\color{gray}\scriptsize($16.52$)} &$0.540${\color{gray}\scriptsize($0.414$)} &$0.371${\color{gray}\scriptsize($0.437$)} &$0.083$ &$12.59${\color{gray}\scriptsize($12.78$)} &$0.537${\color{gray}\scriptsize($0.304$)} &$\mathbf{0.527}${\color{gray}\scriptsize($0.635$)} &$0.291$ \\
\bottomrule
\end{tabular}
}
\newline\newline
\resizebox{0.6\textwidth}{!}{  
\setlength{\tabcolsep}{4pt}
\begin{tabular}{l cccc}
\toprule
& \multicolumn{4}{c}{\textsc{Wheel} ($\Omega=0.03$, $\omega=58.5$)}\\
Method &mPSNR$\uparrow$ &mSSIM$\uparrow$ &mLPIPS$\downarrow$ &PCK-T$\uparrow$\\
\cmidrule(lr){1-1}\cmidrule(lr){2-5}
T-NeRF &$\mathbf{15.65}${\color{gray}\scriptsize($14.42$)} &$\mathbf{0.548}${\color{gray}\scriptsize($0.405$)} &$0.292${\color{gray}\scriptsize($0.363$)} &- \\
NSFF &$13.82${\color{gray}\scriptsize($13.19$)} &$0.458${\color{gray}\scriptsize($0.312$)} &$0.310${\color{gray}\scriptsize($0.366$)} &$0.394$ \\
Nerfies &$13.99${\color{gray}\scriptsize($13.35$)} &$0.455${\color{gray}\scriptsize($0.307$)} &$0.310${\color{gray}\scriptsize($0.366$)} &$\mathbf{0.408}$ \\
HyperNeRF &$14.59${\color{gray}\scriptsize($13.31$)} &$0.511${\color{gray}\scriptsize($0.359$)} &$\mathbf{0.331}${\color{gray}\scriptsize($0.402$)} &$0.346$ \\
\bottomrule
\end{tabular}
}
\vspace{0.5em}
\caption{\textbf{Per-scene breakdowns of the quantitative results on the proposed iPhone dataset.} Numbers in gray are calculated without using the co-visibility mask.}
\label{tab:app_breakdown_iphone_results}
\end{table}

We document the per-sequence quantitative performances of different models on both the Nerfies-HyperNeRF dataset (under non-teleporting setting) in Table~\ref{tab:app_breakdown_nfhn_results} and the proposed iPhone dataset in Table~\ref{tab:app_breakdown_iphone_results}.

\section{Additional results on novel-view synthesis}
\label{sec:app_nvs_results}
\providecommand\animage{}
\renewcommand{\animage}[2]{
    \frame{\includegraphics[width=\linewidth,clip,trim=#1]{figures/assets/nerfies_and_hypernerf_benchmark/#2}}
}
\providecommand\textimage{}
\renewcommand{\textimage}[4]{
	\frame{\begin{overpic}[width=\linewidth,clip,trim=#1]{figures/assets/nerfies_and_hypernerf_benchmark/#2}\put(0.3,920){\footnotesize\sethlcolor{black}\textcolor{white}{\hl{$#3/#4$}}}\end{overpic}}
}
\begin{figure}[t!]
    \setlength{\tabcolsep}{0.8pt}
    \renewcommand{\arraystretch}{0.5}
    \begin{tabularx}{\textwidth}{@{}c*{6}{C}@{}}
        \makebox[20pt]{\raisebox{40pt}{\rotatebox[origin=c]{90}{\textsc{Broom}}}}\hspace{-6pt} &
        \textimage{0 68 0 13}{nerfies/broom/174_src.png}{2.57}{60.36} &
        \animage{0 68 0 13}{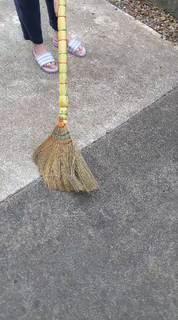} &
        \animage{0 68 0 13}{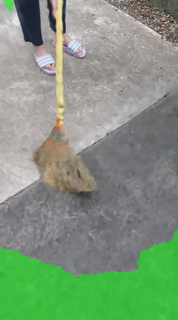} &
        \animage{0 68 0 13}{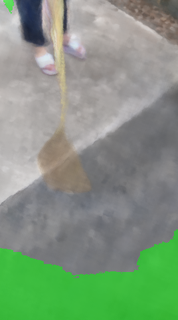} &
        \animage{0 68 0 13}{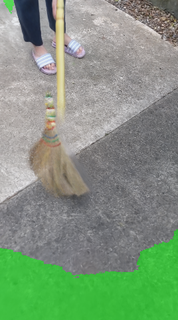} &
        \animage{0 68 0 13}{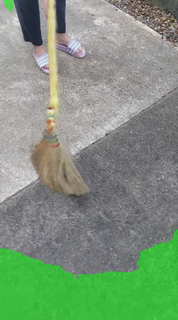}
        \\
        \makebox[13pt]{\raisebox{40pt}{\rotatebox[origin=c]{90}{\textsc{Curls}}}}\hspace{-6pt} &
        \textimage{0 0 0 0}{nerfies/curls/35_src.png}{0.90}{118.65} &
        \animage{0 0 0 0}{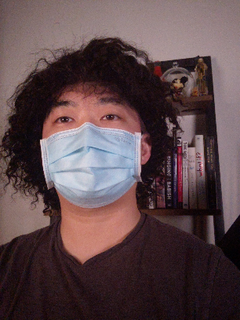} &
        \animage{0 0 0 0}{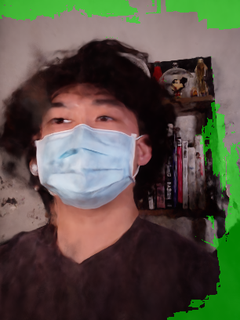} &
        \animage{0 0 0 0}{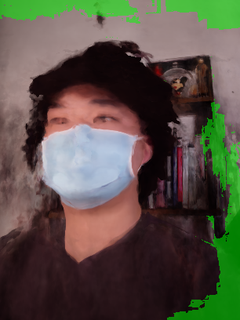} &
        \animage{0 0 0 0}{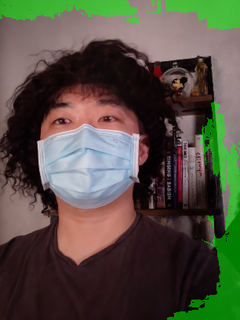} &
        \animage{0 0 0 0}{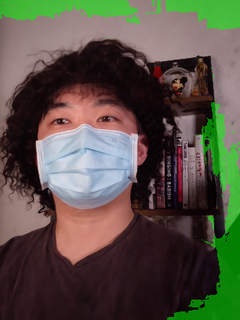}
        \\
        \makebox[13pt]{\raisebox{40pt}{\rotatebox[origin=c]{90}{\textsc{Tail}}}}\hspace{-6pt} &
        \textimage{0 68 0 13}{nerfies/tail/121_src.png}{1.31}{28.63} &
        \animage{0 68 0 13}{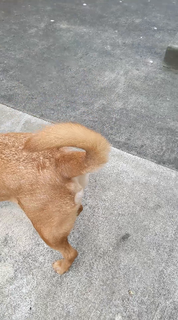} &
        \animage{0 68 0 13}{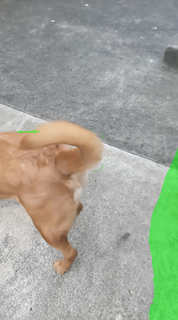} &
        \animage{0 68 0 13}{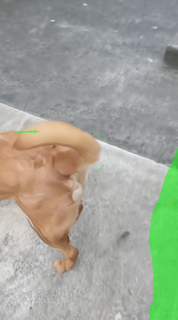} &
        \animage{0 68 0 13}{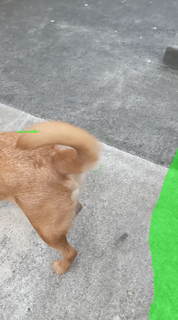} &
        \animage{0 68 0 13}{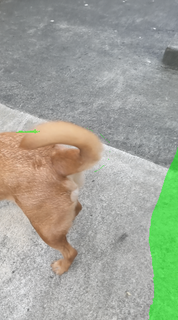}
        \\
        \makebox[13pt]{\raisebox{40pt}{\rotatebox[origin=c]{90}{\textsc{Toby Sit}}}}\hspace{-6pt} &
        \textimage{0 68 0 13}{nerfies/toby-sit/121_src.png}{1.28}{26.43} &
        \animage{0 68 0 13}{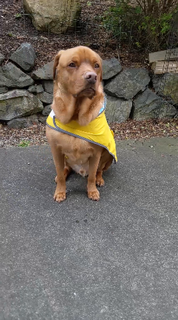} &
        \animage{0 68 0 13}{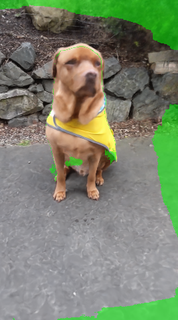} &
        \animage{0 68 0 13}{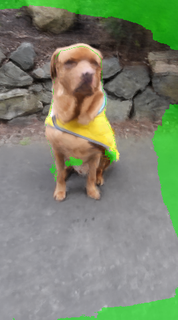} &
        \animage{0 68 0 13}{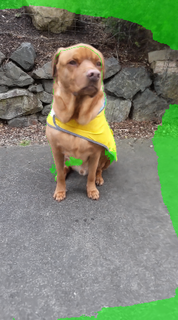} &
        \animage{0 68 0 13}{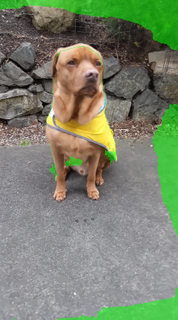}
        \\
        \makebox[20pt]{\raisebox{40pt}{\rotatebox[origin=c]{90}{\textsc{3D Printer}}}}\hspace{-6pt} &
        \textimage{0 14 0 66}{hypernerf/vrig-3dprinter/182_src.png}{1.14}{59.40} &
        \animage{0 14 0 66}{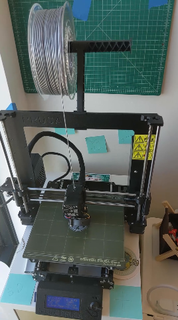} &
        \animage{0 14 0 66}{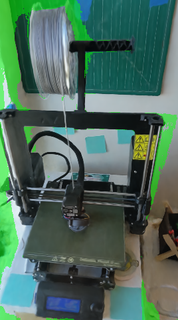} &
        \animage{0 14 0 66}{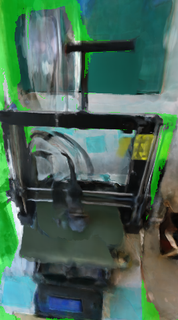} &
        \animage{0 14 0 66}{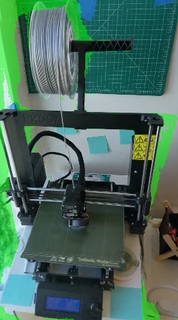} &
        \animage{0 14 0 66}{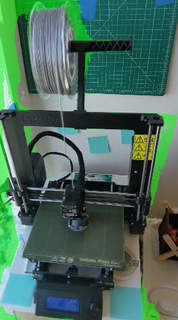}
        \\
        \makebox[20pt]{\raisebox{40pt}{\rotatebox[origin=c]{90}{\textsc{Chicken}}}}\hspace{-6pt} &
        \textimage{0 68 0 13}{hypernerf/vrig-chicken/125_src.png}{1.52}{33.46} &
        \animage{0 68 0 13}{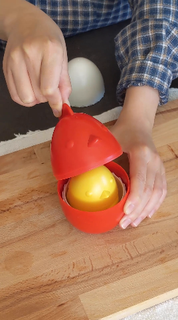} &
        \animage{0 68 0 13}{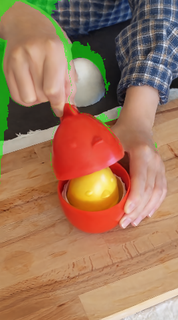} &
        \animage{0 68 0 13}{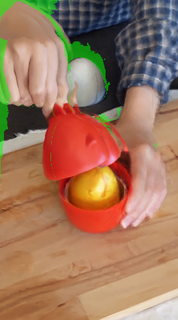} &
        \animage{0 68 0 13}{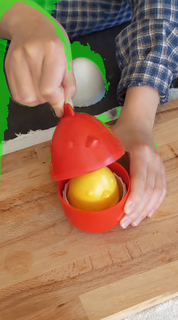} &
        \animage{0 68 0 13}{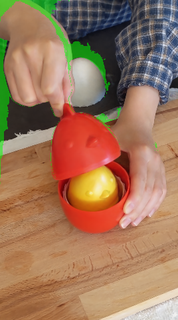}
        \\
        \makebox[20pt]{\raisebox{40pt}{\rotatebox[origin=c]{90}{\textsc{Peel Banana}}}}\hspace{-6pt} &
        \textimage{0 41 0 40}{hypernerf/vrig-peel-banana/372_src.png}{0.33}{33.77} &
        \animage{0 41 0 40}{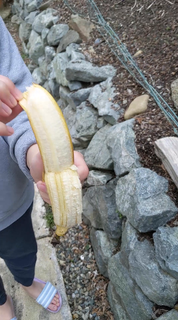} &
        \animage{0 41 0 40}{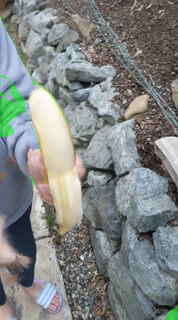} &
        \animage{0 41 0 40}{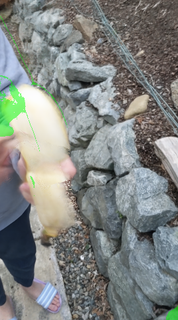} &
        \animage{0 41 0 40}{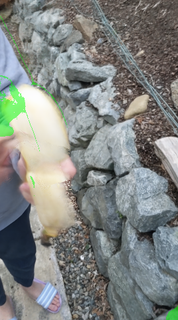} &
        \animage{0 41 0 40}{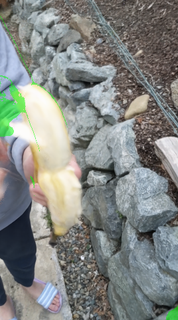}
        \\
        [2pt] &
        {\small Training view} &
        {\small Test view} &
        {\small T-NeRF} &
        {\small NSFF~\cite{li2021neural}} &
        {\small Nerfies~\cite{park2021nerfies}} &
        {\small HyperNeRF~\cite{park2021hypernerf}}
    \end{tabularx}
    \vspace{-4pt}
    \caption{
        \textbf{Additional qualitative results on the Nerfies-HyperNeRF dataset without camera teleportation.}
        $\Omega/\omega$ metrics of the input sequence are shown on the top-left.
    }
    \vspace{-1em}
    \label{fig:app_nvs_nfhn_results}
\end{figure}

\providecommand\animage{}
\renewcommand{\animage}[2]{
    \frame{\includegraphics[width=\linewidth,clip,trim=#1]{figures/assets/iphone_benchmark/#2}}
}
\providecommand\textimage{}
\renewcommand{\textimage}[4]{
	\frame{\begin{overpic}[width=\linewidth,clip,trim=#1]{figures/assets/iphone_benchmark/#2}\put(0.3,905){\footnotesize\sethlcolor{black}\textcolor{white}{\hl{$#3/#4$}}}\end{overpic}}
}
\begin{figure}[t!]
    \setlength{\tabcolsep}{0.8pt}
    \renewcommand{\arraystretch}{0.5}
    \begin{tabularx}{\textwidth}{@{}c*{7}{C}@{}}
        \makebox[20pt]{\raisebox{35pt}{\rotatebox[origin=c]{90}{\textsc{Apple}}}}\hspace{-6pt} &
        \textimage{0 0 0 0}{iphone2/apple/200_src.png}{0.75}{3.79} &
        \animage{0 0 0 0}{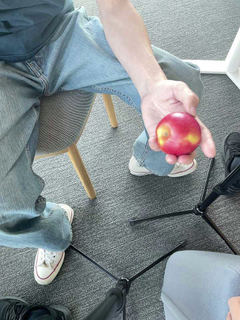} &
        \animage{0 0 0 0}{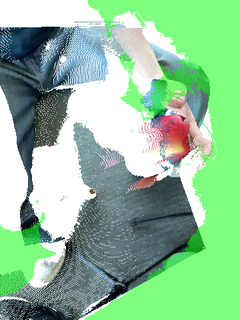} &
        \animage{0 0 0 0}{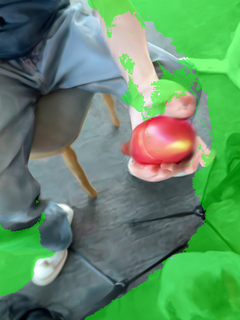} &
        \animage{0 0 0 0}{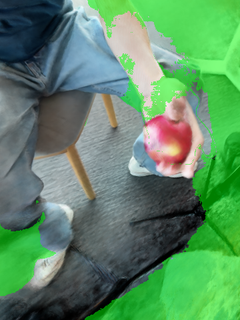} &
        \animage{0 0 0 0}{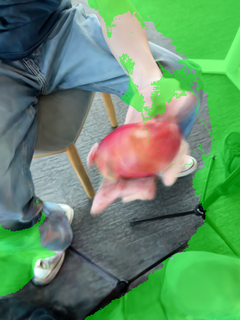} &
        \animage{0 0 0 0}{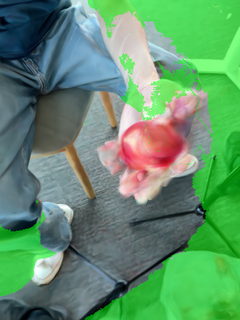}
        \\
        \makebox[20pt]{\raisebox{35pt}{\rotatebox[origin=c]{90}{\textsc{Block}}}}\hspace{-6pt} &
        \textimage{0 0 0 0}{iphone2/block/49_src.png}{0.75}{3.79} &
        \animage{0 0 0 0}{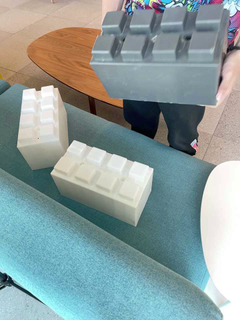} &
        \animage{0 0 0 0}{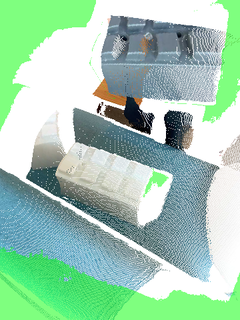} &
        \animage{0 0 0 0}{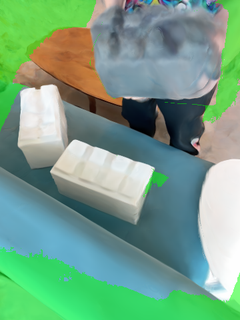} &
        \animage{0 0 0 0}{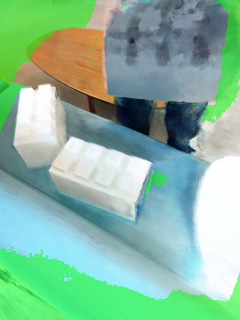} &
        \animage{0 0 0 0}{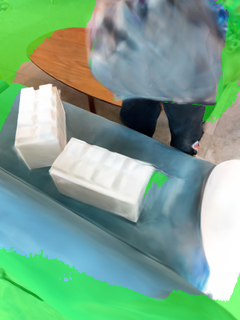} &
        \animage{0 0 0 0}{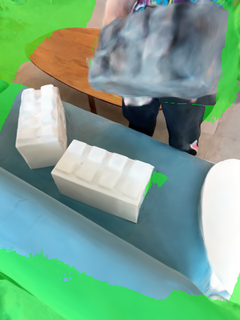}
        \\
        \makebox[20pt]{\raisebox{32pt}{\rotatebox[origin=c]{90}{\footnotesize\textsc{Paper Windmill}}}}\hspace{-6pt} &
        \textimage{0 0 0 0}{iphone2/paper-windmill/84_src.png}{0.38}{10.71} &
        \animage{0 0 0 0}{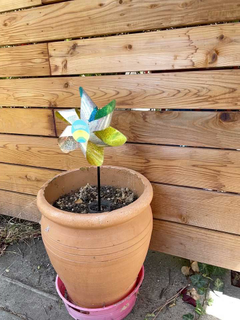} &
        \animage{0 0 0 0}{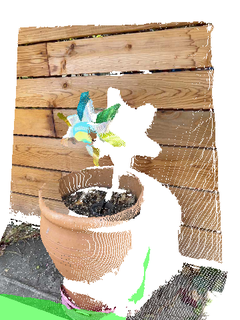} &
        \animage{0 0 0 0}{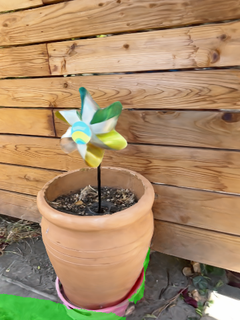} &
        \animage{0 0 0 0}{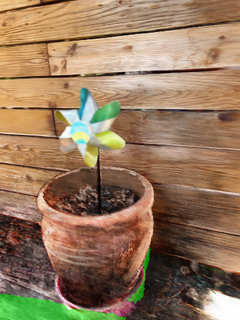} &
        \animage{0 0 0 0}{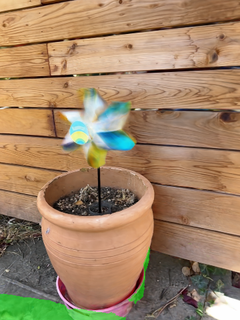} &
        \animage{0 0 0 0}{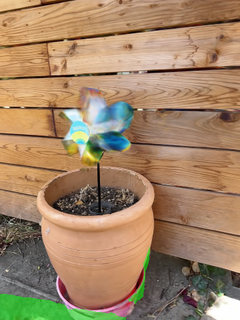}
        \\
        \makebox[20pt]{\raisebox{35pt}{\rotatebox[origin=c]{90}{\textsc{Space Out}}}}\hspace{-6pt} &
        \textimage{10 8 10 8}{iphone2/space-out/207_src.png}{0.13}{6.37} &
        \animage{10 8 10 8}{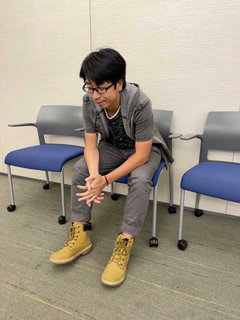} &
        \animage{10 8 10 8}{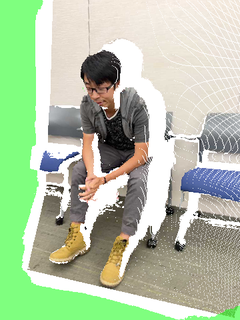} &
        \animage{10 8 10 8}{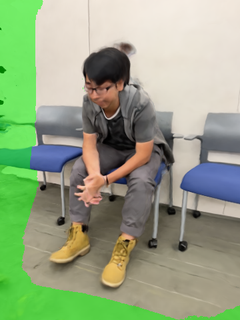} &
        \animage{10 8 10 8}{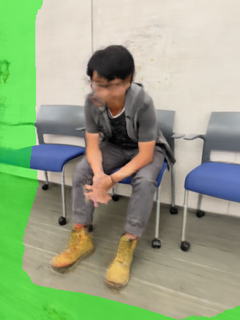} &
        \animage{10 8 10 8}{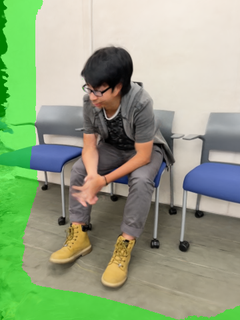} &
        \animage{10 8 10 8}{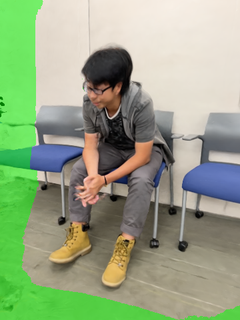}
        \\
        \makebox[20pt]{\raisebox{35pt}{\rotatebox[origin=c]{90}{\textsc{Spin}}}}\hspace{-6pt} &
        \textimage{0 0 0 0}{iphone2/spin/53_src.png}{0.15}{7.86} &
        \animage{0 0 0 0}{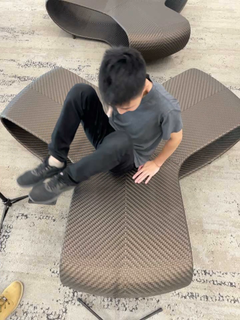} &
        \animage{0 0 0 0}{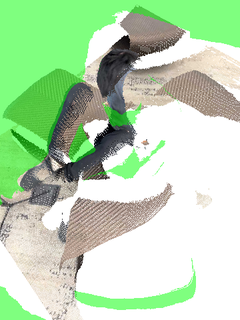} &
        \animage{0 0 0 0}{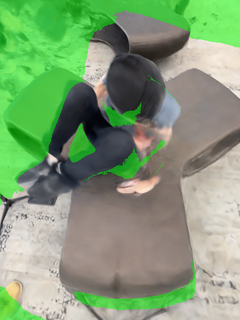} &
        \animage{0 0 0 0}{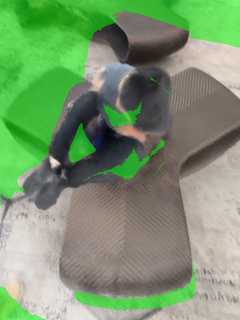} &
        \animage{0 0 0 0}{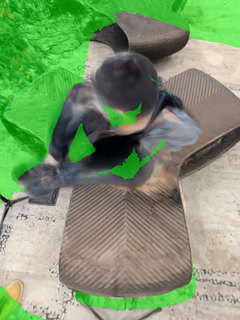} &
        \animage{0 0 0 0}{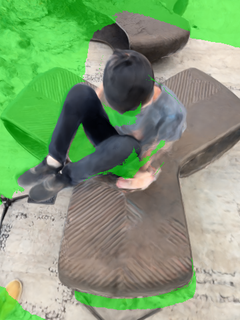}
        \\
        \makebox[20pt]{\raisebox{35pt}{\rotatebox[origin=c]{90}{\textsc{Teddy}}}}\hspace{-6pt} &
        \textimage{0 0 0 0}{iphone2/teddy/0_src.png}{0.20}{7.62} &
        \animage{0 0 0 0}{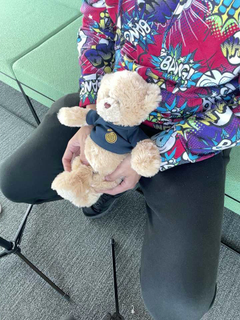} &
        \animage{0 0 0 0}{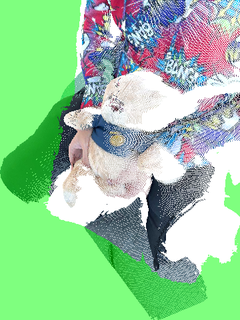} &
        \animage{0 0 0 0}{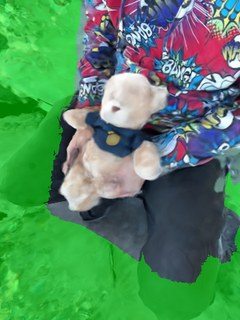} &
        \animage{0 0 0 0}{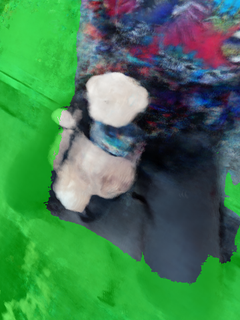} &
        \animage{0 0 0 0}{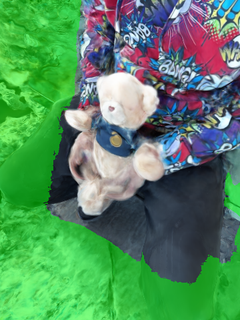} &
        \animage{0 0 0 0}{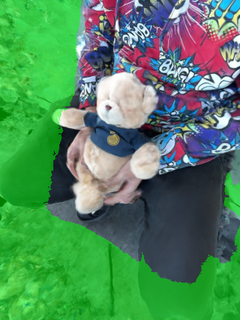}
        \\
        \makebox[20pt]{\raisebox{35pt}{\rotatebox[origin=c]{90}{\textsc{Wheel}}}}\hspace{-6pt} &
        \textimage{0 0 0 0}{iphone2/wheel/150_src.png}{0.03}{58.45} &
        \animage{0 0 0 0}{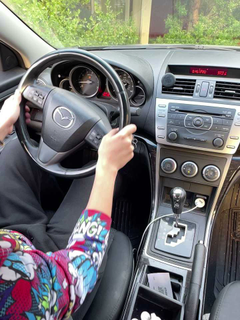} &
        \animage{0 0 0 0}{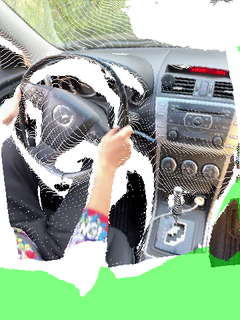} &
        \animage{0 0 0 0}{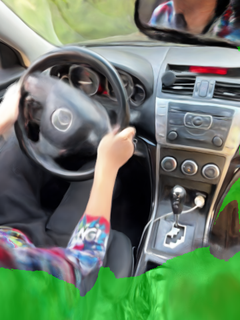} &
        \animage{0 0 0 0}{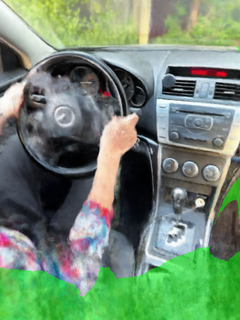} &
        \animage{0 0 0 0}{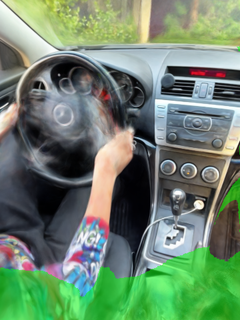} &
        \animage{0 0 0 0}{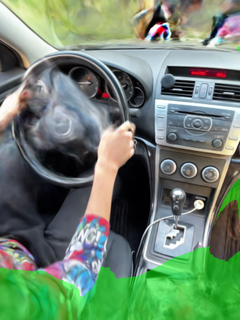}
        \\
        [2pt] &
        {\small Training view} &
        {\small Test view} &
        {\small Depth re-proj.} &
        {\small T-NeRF++} &
        {\small NSFF~\cite{li2020nsff}} &
        {\small Nerfies++} &
        {\small HyperNeRF++}
    \end{tabularx}
    \vspace{-4pt}
    \caption{
        \textbf{Additional qualitative results on the multi-camera captures from the proposed iPhone dataset.}
        $\Omega/\omega$ metrics of the input sequence are shown on the top-left.
        The models shown here are trained with all the additional regularizations (\textsc{+B+D+S}) except NSFF.
    }
    \vspace{-1em}
    \label{fig:app_nvs_iphone_mv_results}
\end{figure}

\providecommand\animage{}
\renewcommand{\animage}[2]{
    \frame{\includegraphics[width=\linewidth,clip,trim=#1]{figures/assets/appendix/app_nvs_iphone_sv_results/#2}}
}
\providecommand\textimage{}
\renewcommand{\textimage}[4]{
	\frame{\begin{overpic}[width=\linewidth,clip,trim=#1]{figures/assets/appendix/app_nvs_iphone_sv_results/#2}\put(0.3,915){\footnotesize\sethlcolor{black}\textcolor{white}{\hl{$#3/#4$}}}\end{overpic}}
}
\begin{figure}[t!]
    \setlength{\tabcolsep}{0.8pt}
    \renewcommand{\arraystretch}{0.5}
    \begin{tabularx}{\textwidth}{@{}c*{6}{C}@{}}
        \makebox[20pt]{\raisebox{38pt}{\rotatebox[origin=c]{90}{\textsc{Creeper}}}}\hspace{-6pt} &
        \textimage{0 0 0 0}{iphone2/creeper/5_src.png}{0.23}{14.05} &
        \animage{0 0 0 0}{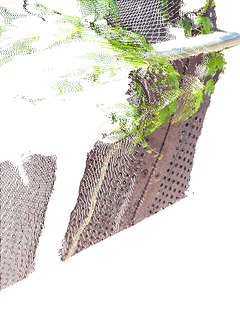} &
        \animage{0 0 0 0}{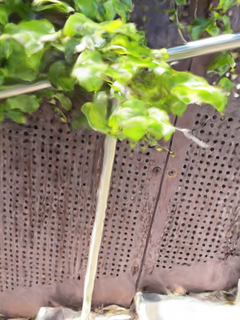} &
        \animage{0 0 0 0}{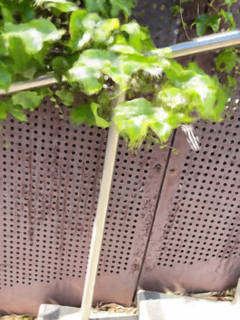} &
        \animage{0 0 0 0}{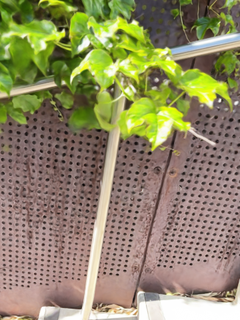} &
        \animage{0 0 0 0}{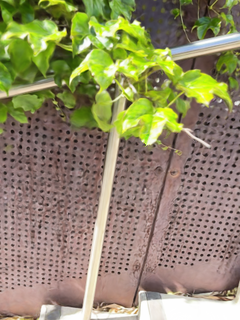}
        \\
        \makebox[20pt]{\raisebox{38pt}{\rotatebox[origin=c]{90}{\textsc{Backpack}}}}\hspace{-6pt} &
        \textimage{0 0 0 0}{iphone2/backpack/5_src.png}{0.26}{5.59} &
        \animage{0 0 0 0}{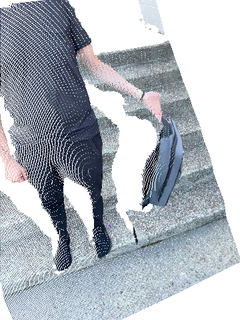} &
        \animage{0 0 0 0}{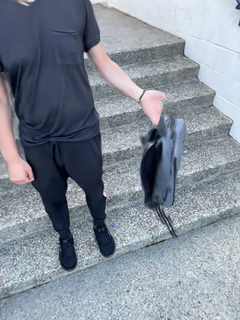} &
        \animage{0 0 0 0}{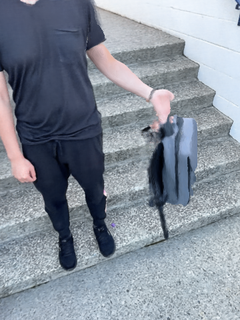} &
        \animage{0 0 0 0}{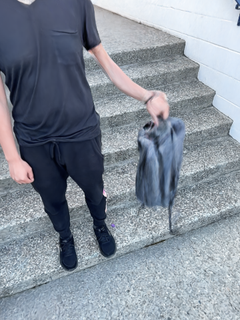} &
        \animage{0 0 0 0}{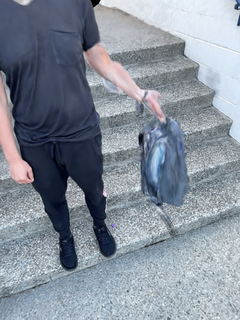}
        \\
        \makebox[20pt]{\raisebox{38pt}{\rotatebox[origin=c]{90}{\textsc{Handwavy}}}}\hspace{-6pt} &
        \textimage{0 0 0 0}{iphone2/handwavy/5_src.png}{0.05}{13.66} &
        \animage{0 0 0 0}{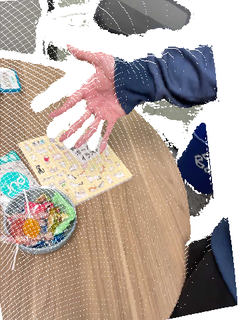} &
        \animage{0 0 0 0}{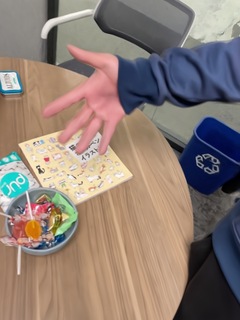} &
        \animage{0 0 0 0}{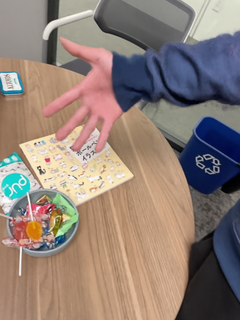} &
        \animage{0 0 0 0}{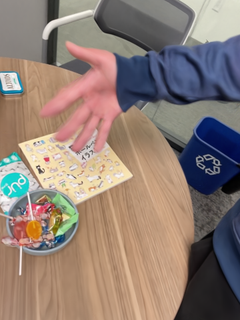} &
        \animage{0 0 0 0}{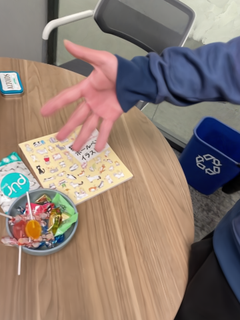}
        \\
        \makebox[20pt]{\raisebox{38pt}{\rotatebox[origin=c]{90}{\textsc{Haru}}}}\hspace{-6pt} &
        \textimage{70 0 70 0}{iphone2/haru-sit/5_src.png}{0.30}{30.32} &
        \animage{70 0 70 0}{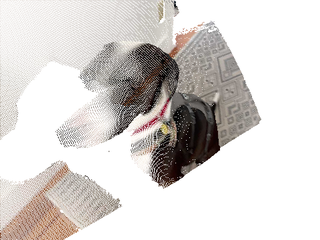} &
        \animage{70 0 70 0}{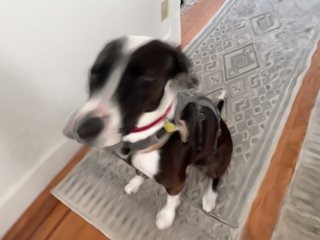} &
        \animage{70 0 70 0}{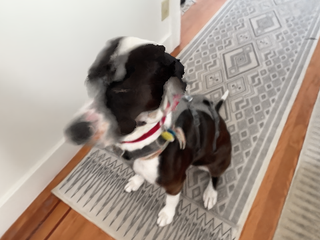} &
        \animage{70 0 70 0}{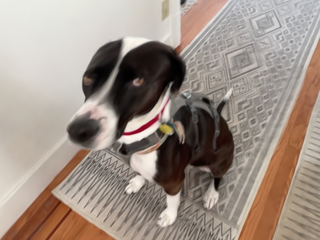} &
        \animage{70 0 70 0}{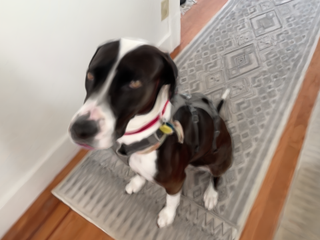}
        \\
        \makebox[20pt]{\raisebox{38pt}{\rotatebox[origin=c]{90}{\textsc{Mochi}}}}\hspace{-6pt} &
        \textimage{0 0 0 0}{iphone2/mochi-high-five/5_src.png}{0.07}{14.49} &
        \animage{0 0 0 0}{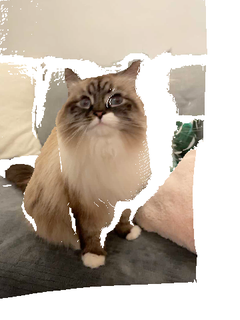} &
        \animage{0 0 0 0}{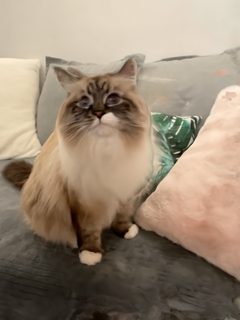} &
        \animage{0 0 0 0}{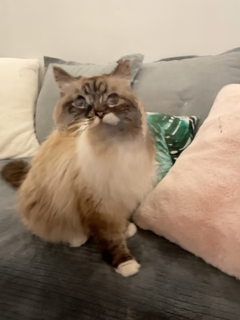} &
        \animage{0 0 0 0}{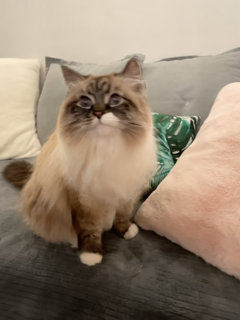} &
        \animage{0 0 0 0}{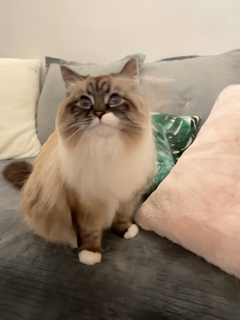}
        \\
        \makebox[20pt]{\raisebox{38pt}{\rotatebox[origin=c]{90}{\textsc{Pillow}}}}\hspace{-6pt} &
        \textimage{0 0 0 0}{iphone2/pillow/1_src.png}{0.06}{13.19} &
        \animage{0 0 0 0}{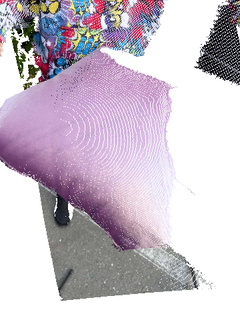} &
        \animage{0 0 0 0}{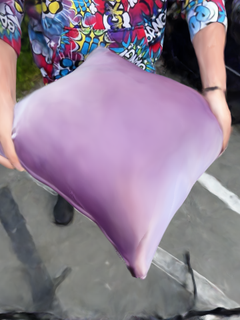} &
        \animage{0 0 0 0}{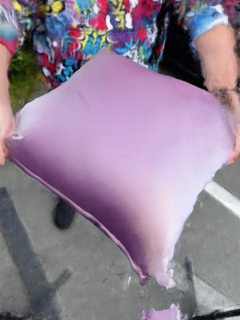} &
        \animage{0 0 0 0}{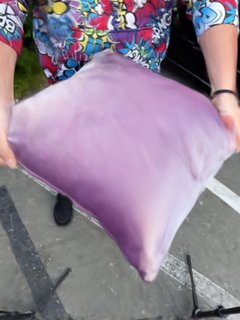} &
        \animage{0 0 0 0}{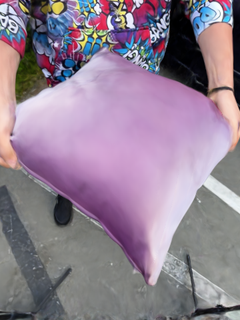}
        \\
        \makebox[20pt]{\raisebox{38pt}{\rotatebox[origin=c]{90}{\textsc{Sriracha}}}}\hspace{-6pt} &
        \textimage{0 0 0 0}{iphone2/sriracha-tree/5_src.png}{0.18}{18.56} &
        \animage{0 0 0 0}{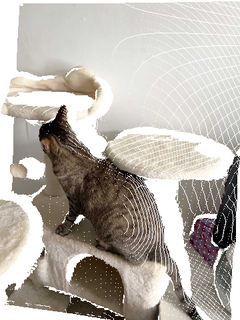} &
        \animage{0 0 0 0}{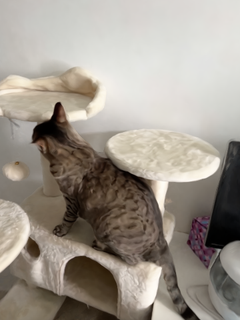} &
        \animage{0 0 0 0}{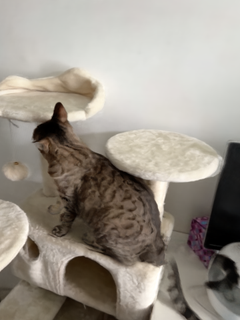} &
        \animage{0 0 0 0}{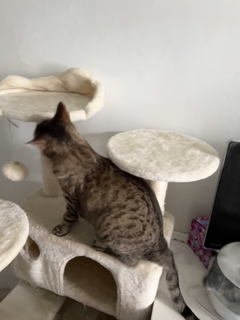} &
        \animage{0 0 0 0}{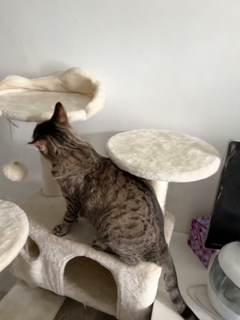}
        \\
        [2pt] &
        {\small Training view} &
        {\small Depth re-proj.} &
        {\small T-NeRF++} &
        {\small NSFF~\cite{li2020nsff}} &
        {\small Nerfies++} &
        {\small HyperNeRF++}
    \end{tabularx}
    \vspace{-4pt}
    \caption{
        \textbf{Additional qualitative results on the single-camera captures from the proposed iPhone dataset.}
        $\Omega/\omega$ metrics of the input sequence are shown on the top-left.
        The models shown here are trained with all the additional regularizations (\textsc{+B+D+S}) except NSFF.
        We re-render the scene from the first viewpoint in each sequence.
        Note that there are no ground-truth validation frames.
    }
    \vspace{-1em}
    \label{fig:app_nvs_iphone_sv_results}
\end{figure}

\providecommand\animage{}
\renewcommand{\animage}[2]{
    \frame{\includegraphics[width=\linewidth,clip,trim=#1]{figures/assets/appendix/app_nvs_full_frame_results/#2}}
}
\providecommand\textimage{}
\renewcommand{\textimage}[4]{
	\frame{\begin{overpic}[width=\linewidth,clip,trim=#1]{figures/assets/appendix/app_nvs_full_frame_results/#2}\put(0,918){\footnotesize\sethlcolor{black}\textcolor{white}{\hl{$#3/#4$}}}\end{overpic}}
}
\begin{figure}[t!]
    \centering
    \begin{minipage}{\linewidth}
        \setlength{\tabcolsep}{0.8pt}
        \renewcommand{\arraystretch}{0.5}
        \begin{tabularx}{\textwidth}{@{}c*{6}{C}@{}}
            \multirow{2}{*}[25pt]{\makebox[20pt]{\raisebox{0pt}{\rotatebox[origin=c]{90}{\textsc{3D Printer}}}}\hspace{-6pt}} &
            \textimage{0 40 0 40}{hypernerf/vrig-3dprinter/62_src.png}{1.22}{59.40} &
            \animage{0 40 0 40}{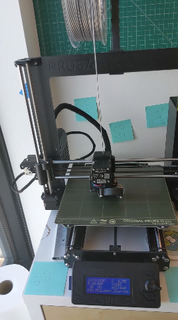} &
            \animage{0 40 0 40}{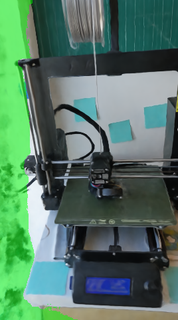} &
            \animage{0 40 0 40}{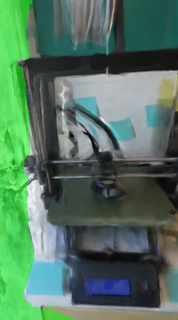} &
            \animage{0 40 0 40}{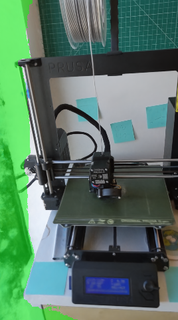} &
            \animage{0 40 0 40}{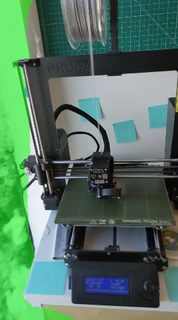}
            \\
            &
            &
            &
            \animage{0 40 0 40}{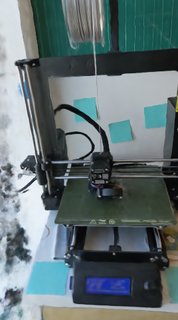} &
            \animage{0 40 0 40}{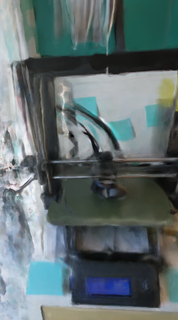} &
            \animage{0 40 0 40}{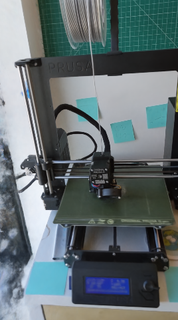} &
            \animage{0 40 0 40}{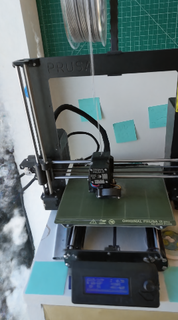}
            \\
            \multirow{2}{*}[20pt]{\makebox[20pt]{\raisebox{0pt}{\rotatebox[origin=c]{90}{\textsc{Chicken}}}}\hspace{-6pt}} &
            \textimage{0 40 0 40}{hypernerf/vrig-chicken/100_src.png}{1.52}{33.46} &
            \animage{0 40 0 40}{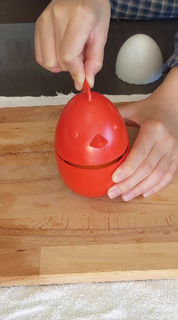} &
            \animage{0 40 0 40}{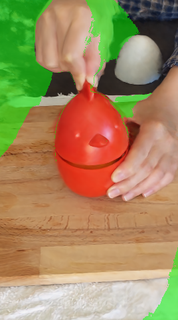} &
            \animage{0 40 0 40}{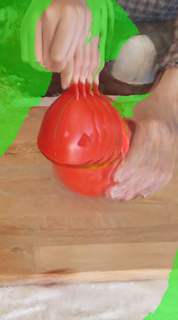} &
            \animage{0 40 0 40}{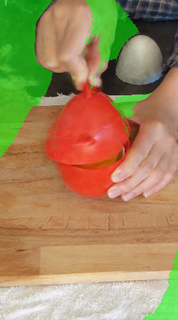} &
            \animage{0 40 0 40}{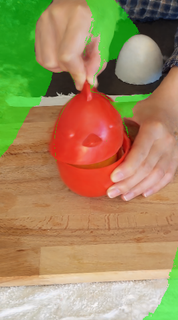}
            \\
            &
            &
            &
            \animage{0 40 0 40}{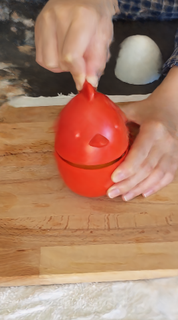} &
            \animage{0 40 0 40}{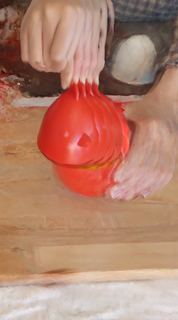} &
            \animage{0 40 0 40}{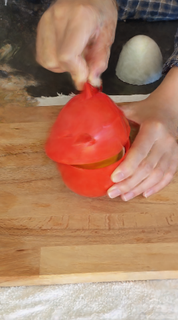} &
            \animage{0 40 0 40}{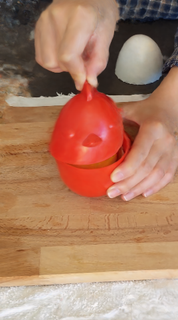}
            \\
            \multirow{2}{*}[10pt]{\makebox[20pt]{\raisebox{0pt}{\rotatebox[origin=c]{90}{\textsc{Spin}}}}\hspace{-6pt}} &
            \textimage{0 0 0 0}{iphone2/spin/0_src.png}{0.15}{7.86} &
            \animage{0 0 0 0}{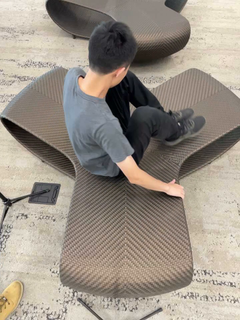} &
            \animage{0 0 0 0}{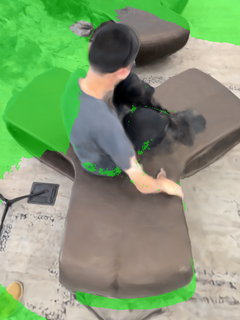} &
            \animage{0 0 0 0}{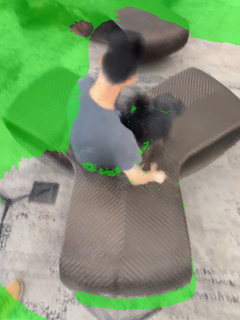} &
            \animage{0 0 0 0}{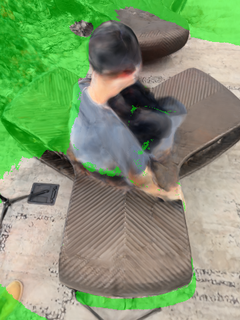} &
            \animage{0 0 0 0}{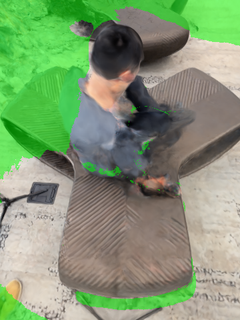}
            \\
            &
            &
            &
            \animage{0 0 0 0}{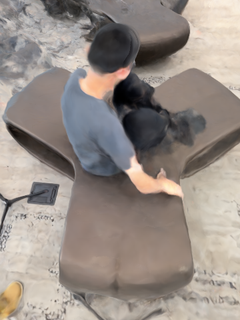} &
            \animage{0 0 0 0}{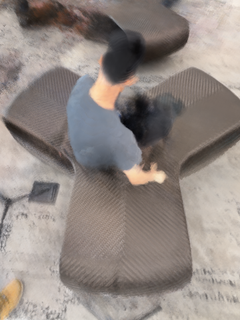} &
            \animage{0 0 0 0}{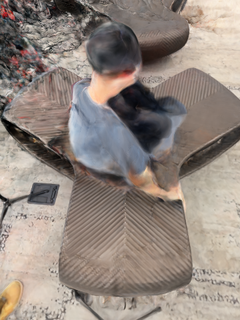} &
            \animage{0 0 0 0}{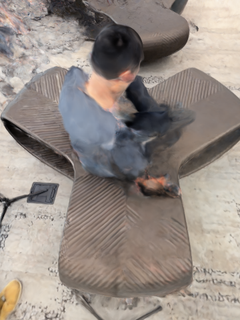}
            \\
            [2pt] &
            {\small Training view} &
            {\small Test view} &
            {\small T-NeRF} &
            {\small NSFF~\cite{li2020nsff}} &
            {\small Nerfies~\cite{park2021nerfies}} &
            {\small HyperNeRF~\cite{park2021hypernerf}}
        \end{tabularx}
    \end{minipage}
    \vspace{-2pt}
    \caption{
        \textbf{Additional qualitative results on the full image rendering on both the Nerfies-HyperNeRF dataset and the proposed iPhone dataset.}
        $\Omega/\omega$ metrics of the input sequence are shown on the top-left.
        All models are trained under non-teleporting setting.
        Note that the models on the proposed iPhone dataset are trained with additional regularizations (\textsc{+B+D+S}) except NSFF.
        For every two rows, we show the results with and without applying co-visibility mask.
        All models are not able to reconstruct the unseen regions.
    }
    \label{fig:app_nvs_full_frame_results}
\end{figure}

We provide additional novel-view synthesis qualitative results under the non-teleporting setting.
In Figure~\ref{fig:app_nvs_nfhn_results}, we show qualitative results on the Nerfies-HyperNeRF dataset.
In Figure~\ref{fig:app_nvs_iphone_mv_results}, we show qualitative results on the multi-camera captures from the proposed iPhone dataset. 
All models except NSFF~\cite{li2020nsff} are trained with all the additional regularizations that we find helpful through ablation, denoted with ``++'' to distinguish with the original models.
In Figure~\ref{fig:app_nvs_iphone_sv_results}, we show qualitative results on the single-camera captures from the proposed iPhone dataset.
We render novel views using the camera pose from the first captured frame.
Finally, in Figure~\ref{fig:app_nvs_full_frame_results}, we show the rendering results with and without co-visibility mask applied.

\section{Additional results on inferred correspondence}
\label{sec:app_corr_results}
\begin{table}[t!]
\centering
\tablestyle{4pt}{1.2}
\resizebox{0.9\textwidth}{!}{%
\setlength{\tabcolsep}{4pt}
\begin{tabular}{l ccccccc c}
\toprule
Method & 
\textsc{Creeper} &
\textsc{Backpack} &
\textsc{Handwavy} &
\textsc{Haru} &
\textsc{Mochi} &
\textsc{Pillow} &
\textsc{Sriracha} &
\textsc{Mean}
\\
\cmidrule(lr){1-1}\cmidrule(lr){2-8}\cmidrule(lr){9-9}
NSFF~\cite{li2020nsff} &
$0.560$&	$0.269$&	$0.178$&	$0.699$&	$0.624$&	$0.154$&	$0.616$&	$0.443$
\\
Nerfies++ &
$\mathbf{0.708}$&	$\mathbf{0.329}$&	$0.685$&	$\mathbf{0.942}$&	$\mathbf{0.908}$&	$0.575$&	$\mathbf{0.737}$& $\mathbf{0.698}$
\\
HyperNeRF++ &
$0.702$&	$0.260$&	$\mathbf{0.708}$&	$0.817$&	$0.891$&	$\mathbf{0.602}$&	$0.617$&	$0.657$
\\
\bottomrule
\end{tabular}%
}
\vspace{0.5em}
\caption{
    \textbf{Additional quantitative results of the PCK-T evaluation on the single-camera captures from the proposed iPhone dataset.}
    The correspondence evaluation is applicable when multi-camera validation is not available.
    All numbers are computed with $\alpha = 0.05$.
}
\label{tab:app_corr_iphone_sv}
\end{table}
\providecommand\animage{}
\renewcommand{\animage}[2]{
    \frame{\includegraphics[width=\linewidth,clip,trim=#1]{figures/assets/correspondences/#2}}
}
\providecommand\textimage{}
\renewcommand{\textimage}[4]{
	\frame{\begin{overpic}[width=\linewidth,clip,trim=#1]{figures/assets/correspondences/#2}\put(0,923){\footnotesize\sethlcolor{black}\textcolor{white}{\hl{$#3/#4$}}}\end{overpic}}
}
\providecommand\cbar{}
\renewcommand{\cbar}[1]{
	\raisebox{-442pt}[0pt][0pt]{\includegraphics[height=540pt,bb=-1 0 74 1398]{figures/assets/correspondences/#1}}
}
\begin{figure}[t!]
    \setlength{\tabcolsep}{0.8pt}
    \renewcommand{\arraystretch}{0.5}
    \begin{tabularx}{\textwidth}{@{}c*{5}{C}c@{}}
        \makebox[20pt]{\raisebox{45pt}{\rotatebox[origin=c]{90}{\textsc{Broom}}}}\hspace{-6pt} &
        \textimage{0 41 0 40}{nerfies/broom/32_kpt_src.png}{2.57}{40.36} &
        \animage{0 41 0 40}{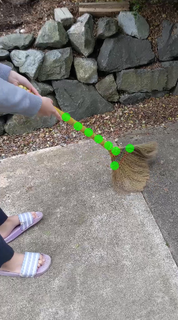} &
        \animage{0 41 0 40}{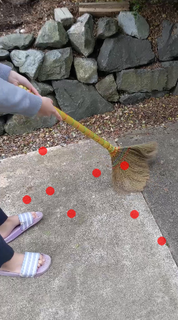} &
        \animage{0 41 0 40}{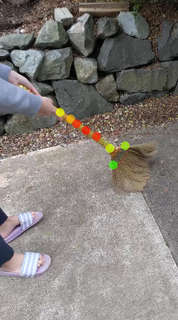} &
        \animage{0 41 0 40}{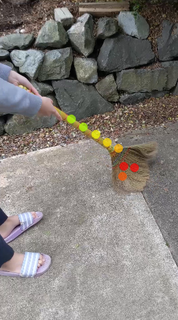} &
        \multirow{6}{*}{\cbar{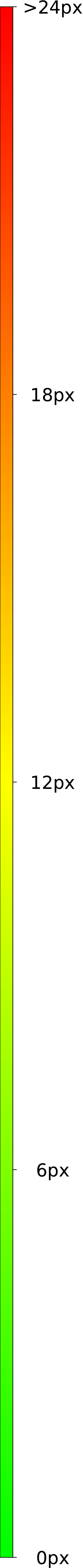}}
        \\
        \makebox[20pt]{\raisebox{45pt}{\rotatebox[origin=c]{90}{\textsc{Tail}}}}\hspace{-6pt} &
        \textimage{0 41 0 40}{nerfies/tail/32_kpt_src.png}{1.31}{28.41} &
        \animage{0 41 0 40}{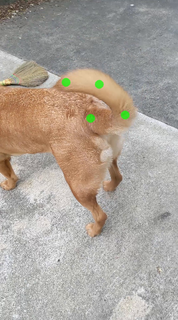} &
        \animage{0 41 0 40}{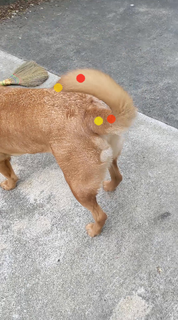} &
        \animage{0 41 0 40}{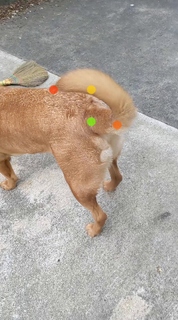} &
        \animage{0 41 0 40}{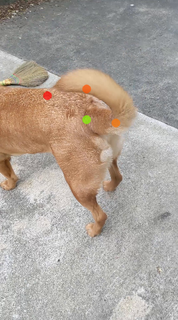}
        \\
        \makebox[20pt]{\raisebox{45pt}{\rotatebox[origin=c]{90}{\textsc{Toby Sit}}}}\hspace{-6pt} &
        \textimage{0 41 0 40}{nerfies/toby-sit/32_kpt_src.png}{1.28}{26.43} &
        \animage{0 41 0 40}{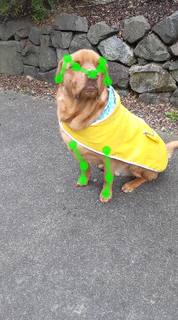} &
        \animage{0 41 0 40}{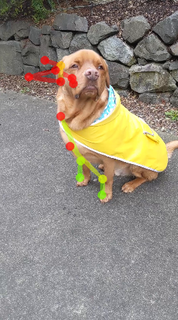} &
        \animage{0 41 0 40}{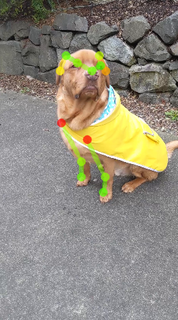} &
        \animage{0 41 0 40}{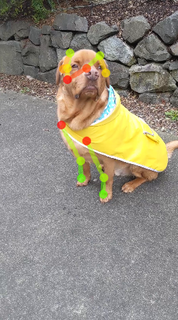}
        \\
        \makebox[20pt]{\raisebox{45pt}{\rotatebox[origin=c]{90}{\textsc{3D Printer}}}}\hspace{-6pt} &
        \textimage{0 41 0 40}{hypernerf/vrig-3dprinter/32_kpt_src.png}{1.22}{59.40} &
        \animage{0 41 0 40}{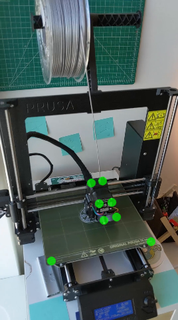} &
        \animage{0 41 0 40}{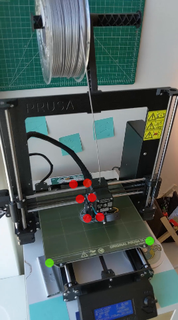} &
        \animage{0 41 0 40}{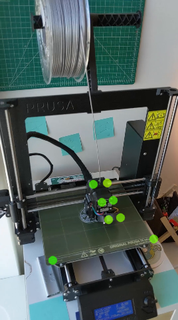} &
        \animage{0 41 0 40}{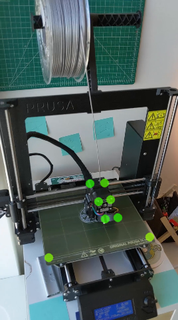}
        \\
        \makebox[20pt]{\raisebox{45pt}{\rotatebox[origin=c]{90}{\textsc{Chicken}}}}\hspace{-6pt} &
        \textimage{0 41 0 40}{hypernerf/vrig-chicken/32_kpt_src.png}{1.52}{33.46} &
        \animage{0 41 0 40}{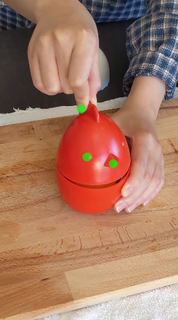} &
        \animage{0 41 0 40}{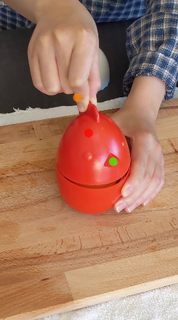} &
        \animage{0 41 0 40}{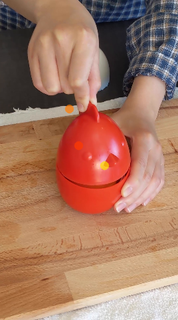} &
        \animage{0 41 0 40}{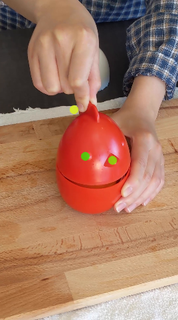}
        \\
        \makebox[20pt]{\raisebox{45pt}{\rotatebox[origin=c]{90}{\textsc{Peel Banana}}}}\hspace{-6pt} &
        \textimage{0 41 0 40}{hypernerf/vrig-peel-banana/32_kpt_src.png}{0.33}{33.77} &
        \animage{0 41 0 40}{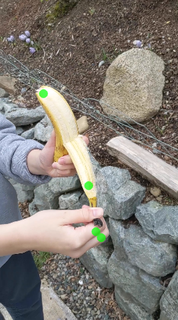} &
        \animage{0 41 0 40}{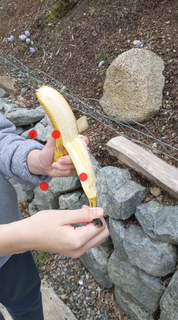} &
        \animage{0 41 0 40}{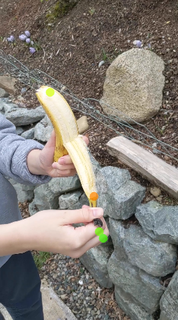} &
        \animage{0 41 0 40}{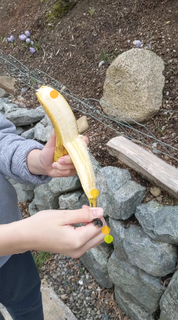}
        \\
        [2pt] &
        {\small Source Kpt.} &
        {\small Target Kpt.} &
        {\small NSFF~\cite{li2020nsff}} &
        {\small Nerfies~\cite{park2021nerfies}} &
        {\small HyperNeRF~\cite{park2021hypernerf}}
    \end{tabularx}
    \vspace{-4pt}
    \caption{
        \textbf{Additional qualitative results of keypoint transferring on the Nerfies-HyperNeRF dataset without camera teleportation.}
        $\Omega/\omega$ metrics of the input sequence are shown on the top-left.
        All models are trained under non-teleporting setting.
        Transferred keypoints are colorized by a heatmap of end-point error, overlaid on the ground-truth target frame.
    }
    \label{fig:correspondences_1}
\end{figure}

\begin{figure}[t!]
    \setlength{\tabcolsep}{0.8pt}
    \renewcommand{\arraystretch}{0.5}
    \begin{tabularx}{\textwidth}{@{}c*{5}{C}c@{}}
        \makebox[20pt]{\raisebox{45pt}{\rotatebox[origin=c]{90}{\textsc{Backpack}}}}\hspace{-6pt} &
        \textimage{0 0 0 0}{iphone2/backpack/32_kpt_src.png}{0.26}{5.59} &
        \animage{0 0 0 0}{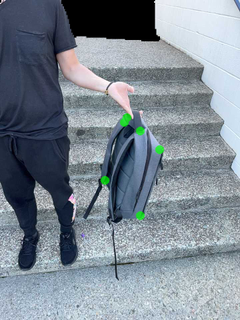} &
        \animage{0 0 0 0}{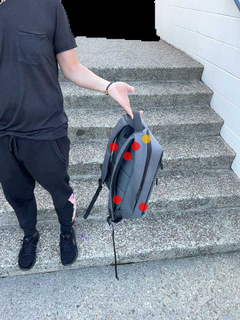} &
        \animage{0 0 0 0}{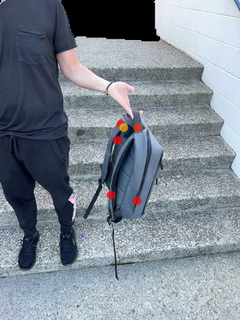} &
        \animage{0 0 0 0}{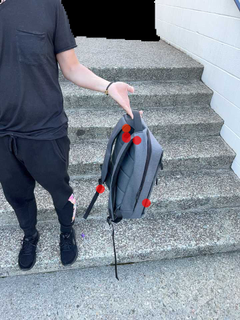} &
        \multirow{6}{*}{\cbar{colorbar.pdf}}
        \\
        \makebox[20pt]{\raisebox{45pt}{\rotatebox[origin=c]{90}{\textsc{Creeper}}}}\hspace{-6pt} &
        \textimage{0 0 0 0}{iphone2/creeper/32_kpt_src.png}{0.23}{14.05} &
        \animage{0 0 0 0}{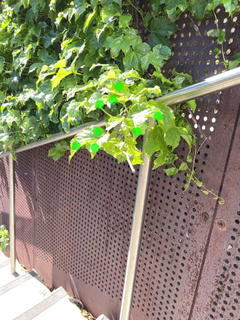} &
        \animage{0 0 0 0}{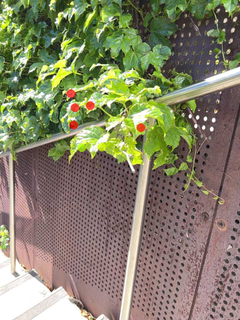} &
        \animage{0 0 0 0}{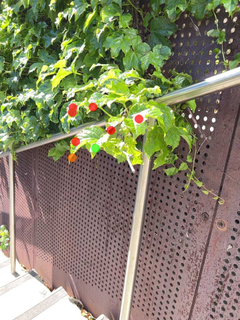} &
        \animage{0 0 0 0}{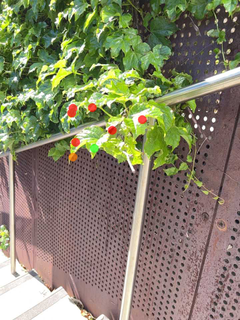}
        &
        \\
        \makebox[20pt]{\raisebox{45pt}{\rotatebox[origin=c]{90}{\textsc{Handwavy}}}}\hspace{-6pt} &
        \textimage{0 0 0 0}{iphone2/handwavy/32_kpt_src.png}{0.05}{13.66} &
        \animage{0 0 0 0}{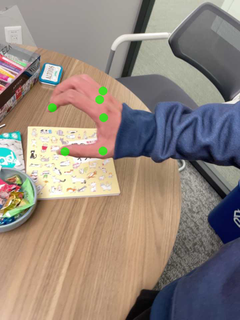} &
        \animage{0 0 0 0}{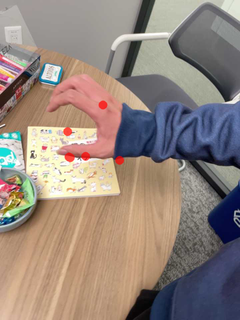} &
        \animage{0 0 0 0}{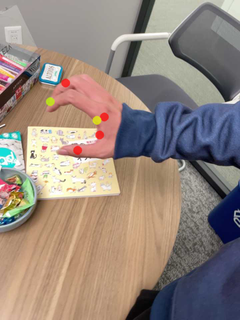} &
        \animage{0 0 0 0}{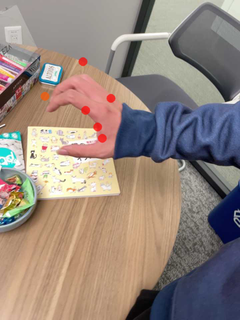}
        &
        \\
        \makebox[20pt]{\raisebox{45pt}{\rotatebox[origin=c]{90}{\textsc{Mochi}}}}\hspace{-6pt} &
        \textimage{0 0 0 0}{iphone2/mochi-high-five/32_kpt_src.png}{0.07}{14.49} &
        \animage{0 0 0 0}{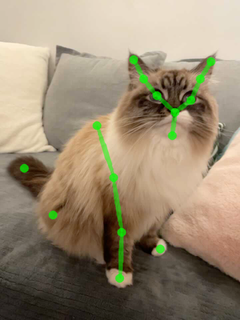} &
        \animage{0 0 0 0}{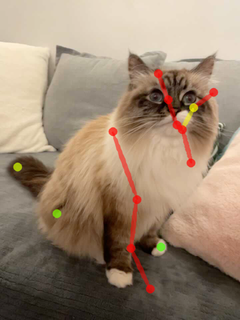} &
        \animage{0 0 0 0}{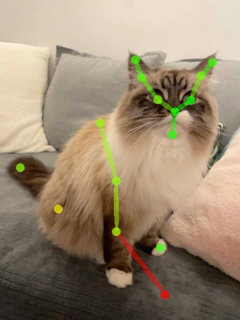} &
        \animage{0 0 0 0}{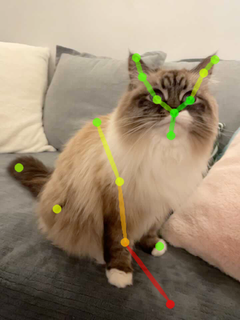}
        &
        \\
        \makebox[20pt]{\raisebox{45pt}{\rotatebox[origin=c]{90}{\textsc{Sriracha}}}}\hspace{-6pt} &
        \textimage{0 0 0 0}{iphone2/sriracha-tree/32_kpt_src.png}{0.18}{18.56} &
        \animage{0 0 0 0}{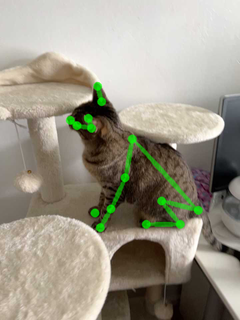} &
        \animage{0 0 0 0}{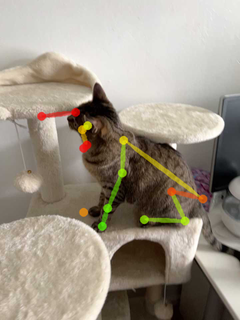} &
        \animage{0 0 0 0}{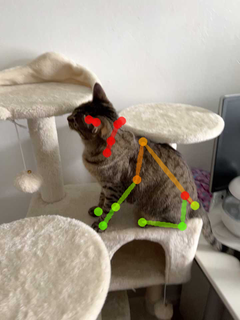} &
        \animage{0 0 0 0}{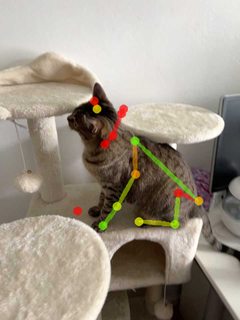}
        &
        \\
        \makebox[20pt]{\raisebox{45pt}{\rotatebox[origin=c]{90}{\textsc{Teddy}}}}\hspace{-6pt} &
        \textimage{0 0 0 0}{iphone2/teddy/32_kpt_src.png}{0.20}{7.62} &
        \animage{0 0 0 0}{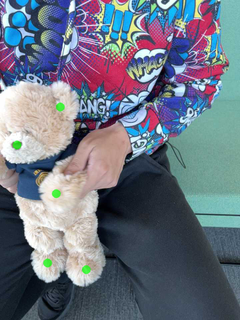} &
        \animage{0 0 0 0}{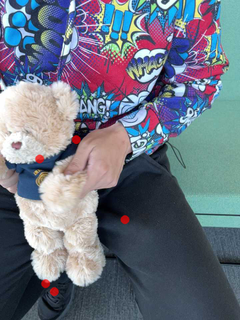} &
        \animage{0 0 0 0}{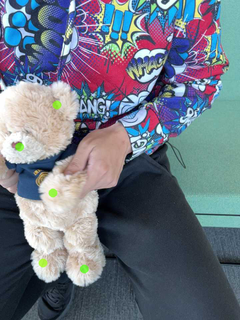} &
        \animage{0 0 0 0}{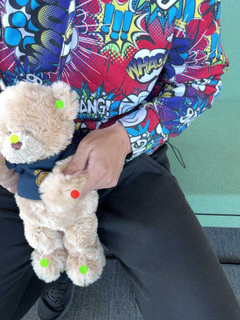}
        &
        \\
        [2pt] &
        {\small Source Kpt.} &
        {\small Target Kpt.} &
        {\small NSFF~\cite{li2020nsff}} &
        {\small Nerfies++} &
        {\small HyperNeRF++}
    \end{tabularx}
    \vspace{-4pt}
    \caption{
        \textbf{Additional qualitative results of keypoint transferring on the proposed iPhone dataset.}
        $\Omega/\omega$ metrics of the input sequence are shown on the top-left.
        All models are trained under non-teleporting setting.
        All models are also trained with all the additional regularizations (\textsc{+B+D+S}) except NSFF.
        Transferred keypoints are colorized by a heatmap of end-point error, overlaid on the ground-truth target frame.
    }
    \label{fig:correspondences_2}
\end{figure}
In Table~\ref{tab:app_corr_iphone_sv}, we provide additional quantitative results of the inferred correspondence on the single-camera captures from the proposed iPhone dataset.
In Figure~\ref{fig:correspondences_1} and~\ref{fig:correspondences_2}, we provide additional qualitative results of the inferred correspondence on both the Nerfies-iPhone dataset and the proposed iPhone dataset.
Note that all models are trained with additional regularizations on the proposed iPhone dataset except NSFF.

\end{document}